\documentclass[10pt]{article}

\usepackage[dvips]{graphicx}
\usepackage{cite}
\usepackage{amsmath,amssymb,subfigure,epsfig,bbm,stmaryrd,MnSymbol,mathrsfs,url}
\usepackage{pstricks}
\usepackage{subfigure}
\usepackage{booktabs}
\usepackage{multirow}
\usepackage{amsmath,amssymb}
\usepackage{pstricks}
\usepackage{amsmath,amssymb,subfigure,epsfig,bbm,stmaryrd,MnSymbol,mathrsfs}
\usepackage{subfigure}
\newcommand{\Zm}{\mathbf{Z}}
\newcommand{\Ym}{\mathbf{Y}}

\newcommand{\bfx}{\mathbf{x}}
\newcommand{\bfm}{\mathbf{m}}

\newcommand{\tho}{^{\text{th}}}

\newcommand{\Xm}{\mathbf{X}}

\newcommand{\unfld}{\mbox{\tt unfold}}
\newcommand{\fld}{\mbox{\tt fold}}

\newcommand{\bcircu}{\mbox{\tt bcirc}}

\title{Tensor-based formulation and nuclear norm regularization for multi-energy computed tomography}

\author{Oguz Semerci\thanks{Schulumberger-Doll Research Center, {\tt osemerci@slb.com}}, Ning Hao\thanks{Department of Mathematics, Tufts University.}, Misha E. Kilmer$^{\dagger}$ and Eric L. Miller\thanks{School of Electrical and Computer Engineering, Tufts University.}}

\begin{document}
\maketitle

%

%
%
%

        
%



\begin{abstract}

The development of energy selective, photon counting X-ray detectors allows for a wide range of new possibilities in the area of computed tomographic image formation. Under the assumption of perfect energy resolution, here we propose a tensor-based iterative algorithm that simultaneously reconstructs the X-ray attenuation distribution for each energy. We use a multi-linear image model rather than a more standard "stacked vector" representation in order to develop novel tensor-based regularizers. Specifically, we model the multi-spectral unknown as a 3-way tensor where the first two dimensions are space and the 3$^{\text{rd}}$ dimension is energy. This approach allows for the design of tensor nuclear norm regularizers, which like its two dimensional counterpart, is a convex function of the multi-spectral unknown. The solution to the resulting convex optimization problem is obtained using an alternating direction method of multipliers (ADMM) approach. Simulation results shows that the generalized tensor nuclear norm can be used as a stand alone regularization technique for the energy selective (spectral) computed tomography (CT) problem and when combined with total variation regularization it enhances the regularization capabilities especially at low energy images where the effects of noise are most prominent.

\end{abstract}




\section{Introduction}


A conventional computed tomography (CT) imaging system utilizes energy integrating detector technology \cite{whiting2006properties} and provides a monochromatic reconstruction of the linear attenuation coefficient distribution of an object under investigation. The polychromatic nature of the X-ray spectra is either neglected \cite{bouman,pan2009commercial} or incorporated into the model in an iterative reconstruction method to achieve more accurate results \cite{elbakri2002statistical, semerci}. However, neglecting the polychromatic nature may cause the loss of significant energy dependent information  \cite{wang2008outlook, elbakri2002statistical,de2001iterative}. A multi-energy CT system, on the other hand, distinguishes specific energy regions of the polychromatic spectra at the detector side. Energy selective detection is accomplished with the use of photon counting detectors (PCDs) \cite{shikhaliev2008energy}, instead of the energy integrating detectors \cite{whiting2006properties} used in conventional and dual energy CT.

PCDs, which are also referred to as energy discriminating detectors, have the ability to identify individual photons and classify them according to their energy. This property allows the recovery of spectral properties of the object being imaged and opens the door to ``color'' CT technology with the simplicity of monochromatic reconstruction models \cite{gao2011multi}. Multi-energy CT promises improved diagnostic medical imaging \cite{shikhaliev2011photon, schlomka2008experimental} as well as in the security domain \cite{ivakhnenko2010novel} due to its contrast enhancement and ability to characterize material composition.

Within an energy integrating detector, incoming photons are converted to electrical charge and accumulated on a detector, which is read out to determine the output signal. The latter step is the source of so called \emph{detector-read-out} noise, which degrades the image quality \cite{semerci}. In a photon counting detector, on the other hand, an incoming photon is converted to an electrical pulse, whose amplitude is determined by the energy of the photon and the output signal is based on a counter that is incremented according to the charge of the electric pulse \cite{iwanczyk2009photon}. This direct relationship between the counter and an incoming photon with a certain energy eliminates the main cause of the detector-read-out noise. Hence, in addition to their energy discriminating properties, PCDs offer better signal quality compared to energy integrating detectors \cite{barber2009characterization}. 


The driving application of our work is security \cite{ying, semerci,singh2003explosives}. More specifically, the possibility of reconstructing the total attenuation distribution as a function of energy indicates the applicability of multi-energy CT to the luggage screening problem, as accurately reconstructed attenuation curves of nominal objects in luggage potentially lead to material identification. Nevertheless, the methods considered here are more broadly applicable both to the application of multi-energy CT for medical imaging as well as to other multi-linear inverse problems.


We propose an iterative reconstruction method for the multi-energy CT problem where we model the multi-spectral unknown as a low rank 3-way (third order) \emph{tensor}. With the term \emph{tensor} we refer to the multidimensional generalization of matrices, \emph{i.e.}, matrices are two-dimensional (2-way) tensors. Recently, there has been considerable work on recovering corrupted matrices or tensors based on low-rank and sparse decomposition \cite{candes2009robust} or solely on low-rank assumptions \cite{cai2008singular,candes2009exact,tomioka2010estimation,liu2009tensor}. These ideas were also applied to 4D cone beam CT  \cite{cai2012cine} and spectral tomography \cite{gao2011multi}, where the multi-linear unknown is modelled as a superposition of low rank and sparse matrices. In those efforts, the multi-linear unknown is represented as a matrix where each column is the lexicographically ordered collection of pixels at a given energy or time. The authors applied low rank plus sparse decomposition to this multi-linear unknown where the matrix nuclear norm penalty is applied to the low rank component. 


We take a different approach and exploit the inherent tensorial nature of the multi-energy CT problem allowing us to make use of a broader collection of tools for the analysis of these structures. To date, tensor decomposition tools such CANDECOMP/PARAFAC (CP) and \cite{carroll1970analysis}, Tucker \cite{tucker1966some} decompositions have found application in data mining and analysis for chemistry, neuroscience, computer vision and communications \cite{acar2009unsupervised, andersen2003practical,kolda2008scalable}. The higher-order generalization of the singular value decomposition (SVD), which is also referred to as multi-dimensional SVD \cite{de2000multilinear}, has been used for image processing applications such as facial recognition \cite{vasilescu2002multilinear}.

Despite being efficient tools for multidimensional data processing, to find these decompositions requires the solution of a difficult non-convex optimization problem that also has poor convergence properties. Moreover, for CP and Tucker methods the number of components (unknowns) needs to be known \emph{a priori} \cite{acar2009unsupervised,kolda2009tensor,bro2003new}. Thus here we consider an alternate approaches in which these tensor decomposition ideas form the basis for a generalization of the sparsity promoting nuclear norm concepts that have received so much attention recently \cite{candes2009exact,recht2010guaranteed}. 

For the first approach, we are motivated by \cite{tomioka2010estimation,liu2009tensor,gandy2011tensor} where the idea of matrix completion via nuclear norm minimization is generalized to the tensor case using the matricization (unfolding) operation. The unfolding operation refers to rearranging the columns of a tensor along a certain mode or dimension into a matrix \cite{kolda2009tensor} (see Section \ref{sec:preliminaries} for a more detailed explanation). The multidimensional nuclear norm, or the \emph{generalized tensor nuclear norm}, is obtained by the summation of nuclear norms of the unfoldings in each mode. Successful results were reported for tensor completion for multi-spectral imaging \cite{signoretto2011tensor}, color image impainting \cite{signoretto2010nuclear,signoretto2011learning} and multi-linear classification and data analysis \cite{signoretto2011learning,tomioka2010estimation} but, to the best of our knowledge have not been considered for use in a linear inverse problems context before.  This then represents a first contribution of this paper.

We use this simple, yet effective generalization in the multi-energy CT problem \cite{semerci:12:air} where we assume the multi-spectral unknown is low rank in each of its unfoldings and construct a regularizer. The resulting tensor nuclear norm regularizer (TNN-1) allows fast processing and has satisfactory noise reduction capabilities. Applying the low rank prior to the multi-spectral matrix, which has vectorized images of different energies in its columns, is a special case of our tensor model where only the unfolding in the energy dimension is considered \cite{signoretto2011tensor}. Our approach provides a more powerful regularization method for the case where the number of energy bins is limited and redundancy in the spatial dimensions can be exploited with the incorporation of unfoldings in spatial dimensions \cite{liu2009tensor, signoretto2011tensor}. One of the contributions of this work is to demonstrate the benefits of low-rank assumptions on the unfoldings in the spatial dimensions to design regularizers.

The generalized tensor nuclear norm is based on the rank of each unfolding which give a weak upper bound on the rank of a tensor  \footnote{Tensor rank is defined as the minimal number of 3-way outer products of vectors needed to express the tensor. We refer the reader to Section 3 of Kolda and Bader \cite{kolda2009tensor} for more details.} \cite{kolda2009tensor}. However, it does not exploit the correlations among all the dimensions simultaneously. With this motivation, as the second approach, we propose a new tensor nuclear norm based on tensor singular value decomposition (t-SVD), which is introduced by Kilmer and Martin \cite{Kilmer2011641} and has been proved to be useful for applications such as facial recognition and image deblurring \cite{kilmer2011third}. The t-SVD is based on a new tensor multiplication scheme and has similar structure to the matrix SVD which allows optimal low-rank representation (in terms of the Frobenius norm) of a tensor by the sum of outer product of matrices \cite{Kilmer2011641}. We devise a new tensor nuclear norm based on t-SVD, which leads to our second regularizer (TNN-2). Similar to TNN-1, TNN-2 can be written in a matrix nuclear norm form. Introduction of this new tensor nuclear norm and its utilization for regularization is the second contribution of this paper.

In addition, we combine TNN-1 and TNN-2 with total variation (TV) regularization \cite{rudin1992nonlinear}. Typically, edge enhancement/preservation is crucial for all imaging applications. One of the most widely used edge preserving regularization technique is total variation (TV) \cite{rudin1992nonlinear} which has been applied to CT as well \cite{sidky2008image,tang2009performance}. With the expectation that the spatial structure of the image at each energy is appropriately regularized using total variation, in  this work, we use the summation of the TV of images at each energy as a regularizer.

Although the images at different energies are treated independently with TV, the low rank assumptions on the multi-dimensional unknown results in the implicit coupling of information across the energy dimension. Therefore, when we combine TV with TNN-1 or TNN-2, the accuracy of the reconstructions, especially at low energies where the noise level is higher due to reduced photon counts, are enhanced. As materials are better distinguished at low energies reliable recovery of low energy images is a significant benefit of our approach.



The paper is organized as follows: Section \ref{sec:preliminaries} describes the preliminaries on tensors and gives the notation that will be used throughout the paper. Section \ref{sec:forward} describes the measurement model and the multi-spectral phantom in the form of a 3-way tensor. Section \ref{sec:reg}, after a brief introduction to the rank minimization problem, provides the details of the tensor based modeling of the unknown and mathematical details about our nuclear norm regularizers. In Section \ref{sec:inversion} we give the details of the ADMM algorithm that is used in inversion. Section  \ref{sec:results} shows simulation results and Section \ref{sec:conclusion} gives concluding remarks and future directions.

%
%
%

\section{Preliminaries on Tensors}\label{sec:preliminaries}

\begin{figure*}[t!]
\centering
\includegraphics[width=4in, trim = 0mm 0mm 0mm 0mm, clip=true]{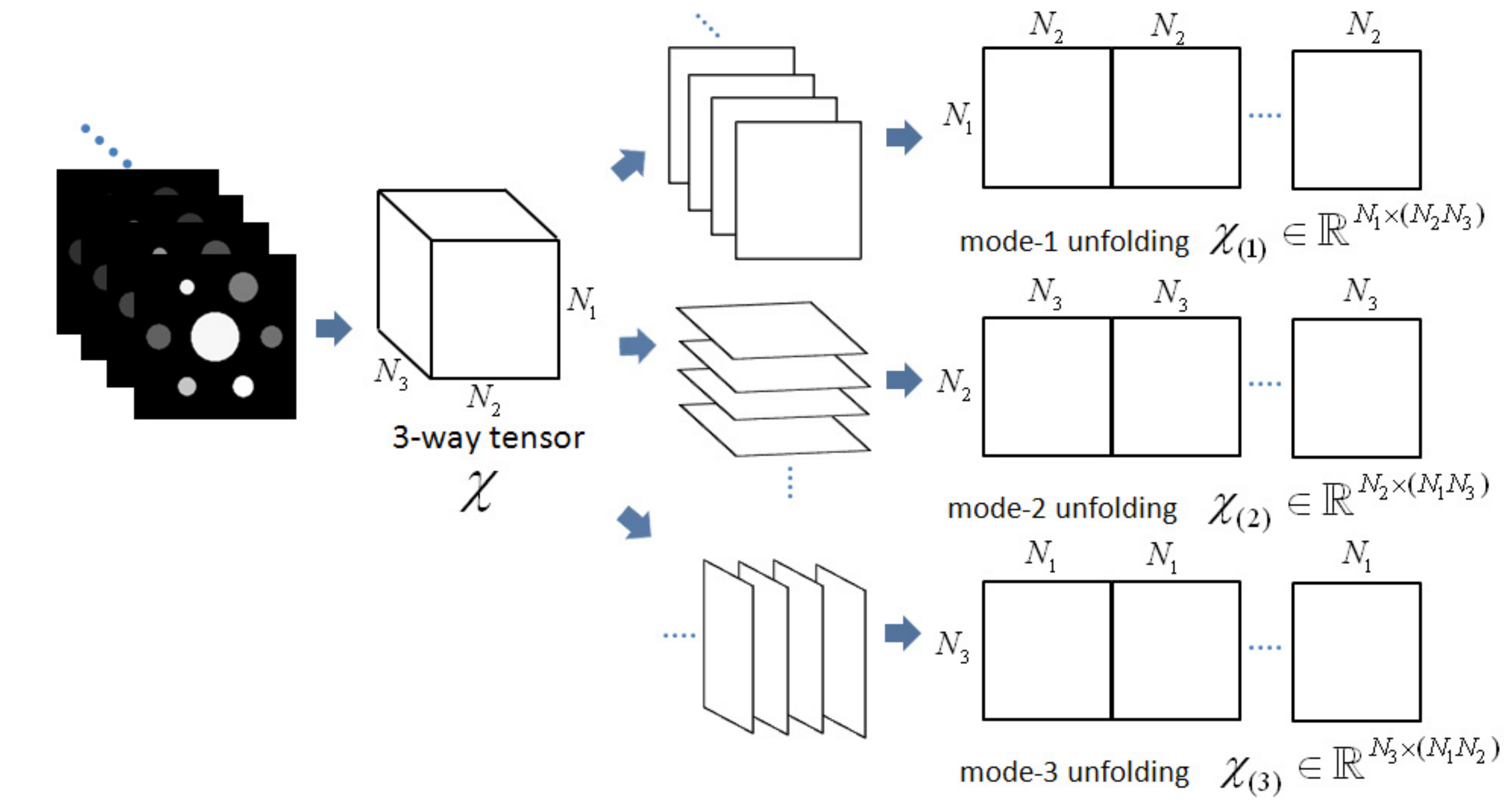}
\caption{Mode-1, mode-2 and mode-3 unfoldings of a 3-way tensor $\chi$. It is easy to visualize unfolding operations in terms of frontal, horizontal and lateral slices. The unfolding operation corresponds to aligning the corresponding slices for each mode next to each other.}
\label{fig:unfoldings}
\end{figure*}

In this section, we give the definitions and the notation that will be used throughout the paper. For a more comprehensive discussion, we refer the reader to the review by Kolda and Bader \cite{kolda2009tensor}, and Kilmer and Martin \cite{Kilmer2011641}. A $K$-way tensor is a multi-linear structure in $\mathbb{R}^{N_1\times N_2\times\hdots\times N_K}$. The unfolding (matricization) operation is defined as the following: For a $K$-way tensor, the mode-$l$ unfolding $\chi_{(l)} \in \mathbb{R}^{N_l\times \prod_{k'\neq l}N_k'} $ is a matrix whose columns are mode-$l$ fibers, where mode-$l$ fibers are vectors in $\mathbb{R}^{N_l}$ that are obtained by varying the index in the $l^{\text{th}}$ dimension of the tensor and fixing the others \cite{kolda2009tensor}. As shown in Fig. \ref{fig:unfoldings}, for 3-way tensors, we can visualize the unfolding operations in terms of frontal, horizontal and lateral slices. Although we do not give formal definitions for horizontal and lateral slices, which are obvious from Fig. \ref{fig:unfoldings}, we denote the $k\tho$ frontal face of a 3-way tensor $\chi$ by $\mathbf{X}^{(k)}$, as this notation will be useful.

Folding and unfolding operators can be represented with permutation of lexicographically ordered vectors \cite{tomioka2010estimation}. Specifically, the mode-$l$ unfolding maps the tensor element $(i_1,\hdots,i_{N_K})$ to the matrix element $(i_l,j)$ according to \cite{kolda2009tensor}
\begin{equation}
j=1+\sum_{\substack{k=1 \\ k\neq l}}^N(i_k-1)J_k,\;\; \text{where}\;\; J_k=\prod_{\substack{m=1 \\ m\neq l}}^{k-1}N_m.
\label{permut}
\end{equation}
Let us denote the vectorized form of $\chi$ by $\bfx$ and of $\chi_{(l)}$ by $\bfx_l$. Then the relationship between $\bfx$ and $\bfx_l$ is
\begin{equation*}
\bfx_l = \mathbf{P}_l\bfx \;\;\text{and}\;\; \bfx = \mathbf{P}_l^{\mathrm{T}}\bfx_l,
\end{equation*}
where $\textbf{P}_l \in \mathbb{R}^{N_1N_2...N_{K}\times N_1N_2...N_{K}}$ is the permutation matrix that corresponds to the $l^{\text{th}}$ unfolding operation given by (\ref{permut}). Note that $\bfx$ and $\bfx_1$ are equal and $\mathbf{P}_1$ is the identity matrix. This notation will be useful in Section \ref{sec:inversion}.

To construct our second tensor nuclear norm regularization approach in Section \ref{sec:reg2} we need the t-SVD of \cite{Kilmer2011641} which in turn requires that we introduce three operators: $\fld$, $\unfld$ and $\bcircu$. While the mode-1 unfolding aligns frontal slices next to each other, the $\unfld(\chi)$ operation aligns them on top of each other:
\begin{equation*}
\unfld(\chi)= \begin{bmatrix}{\Xm}^{(1)}\\
{\Xm}^{(2)}\\
\vdots\\
{\Xm}^{(N_3)}\\
\end{bmatrix} 
\end{equation*}
and $\fld(\unfld(\chi))$ folds them back to a tensor form:
\begin{equation*}
\fld(\unfld(\chi))=\chi.
\end{equation*}
Using the ${\bold X}^{(k)}$'s, one can form the block circulant matrix $\bcircu(\chi) \in \mathbb{R}^{N_3N_1 \times N_3N_2}$ as follows:
\begin{equation}
\label{block_circ}
\bcircu(\chi)=\begin{bmatrix}{\bold X}^{(1)}&{ \bold X}^{(N_3)}&{\bold X}^{(N_3-1)}&\hdots&{\bold X}^{(2)}\\
{\bold X}^{(2)}&{\bold X}^{(1)}&{\bold X}^{(N_3)}&\hdots&{\bold X}^{(3)}\\
 \vdots&\ddots&\ddots&\ddots&\vdots\\
 {\bold X}^{(N_3)}&{\bold X}^{(N_3-1)}&{\bold X}^{(N_3-2)}&\hdots&{\bold X}^{(1)}\\
 \end{bmatrix}. 
\end{equation}

The $n$-mode product of a $K$-way tensor $\chi  \in \mathbb{R}^{N_1\times N_2\times\hdots N_K}$ with matrix $\mathbf{U}\in \mathbb{R}^{J\times N_n}$ produces a tensor in with size $N_1\times \hdots \times N_{n-1} \times J \times N_{n+1} \times \hdots \times N_K$ and defined as
\begin{equation*}
(\chi \times_n \mathbf{U})_{i_1\hdots i_{n-1}j\,i_{n+1}\hdots i_K} = \sum_{i_n=1}^{N_n}x_{i_1i_2\hdots i_K}\,u_{ji_n},
\end{equation*}
where $\times_n$ is the $n$-mode product operation.

While the $n$-mode product defines an operation between a tensor and a matrix, multiplication of 3-way tensors can be performed using the \emph{t-product} \cite{Kilmer2011641}. For $\chi_1 \in \mathbb{R}^{N_1\times N_2\times N_3}$ and $\chi_2 \in \mathbb{R}^{N_2\times \ell \times N_3}$ the t-product is given as
\begin{equation}
\chi_1\ast \chi_2 =\fld(\bcircu(\chi_1) \unfld(\chi_2)).
\label{tensorprod}
\end{equation}
Notice that, $\chi_1\ast \chi_2$ is in $\mathbb{R}^{N_1 \times \ell \times N_3}$. The t-product defined in (\ref{tensorprod}) is the basis for the t-SVD (tensor SVD) and the regularizer, TNN-2, which is introduced in Section \ref{sec:reg2}. Given $\chi \in \mathbb{R}^{N_1\times N_2\times N_3}$, its transpose, $\chi^{\mathrm{T}} \in \mathbb{R}^{N_2\times N_1\times N_3}$, is obtained by applying matrix transpose to each frontal face and then reversing the order of transposed frontal slices 2 through $N_3$:
\begin{equation*}
 \chi^{\mathrm{T}}=\fld \left( \begin{bmatrix}({\bold X}^{(1)})^{\mathrm{T}}\\
({\bold X}^{(N_3)})^{\mathrm{T}}\\
\vdots \\
({\bold X}^{(3)})^{\mathrm{T}}\\
({\bold X}^{(2)})^{\mathrm{T}}\end{bmatrix}  \right).
\end{equation*}
The tensor $\mathcal{Q}$ is orthogonal in the sense of the t-product if
\begin{equation*}
\mathcal{Q}^{\mathrm{T}}\ast\mathcal{Q}=\mathcal{Q}\ast \mathcal{Q}^{\mathrm{T}}=\mathcal{I},
\end{equation*}
where $\mathcal{I}$ is the identity tensor whose first frontal face is the $\ell\times \ell$ identity matrix, and whose other frontal slices are all zeros.

We now review the block diagonalization property of block circulant matrices. For any block circulant matrix $\bcircu(\chi)\in \mathbb{R}^{N_3N_1 \times N_3N_2}$ we have
\begin{equation}
\label{block_diag}
\begin{split}
({\bold F}_{N_3}\otimes {\bold I}_{N_1})\cdot \bcircu
(\chi) & \cdot({\bold F}_{N_3}^{\ast}\otimes \bold{I}_{N_2}) \\& = 
 \begin{bmatrix}
\hat{{\bold X}}^{(1)}& & & \\
 & \hat{{\bold X}}^{(2)} & & \\
 & &\ddots & \\
  & & & \hat{{\bold X}}^{(N_3)}
\end{bmatrix},
\end{split}
\end{equation}
where ${\bold I}_{N_1}$ and ${\bold I}_{N_2}$ are the identity matrices in $\mathbb{R}^{N_1\times N_1}$ and $\mathbb{R}^{N_2\times N_2}$, respectively, ${\bold F}_{N_3}\in \mathbb{R}^{N_3 \times N_3}$ is the normalized discrete Fourier Transform matrix \cite{golub1996matrix} and $\hat{\Xm}^{(n)}$'s are the frontal faces of the tensor $\hat{\chi}$, which is obtained by applying the Fast Fourier Transform (FFT) to the mode-3 fibers of $\chi$. We will use this notation, \emph{ie.,} $\hat{\chi}$ and $\hat{\Xm}^{(n)}$ for the tensor $\chi$ and its $n\tho$ frontal face in the Fourier domain in Section \ref{sec:reg2}.

Finally, we note that $\bcircu(.)$ is a linear operation, which can be written in terms of permutation matrices. Let ${\bold x}$ denote the vectorized version of $\chi$ as described before, and let ${\bold x}_\mathrm{c}$ denote the vectorized version of  $\bcircu(\chi)$. Then, we have
\begin{equation}
\label{bcircu_matrix}
\bfx_\mathrm{c} = \mathbf{P}_\mathrm{c}\bfx = \begin{bmatrix}
\mathbf{P}_{\mathrm{c},1}\\
\mathbf{P}_{\mathrm{c},2}\\
\vdots \\
\mathbf{P}_{\mathrm{c},N_3}
\end{bmatrix}\bfx,
\end{equation}
where $\mathbf{P}_{\mathrm{c},i}$'s reorder the elements of $\bfx$ according to the column blocks of (\ref{block_circ}).

\section{The Measurement Model and The Multi-spectral Unknown as a Tensor}\label{sec:forward}

\begin{figure}[t!]
\centering
\includegraphics[width=2in, trim = 0mm 0mm 0mm 0mm, clip=true]{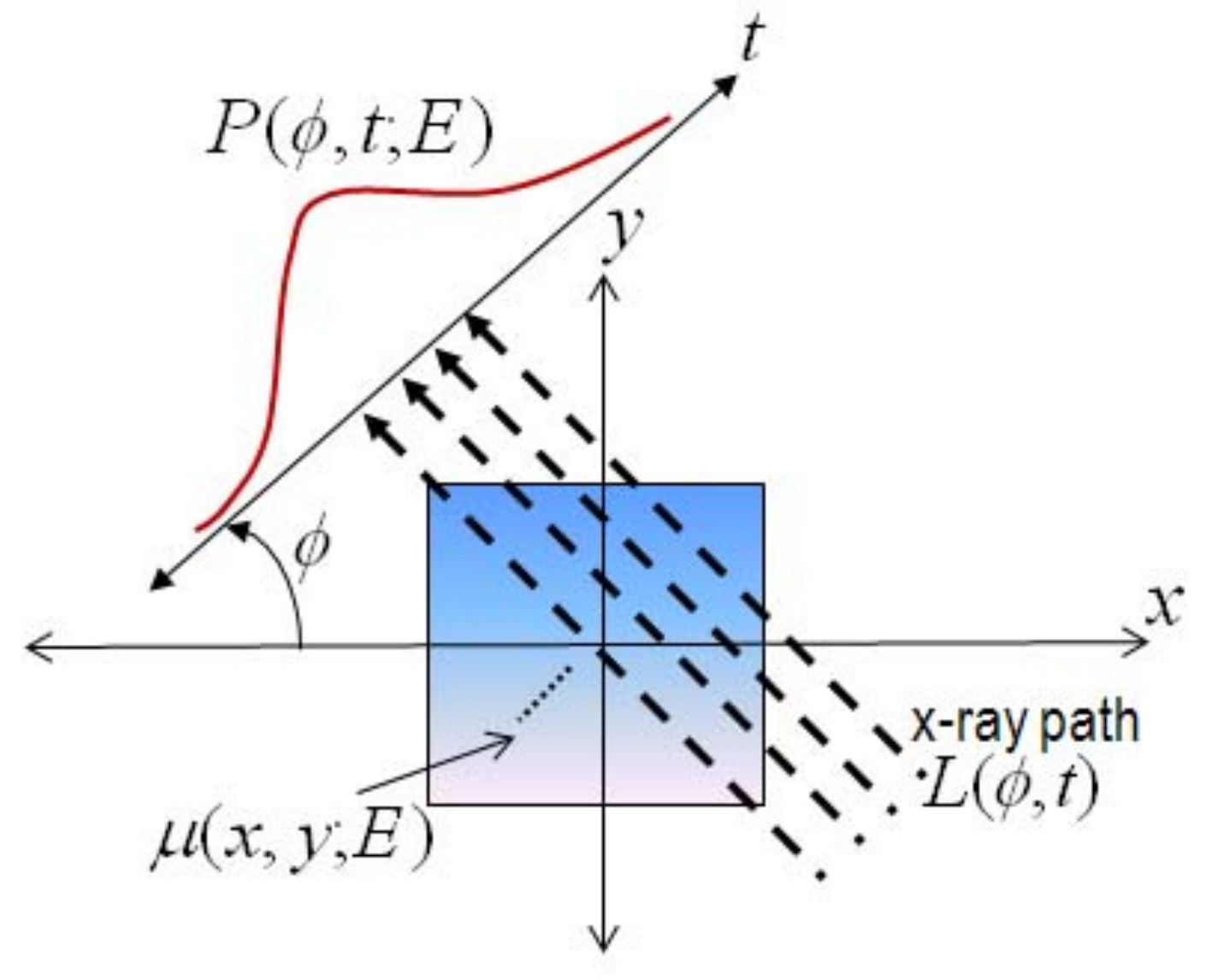}
\label{fig:radon}\caption{Parallel beam X-ray measurement geometry.}
\end{figure}

Polychromatic CT \cite{semerci2012tomographic, elbakri2002statistical, de2001iterative} is based on the projection model
\begin{equation}
\label{poly}
P(\phi,t)=\int S(E)\exp \left( - \int_{L(\phi,t)} \mu(\mathbf{r},E)\; \mathrm{d}\mathbf{r} \right)\mathrm{d}E
\end{equation}
where $\mu(\mathbf{r},E)$ is the energy dependent attenuation coefficient, $S(E)$ is the source spectrum and $(\phi,t)$ parametrizes the x-ray path $L(\phi,t)$. In this work, we used parallel beam measurement geometry \cite{beutel2000handbook} as depicted in Fig. 2. Under the assumption of infinitesimal detector bin width (\emph{i.e.}, perfect energy resolution) the polychromatic projection given in (\ref{poly}) simplifies to a monochromatic one, where $s_k = S(E)\delta(E-E_k)$, resulting in the following model for the data corresponding to the $k^{\text{th}}$ bin:
\begin{equation}
\label{mono}
\begin{split}
P(\phi,t;k)= &s_k\exp \left( - \int_{L(\phi,t)} \mu(\mathbf{r},E_k)\; \mathrm{d}\mathbf{r} \right)\\
&\text{for}\,\, k=1,...,N_3,
\end{split}
\end{equation}
where $N_3$ is the number of energy bins. We refer the reader to \cite{schmidt2012empirical} for an example of a fully polychromatic energy-resolved CT model.

In order to obtain a discrete representation of (\ref{mono}), we discretize each $\mu(\mathbf{r},E_k)$ into images of $N_p=N_1N_2$ pixels: $\mathbf{x}_k\in \mathbb{R}^{N_p}\; \text{for}\; k=1,...,N_3$ where $N_1$ and $N_2$ refers to the number of pixels in spatial dimensions $x$ and $y$. We also discretize the $(\phi,t)$ space into $N_m$ source detector pairs and introduce the system matrix $\mathbf{A} \in \mathbb{R}^{N_m\times N_p}$ where $[\mathbf{A}]_{ij}$ represents the length of that segment of ray $i$ passing through pixel $j$. Incorporating the Poisson statistics of X-ray interactions, the multi-energy measurement model is written
\begin{equation}
\label{measurement}
y_{k,j} = \text{Poisson} \left\{ s_k \exp{[\mathbf{A}\mathbf{x}_k]_j} \right\},
\end{equation}
where $k$ and $j$ index detector bins and source-detector pairs respectively. Note that, the electronic noise can be neglected for PCDs, This is different than conventional CT where energy integrating detectors are used \cite{iwanczyk2009photon, barber2009characterization}.

\begin{figure*}[t!]
\centering
$\begin{array}{cc}
\includegraphics[width=2.1in, trim = 0mm 0mm 0mm 0mm, clip=true]{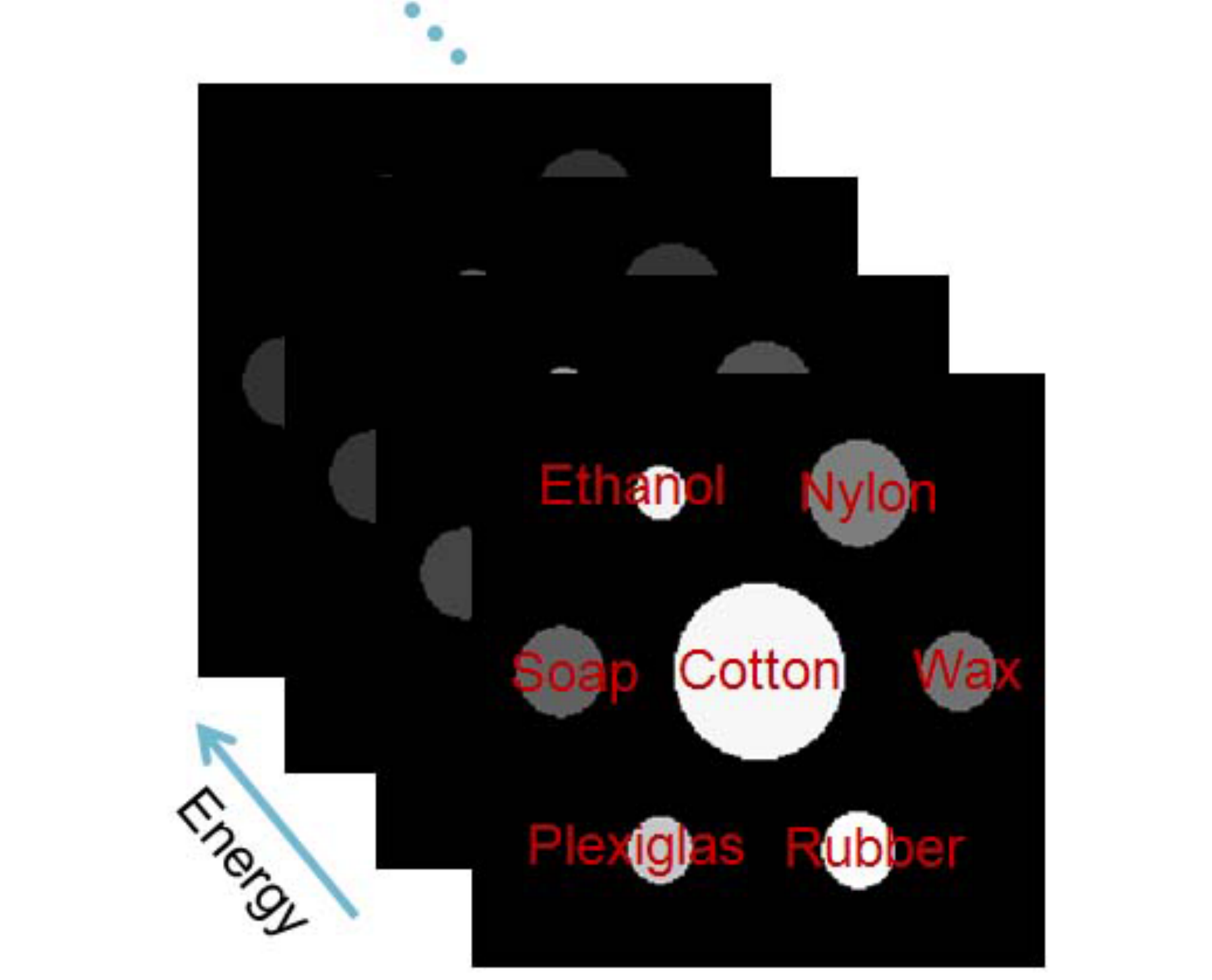}  &
\includegraphics[width=2.2in, trim = 0mm 0mm 0mm 0mm, clip=true]{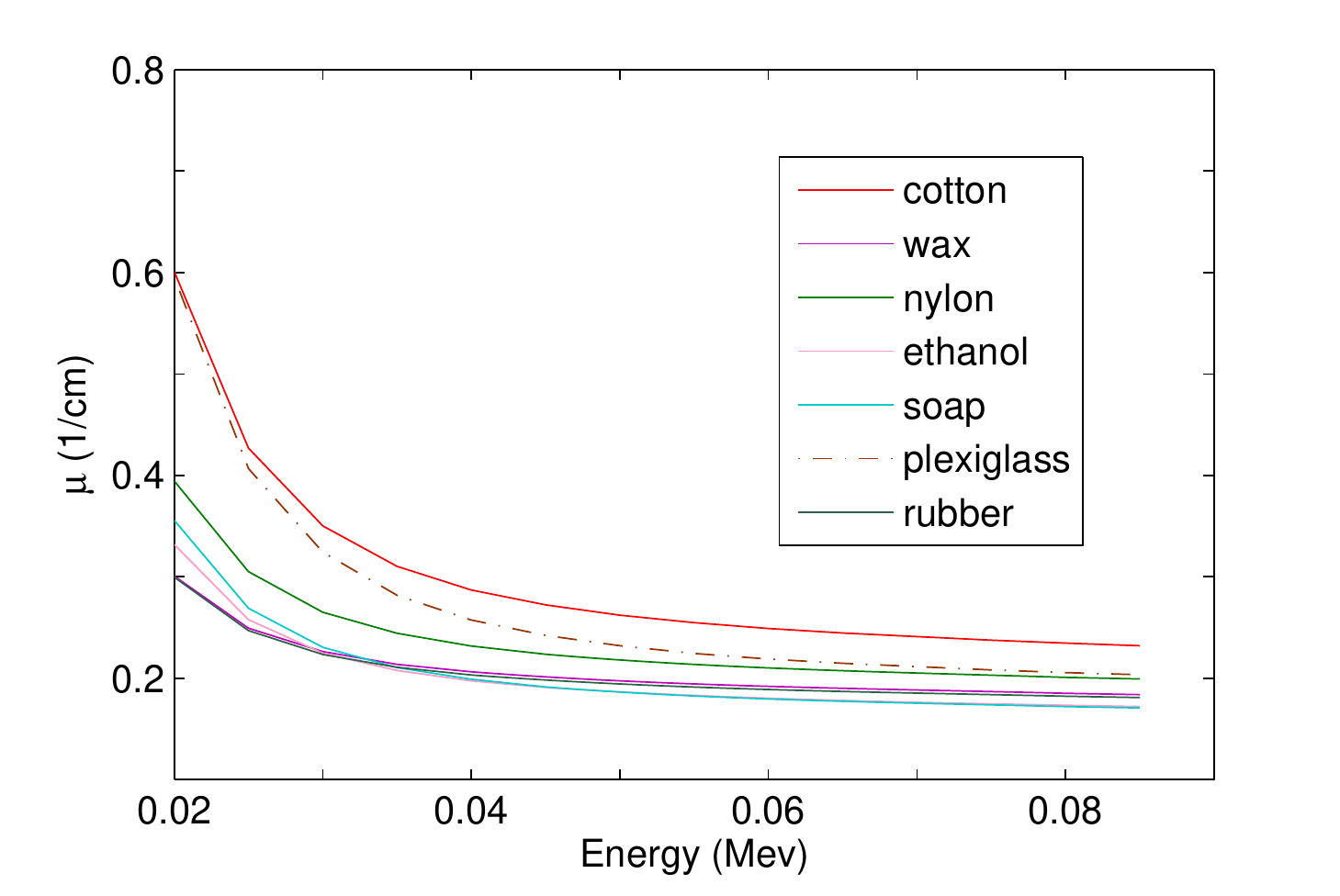}
\end{array}$
\caption{ Multi-spectral phantom and the attenuation curves for existing materials.}
\label{fig:phantom}
\end{figure*}

Our goal is to develop an image formation method that treats all of the $\bfx_k$'s in a unified manner, as done in \cite{gao2011multi}, rather than reconstructing each independently of the others \cite{ivakhnenko2010novel}. Towards this aim, we utilize tensors which are multi-linear generalizations of vectors and matrices. Specifically, we define the three-way ($3^{\text{rd}}$-order) tensor $\chi \in \mathbb{R}^{N_1\times N_2\times N_3}$, where first two are spatial dimensions the third dimension is energy, and the $\bfx_k$'s are the lexicographical ordering of the $N_1\times N_2$ frontal slices. A depiction of the multi-spectral phantom used in this study along with the corresponding attenuation curves are given in Figure \ref{fig:phantom}. Note that the multi-linear structure can be extended to higher dimensions for different classes of problems. For example, one can consider a 5D dynamical problem with additional 3$^{\text{rd}}$ spatial dimension and time dependency. The goal of the multi-energy CT problem in this paper however is to reconstruct $\chi$ given $y_{k,j}$ for $k=1,\hdots,N_3$ and $y=1,\hdots,N_m$.

\section{Low-Rank modeling and Regularization} \label{sec:reg}

As mentioned before, low-rank modeling is an important tool not only for compressing \cite{ye2005generalized} and analyzing \cite{vasilescu2002multilinear,jolliffe2005principal} large data sets but also for regularization and the incorporation of prior information \cite{lefkimmiatis2012hessian,gao2011multi}. Traditionally, low-rank modeling is applied to a matrix variable, which is assumed to be low order or of low complexity \cite{fazel2001rank}. As the multi-linear generalizations, such as our multi-spectral unknown in the form of a 3-way tensor, are closely related to the matrix case, we briefly describe the rank minimization problem of matrices in this section.

Let the matrix $\Xm \in \mathbb{R}^{N\times M}$ denote the unknown variable that is assumed to be low-rank. For instance, $\Xm$ can be the system parameters of a low-order control system \cite{fazel2001rank}, a low dimensional representation of data \cite{turk1991face}, or adjacency matrix of a network graph \cite{hsieh2012low}. The problem of estimating $\hat{\Xm}$ with minimal rank from the output $\bfm$ of a system $K$ can be formulated as a minimization problem:
\begin{equation}
\label{min_rank}
\begin{aligned}
&\underset{\Xm \in  \mathbb{R}^{M\times N}}{\mathrm{minimize}} & & \mathrm{rank}(\Xm) \\
&\mathrm{subject\;to} & & K(\Xm) = \bfm.
\end{aligned}
\end{equation}
%
However, minimization of $\mathrm{rank}(\Xm)$, which is a non-convex function of $\Xm$, is an NP-hard problem \cite{recht2010guaranteed}. Consequently, Fazel \emph{et al.} \cite{fazel2001rank} proposed the replacement of the rank function with the nuclear norm, which is defined as 
\begin{equation*}
\Vert\Xm\Vert_\ast:=\sum_i^{\mathrm{min}(M,N)}\sigma_i(\Xm),
\end{equation*}
where $\sigma_i$'s are the singular values of the matrix $\Xm$. This replacement results in the following optimization problem.
\begin{equation}
\label{min_nuc}
\begin{aligned}
&\underset{\Xm \in  \mathbb{R}^{M\times N}}{\mathrm{minimize}} & & \Vert\Xm\Vert_\ast \\
&\mathrm{subject\;to} & & K(\Xm) = \bfm.
\end{aligned}
\end{equation}
The minimization problem (\ref{min_nuc}) is motivated by the fact that the nuclear norm provides the tightest convex relaxation for the rank operation in matrices \cite{candes2009exact}. The replacement of rank with the nuclear norm is analogous to the use of the $\ell_1$ norm as a proxy to the $\ell_0$ semi-norm to achieve sparse signal reconstructions \cite{donoho2006compressed}. Analysis of this convex relaxation technique and the equivalence of (\ref{min_rank}) and (\ref{min_nuc}) for compressed sensing are analyzed in \cite{recht2010guaranteed}. In the sequel we interpret the nuclear norm term as a regularizer and seek solutions to problems in the form
\begin{equation}
\label{min_nuc2}
\tilde{\Xm} := \underset{\Xm \in  \mathbb{R}^{M \times N}}{\mathrm{argmin}} \, \Vert K(\Xm)-\bfm \Vert^2_2 + \gamma \Vert\Xm\Vert_\ast.
\end{equation}
where $R_\ast(\Xm) = \gamma \Vert\Xm\Vert_\ast$ and  $\gamma$ is the regularization parameter.

The low-rank assumptions and the nuclear norm heuristics have been generalized to the multi-linear case, \emph{i.e.}, to tensors, using the unfolding operations \cite{liu2009tensor, signoretto2010nuclear}. Inspired by these works, we developed the tensor nuclear norm regularizer (TNN-1), which is introduced in Section \ref{sec:reg}-A. Additionally, we define a new tensor nuclear norm, where we exploit the T-SVD \cite{Kilmer2011641}. This new tensor nuclear norm and the regularizer (TNN-2) based on its definition are defined in Section \ref{sec:reg2}. The common property of TNN-1 and TNN-2 is the fact that they can be formulated as a matrix nuclear norm minimization problem, which we shall explain next. 

\subsection{Tensor Rank and the Generalized Tensor Nuclear Norm Regularizer (TNN-1)}\label{sec:reg1}

We start with the definition of Tucker decomposition, as the generalized tensor nuclear norm is related to it. The Tucker model \cite{tucker1966some,kolda2009tensor} is a multi-linear extension of SVD where a $K$-way tensor $\chi \in \mathbb{R}^{N_1\times N_2\times\hdots N_K}$ is decomposed into a core tensor $\mathcal{G} \in \mathbb{R}^{r_1\times r_2\times\hdots r_K}$, which controls the interactions between the modes and $K$ matrices, which multiply the core tensor in each mode:
\begin{equation*} \label{tucker}
\chi = \mathcal{G}\times_1 \mathbf{A}_1\times_2 \mathbf{A}_2 \hdots \mathbf{A}_{K-1}\times_K \mathbf{A}_K.
\end{equation*}
\begin{figure}
\label{fig:tucker}
\centering
\includegraphics[scale=0.4]{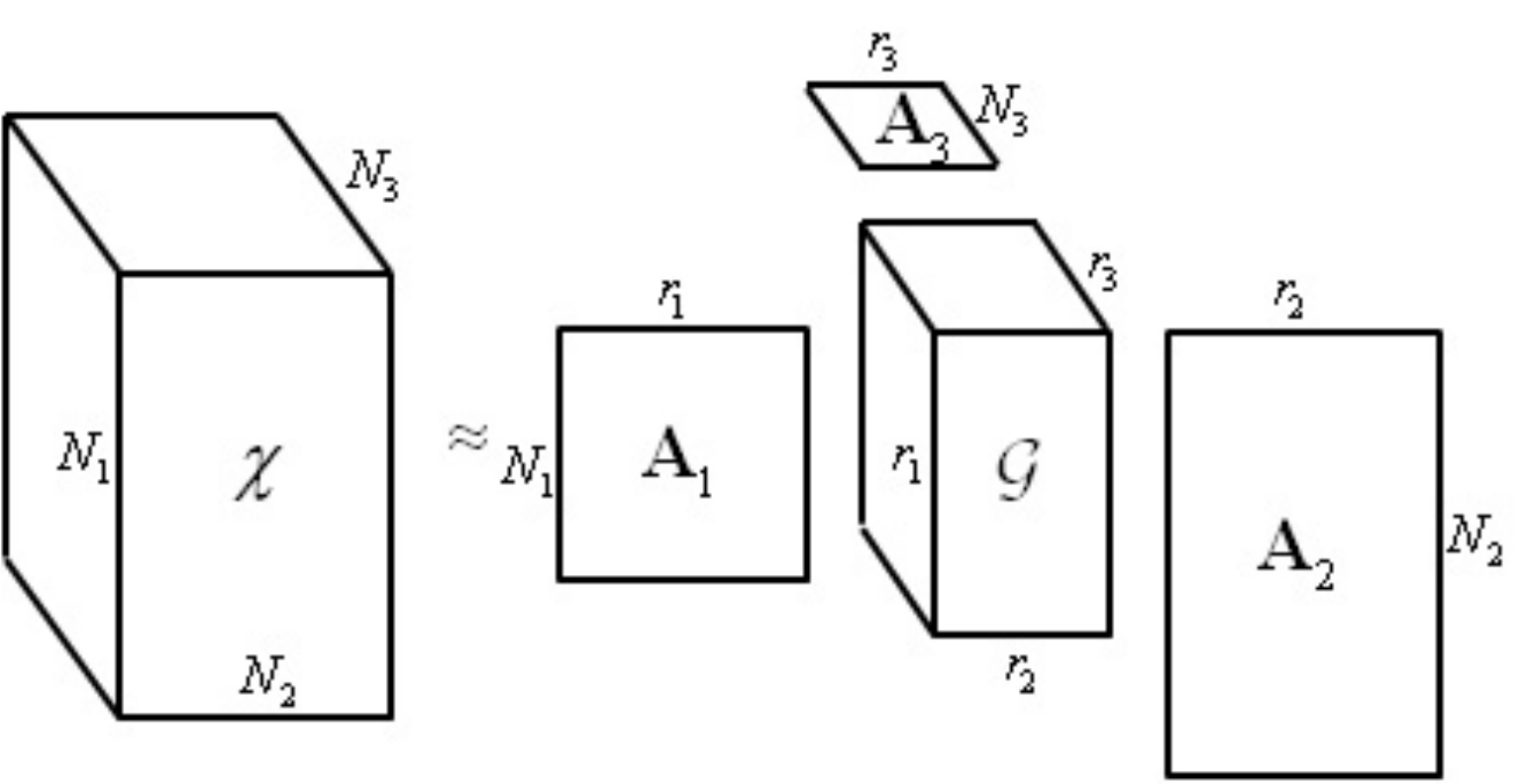}
\caption{Tucker decomposition of a 3-way tensor. The core tensor $\mathcal{G}$ controls the interactions between the modes and matrices that multiply the core tensor in each mode }
\end{figure}
Here, columns of $A_l\in\mathbb{R}^{N_l\times r_l}$ can be considered as left singular vectors of mode-$l$ unfolding and $r_l$ for $l=1,\hdots,K$ is referred to as the $n$-rank\footnote{The CP decomposition can be seen as a special case of Tucker where, the core tensor is super-diagonal. Therefore, $r_1=r_2=r_3=r$ where the tensor rank due to CP, $r$, needs to be known \emph{a priori} \cite{kolda2009tensor}.}. In the general Tucker model, the $\mathbf{A}_l$'s need not be orthogonal. The special case of the Tucker model where $\mathbf{A}_l$'s are orthogonal matrices is referred to as the Higher-Order-Singular-Value-Decomposition (HOSVD) \cite{de2000multilinear}. The Tucker model can also be written in terms of unfoldings. For the three-way case we have
\begin{equation}
\label{tucker2}
\begin{split}
\chi_{(1)} &= \mathbf{A}_1\mathcal{G}_{(1)}(\mathbf{A}_3 \otimes \mathbf{A}_2)^{\mathrm{T}}\\
\chi_{(2)} &= \mathbf{A}_2\mathcal{G}_{(2)}(\mathbf{A}_3 \otimes \mathbf{A}_1)^{\mathrm{T}}\\
\chi_{(3)} &= \mathbf{A}_3\mathcal{G}_{(3)}(\mathbf{A}_2 \otimes \mathbf{A}_1)^{\mathrm{T}},
\end{split}
\end{equation}
%
where $\otimes$ is the \emph{Kronecker} product. Fig. 4 illustrates the Tucker decomposition for the 3-way case. The equations given in (\ref{tucker2}) and definition of $n$-rank shows that if a tensor is low rank in its $l\tho$ mode (\emph{i.e.}, $r_l<\text{min}(N_l,\tiny{\prod_{k'\neq l}N_k'})$), its unfolding in the same mode is a low rank matrix. Due to this connection, the matrix nuclear norm has been generalized to tensors by utilizing the unfolding operation in each mode in order to estimate low-rank tensors via a convex minimization problem \cite{signoretto2010nuclear,liu2009tensor,gandy2011tensor}. The generalized tensor nuclear norm for a $K$-way tensor is given as \cite{signoretto2010nuclear, liu2009tensor}
\begin{equation}
\Vert \chi \Vert_\ast := \frac{1}{N}\sum_{k=1}^{N} \Vert \chi_{(k)} \Vert_\ast
\label{nuclear}
\end{equation}
We refer the reader to \cite{signoretto2010nuclear} for a thorough discussion about the relation of (\ref{nuclear}) to Tucker decomposition and Shatten 1-norm of matrices.  


\begin{figure*}[t!]
\centering

$\begin{array}{ccc}

\includegraphics[width=1.04in, trim = 9mm 3mm 7mm 0mm, clip=true]{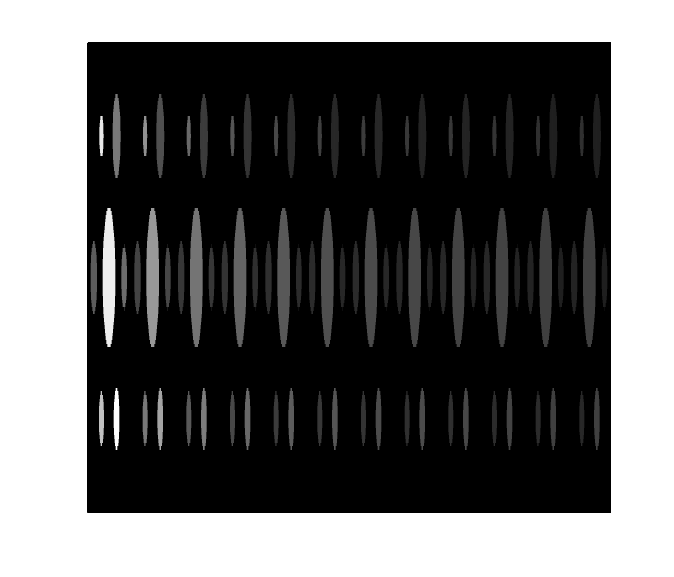}&
\includegraphics[width=0.98in, trim = 9mm 3mm 7mm 0mm, clip=true]{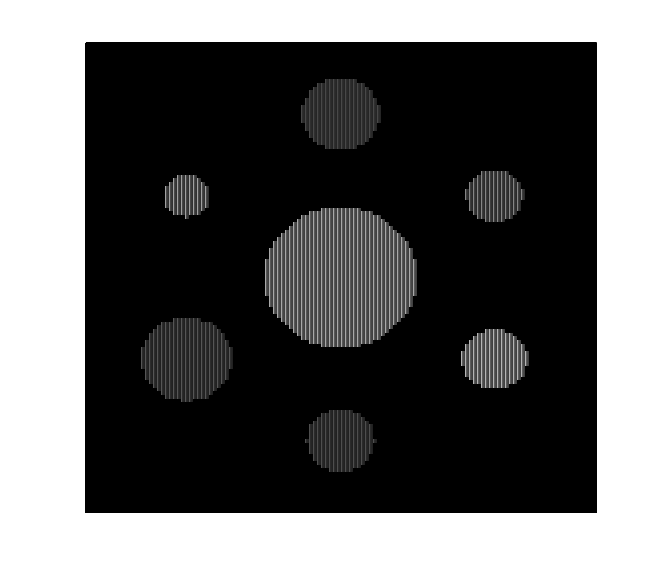}&
\includegraphics[width=0.98in, trim = 9mm 3mm 7mm 0mm, clip=true]{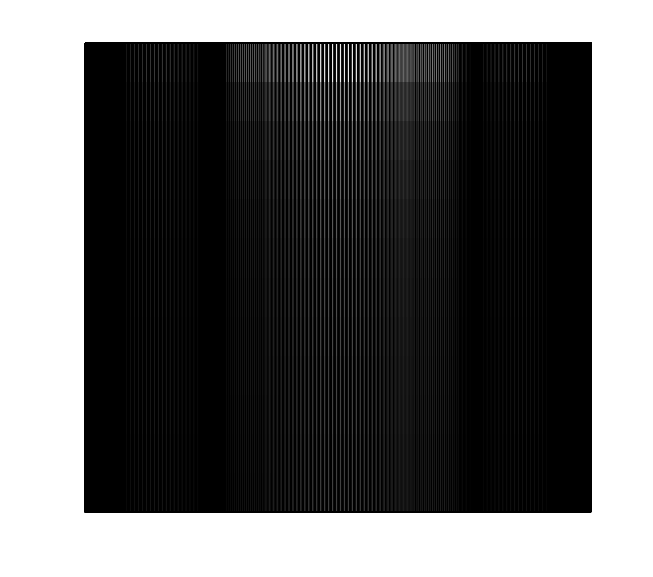}\\
\includegraphics[width=1.29in, trim = 0mm 2mm 2mm 2mm, clip=true]{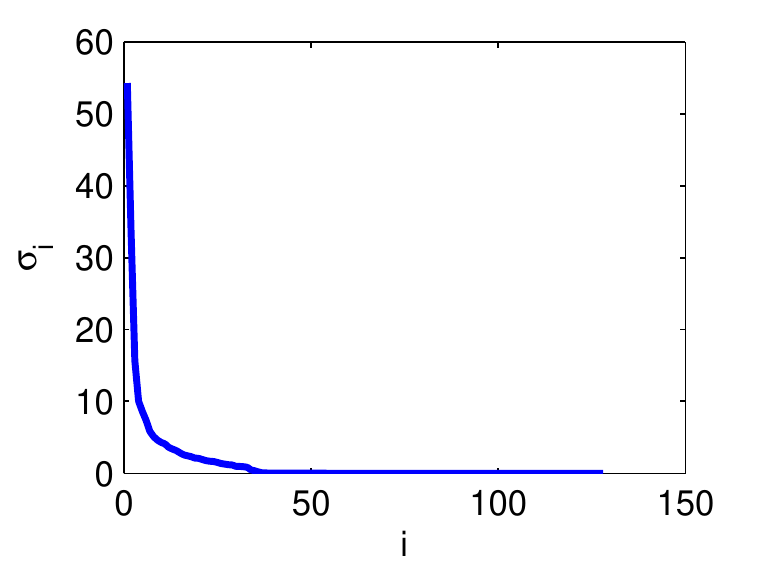}&
\includegraphics[width=1.29in, trim = 0mm 2mm 2mm 2mm, clip=true]{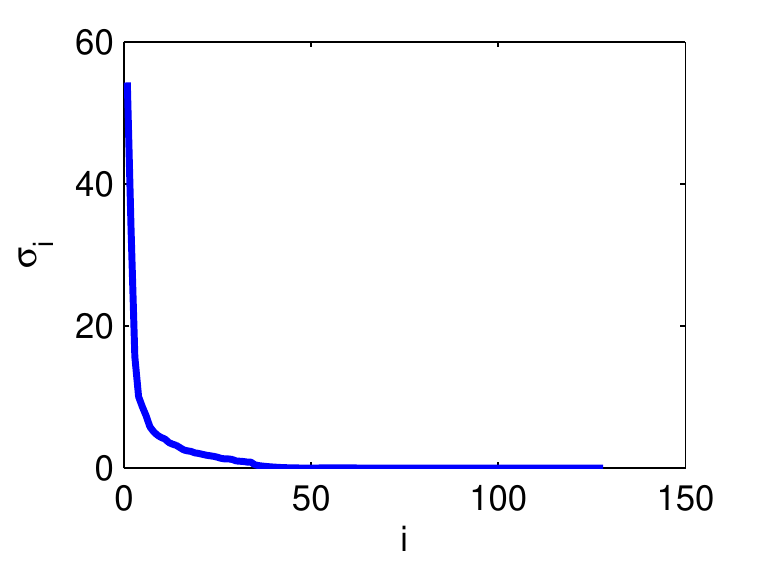}&
\includegraphics[width=1.29in, trim = 0mm 2mm 2mm 2mm, clip=true]{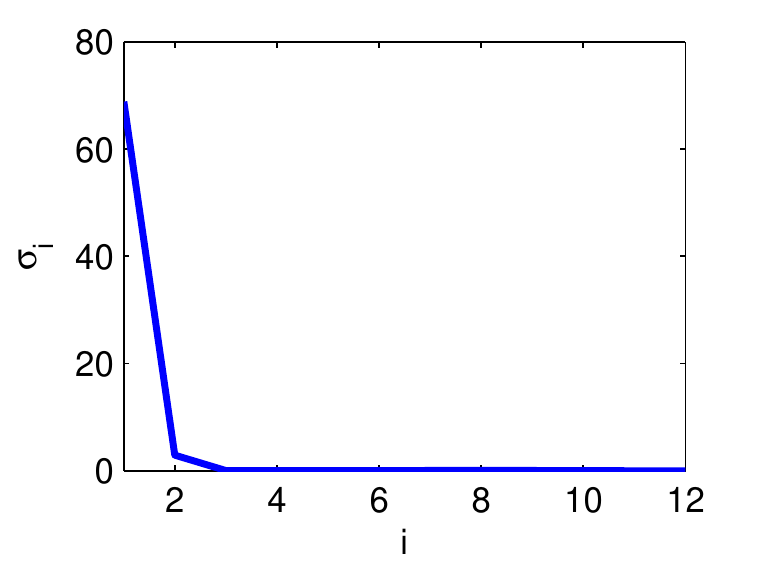}

\end{array}$
\label{fig:sing}
\caption{The unfoldings of $\chi$ and their singular values in log scale. First row: $\chi_{(1)},\chi_{(2)}$ and $\chi_{(3)}$. Second row: $\{\sigma(\chi_{(1)})\}_{i=1,\hdots,N_1}$, $\{\sigma(\chi_{(2)})\}_{i=1,\hdots,N_2}$ and $\{\sigma(\chi_{(3)})\}_{i=1,\hdots,N_3}$. }
\end{figure*}  

The low-rank tensor concept is quite relevant to the multi-energy CT problem. X-ray attenuation at neighboring energies are highly correlated. Therefore, for spectral CT, one expects the third unfolding, $\chi_{(3)}$ to be low-rank \cite{gao2011multi}. However, structural redundancies can also be exploited by enforcing low-rank structure on the other two unfoldings. To this end, we use the more general form of the tensor nuclear norm given in (\ref{nuclear}), which was proposed by Tomioka \emph{et al.} \cite{tomioka2010estimation} as a regularizer:
\begin{equation}
R_{\ast}(\chi)=\sum_{k=1}^{3} \gamma_k\Vert \chi_{(k)} \Vert_\ast,
\label{tnn}
\end{equation} 
where $\gamma_k$'s can be regarded as regularization parameters that tune the importance of each unfolding. A different parameter for each unfolding is assigned in order to make the regularizer more flexible \cite{tomioka2010estimation}, in a way that low-rank assumptions on any mode can be discarded if desired. For instance, see Section \ref{sec:results} for the examples where we set $\gamma_1$ and $\gamma_2$ to zero.  Fig. 5 demonstrates the unfoldings of the phantom with 12 energy levels given in Fig. \ref{fig:phantom} and their singular values. The rapid decay of the singular values provides an indication of the usefulness of (\ref{tnn}) as a regularizer for multi-energy CT reconstruction.

\subsection{A t-SVD Based Tensor Nuclear Norm and Regularization (TNN-2)} \label{sec:reg2}

In \cite{Kilmer2011641} it is shown that any tensor $\chi \in \mathbb{R}^{N_1 \times N_2 \times N_3}$ can be factored as 
\begin{equation*}
\label{tensorsvd}\chi=\mathcal{U}\ast \mathcal{S}\ast \mathcal{V}^{T},
\end{equation*}
where $\mathcal{U} \in \mathbb{R}^{N_1 \times N_1 \times N_3}$ and $\mathcal{V} \in \mathbb{R}^{N_2 \times N_2 \times N_3}$ are orthogonal, and $\mathcal{S}\in \mathbb{R}^{N_1 \times N_2 \times N_3}$ is made up with diagonal frontal faces. It is easy to show that, as is the case with the common matrix SVD, the t-SVD allows the tensor $\chi$ to be written as a finite sum of outer product of matrices \cite{kilmer2011third}:
\begin{equation}
\label{analogSVD}
\chi = \sum_{i=1}^{\mathrm{min}(N_1,N_2)} \mathcal{U}(:,i,:)\ast\mathcal{S}(i,i,:)\ast\mathcal{V}(:,i,:)^{\mathrm{T}},
\end{equation}
where $(:,i,:)$ and $(:,:,i)$ correspond to the $i\tho$ lateral and $i\tho$ frontal faces respectively, and $(i,i,:)$ is the $i\tho$ mode-3 fiber, similar to {\sc Matlab}'s indexing. Now, we have the following relationship in the Fourier domain \cite{Kilmer2011641}
\begin{equation}
\begin{split}
\label{tsvdfft} &\begin{bmatrix}\hat{\chi}^{(1)}& & \\ &\ddots & \\  & &\hat{\chi}^{(n)}\end{bmatrix}=
\begin{bmatrix} \hat{\bold U}^{(1)}& & \\  & \ddots & \\ & & \hat{\bold U}^{(N_3)}\end{bmatrix}\\
& \hspace{3mm}\cdot \begin{bmatrix}\hat{\bold S}^{(1)}& & \\  & \ddots & \\ & & \hat{\bold{S}}^{(N_3)}\end{bmatrix}
\cdot\begin{bmatrix}\hat{\bold V}^{(1)}& & \\  & \ddots & \\ & & \hat{\bold V}^{(N_3)}\end{bmatrix}^T,
\end{split}
\end{equation}
where the left-hand-side is the block diagonalized version of $\chi$ as given in (\ref{block_diag}) and  $\hat{{\bold A}}^{(n)} = \hat{\bold U}^{(n)}\hat{\bold{S}}^{(n)}(\hat{\bold V}^{(n)})^{\mathrm{T}}$ is the SVD of the block, $\hat{{\bold A}}^{(n)}$. In the light of (\ref{analogSVD}) and (\ref{tsvdfft}) we propose the following tensor nuclear norm:
\begin{equation*}
\Vert \chi \Vert_{\circledast}:=\sum_{i=1}^{\mathrm{min}(N_1,N_2)}\sum_{j=1}^{N_3}\hat{\mathcal{S}}(i,i,j).
\end{equation*}
Notice that we use the circled asterisk for this tensor nuclear norm definition. From (\ref{tsvdfft}) we have
\begin{equation}
\label{tnn2}
\begin{split}
\Vert \chi \Vert_{\circledast} & = \Vert ({\bold F}_{n}\otimes {\bold I}_{N_1})\cdot \bcircu
(\chi) \cdot({\bold F}_{N_3}^{\ast}\otimes \bold{I}_{N_2}) \Vert_\ast \\
& = \Vert  \bcircu
(\chi)  \Vert_\ast,
\end{split}
\end{equation}
where the first follows immediately from (\ref{block_diag}) and the second line is due to the unitary invariance of the matrix nuclear norm. A consequence of (\ref{tnn2}) is that $\Vert.\Vert_{\circledast}$ is a valid norm since \begin{itemize}
\item[\emph{i}.] For any tensor $\chi \in \mathbb{R}^{N_1 \times N_2 \times N_3}$, $\Vert\chi\Vert_{\circledast}=\Vert\bcircu(\chi)\Vert_{\ast}\geq 0$, and when $\chi=0$, by definition $\Vert\chi\Vert_{\circledast}=\Vert\bcircu(\chi)\Vert_{\ast} = 0$. \item[\emph{ii}.] Let $a\in\mathbb{R}$, then
\begin{equation*}
\begin{split}
\Vert a\chi\Vert_{\circledast}=\Vert\bcircu(a\chi)\Vert_{\ast}&=\Vert a(\bcircu(\chi))\Vert_{\ast}\\& =\vert a\vert \Vert\bcircu(\chi)\Vert_{\ast} =\vert a \vert \Vert\chi\Vert_{\ast}
\end{split}
\end{equation*}
\item[\emph{iii}.] Let $\chi_1$ and $\chi_2$ be two tensors.  
 \begin{equation*}
 \begin{split}\Vert\chi_1+\chi_2\Vert_{\circledast} & =\Vert\bcircu(\chi_2+\chi_2)\Vert_{\ast} \\
 & = \Vert\bcircu(\chi_1)+\bcircu(\chi_2)\Vert_{\ast}\\
 & \leq \Vert\bcircu(\chi_1)\Vert_{\ast}+\Vert\bcircu(\chi_2)\Vert_{\ast} \\
 & =\Vert\chi_1\Vert_{\circledast}+\Vert\chi_2\Vert_{\circledast}.
 \end{split}
 \end{equation*}
\end{itemize}

\begin{figure*}[t!]
\centering
$\begin{array}{cc}

\includegraphics[width=1.44in, trim = 0mm 3mm 7mm 0mm, clip=true]{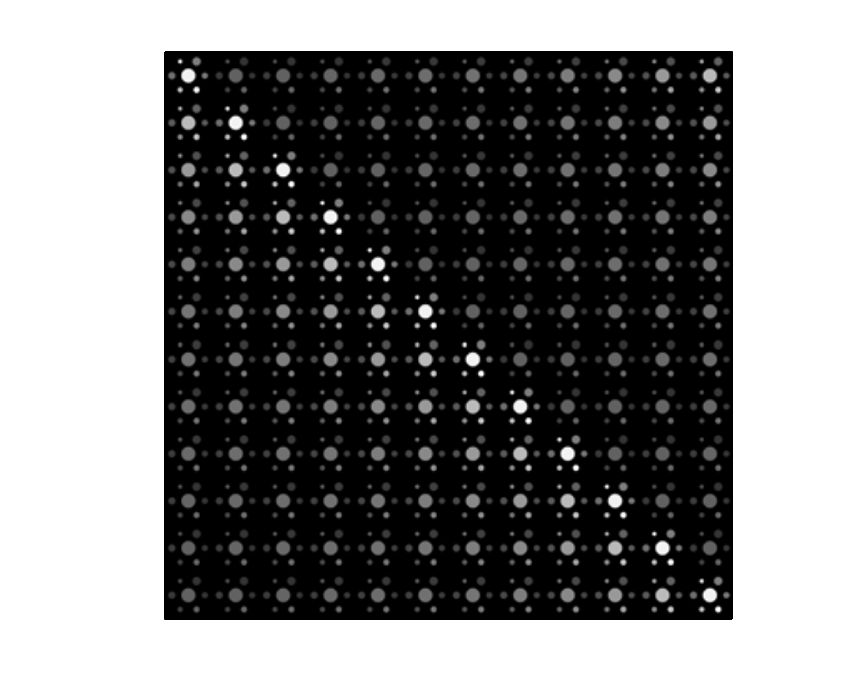}&
\includegraphics[width=1.28in, trim = 0mm 0mm 7mm 0mm, clip=true]{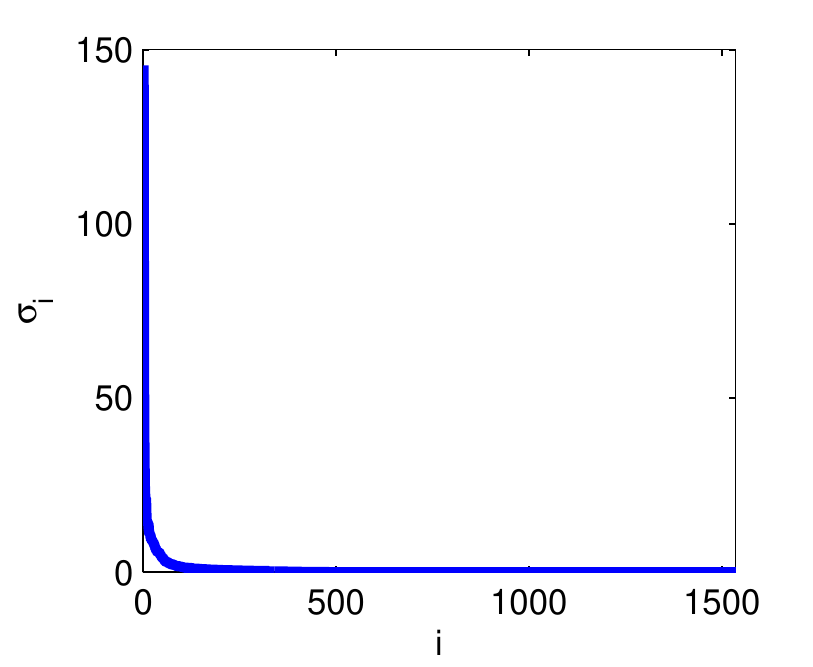}

\end{array}$
\label{fig:sing_circ}
\caption{The block circulant matrix $\bcircu(\chi)$ and its singular values.}
\end{figure*}  

Similar to the TNN-1 case, we use the new tensor nuclear norm given in (\ref{tnn2}) as a regularizer and form TNN-2 as:
\begin{equation*}
\label{tnn2r}
R_{\circledast}= \gamma \Vert \chi \Vert_{\circledast}  = \gamma \Vert\bcircu(\chi)\Vert_{\ast}
\end{equation*}
Fig. 6 demonstrates $\bcircu(\chi)$ of the phantom given in Fig. \ref{fig:phantom} and its singular values, which rapidly decay.

It has been shown that a truncated t-SVD representation provides an optimal representation in the same way that a truncated matrix SVD would give an optimal low rank approximation to the matrix in terms of the Frobenius norm \cite{Kilmer2011641}. Thus, our newly defined tensor nuclear norm, which is based on t-SVD, is more analogous to the matrix nuclear norm than generalized tensor nuclear norm given in (\ref{nuclear}) in this sense. Additionally, compared to TNN-1, TNN-2 has only one regularization parameter that needs to be determined.

\section{Inverse Problem Formulation} \label{sec:inversion}

The measurement model given in (\ref{measurement}) leads to the penalized weighted least squares (PWLS) formulation \cite{sauer1993local} which gives quadratic approximation to the Poisson log-likelihood function for $k^{\text{th}}$ energy bin as
\begin{equation*}
\label{data_misfit}
L_k(\mathbf{x}_k) = (\mathbf{A}\mathbf{x}_k - \mathbf{m}_k)^\mathrm{T}\Sigma_i^{-1}(\mathbf{A}\mathbf{x}_k - \mathbf{m}_k).
\end{equation*}
Here, $[\mathbf{m}_k]_j = \log (s_i/y_{k,j})$ and $\Sigma_k^{-1}$ is a diagonal weighting matrix with  $[\Sigma_k^{-1}]_{jj}= y_{k,j} $. Adding $L_k$'s and the regularization function $R(\chi)$ we obtain a convex objective function:
\begin{equation}
\label{optF1}
\begin{aligned}
&\underset{\chi}{\mbox{minimize}} & & \frac{1}{2} \sum_{k=1}^{N_3} L_k(\mathbf{x}_k)  + R(\chi),
\end{aligned}
\end{equation}
where $R(\chi)$ is a combination of $R_{\ast}(\chi)$ or $R_{\circledast}(\chi)$ with a total variation (TV) regularizer. We have considered two types of TV regularizers. The first one, which is denoted by $TV(\chi)$, is the weighted superposition of isotropic TV operator (2D TV) applied to the frontal slices of $\chi$ as
\begin{equation}
\label{TV}
\begin{split}
TV(\chi) & := \sum_{k=1}^{N_3}\alpha_k TV(\mathbf{x}_k) \\
& := \sum_{k=1}^{N_3}\alpha_k\sum_{i=1}^{N_1-1}\sum_{j=1}^{N_2-1} \vert (\nabla \Xm^{(k)})_{i,j} \vert
\end{split}
\end{equation}
where $ (\nabla \Xm^{(k)})_{i,j} = [(\nabla_x \Xm^{(k)})_{i,j} \;\; (\nabla_y \Xm^{(k)})_{i,j}]^{\mathrm{T}}$ with
$$(\nabla_x \Xm^{(k)})_{i,j} = \Xm^{(k)}_{i+1,j} - \Xm^{(k)}_{i,j}$$ and $$(\nabla_y \Xm^{(k)})_{i,j} = \Xm^{(k)}_{i,j+1} - \Xm^{(k)}_{i,j},$$Here, the $\alpha_k$'s are the regularization parameters. The second one, which is denoted by $TV_{3\text{D}}(\chi)$, is the three dimensional TV (3D TV) operator defined as
\begin{equation*}
\begin{split}
TV_{3\text{D}}(\chi) & := \alpha \sum_{k=1}^{N_1-1}\sum_{i=1}^{N_2-1}\sum_{j=1}^{N_3-1} \vert (\nabla \chi)_{i,j,k} \vert
\end{split}
\end{equation*}
where $ (\nabla \chi)_{i,j,k} $ is the obvious extension of the 2D case to 3D. In the remainder of the paper we refer the first approach as TV and the second as 3D-TV regularization. TV regularization favors images with sparse gradients, hence it works well for piecewise constant image reconstruction. However, due its stair-casing effect TV tends to be problematic for texture recovery \cite{elad2005simultaneous}. As we demonstrate empirically in Section \ref{sec:results}, combining the tensor nuclear norm regularizer with TV reduces the amount of TV needed for reasonable noise cancellation and helps with recovering the texture.

\section{Solution Algorithm via alternating direction method of multipliers (ADMM)}\label{inversion2}

ADMM is a combination of dual decomposition and augmented Lagrangian methods \cite{boyd2011distributed,bertsekas1999nonlinear}. Although it results in a four-fold increase in the number of variables in the minimization procedure (see Section VI-A), ADMM provides a simple splitting scheme that breaks down a cost function, which is hard minimize, into pieces that are more tractable and can be minimized efficiently. Splitting based methods have been used for several problems including iterative CT reconstruction \cite{xu2011statistical,ramani2012splitting,vandeghinste2011split}, image recovery \cite{afonso2010fast} and restoration \cite{figueiredo2010restoration} and tensor completion \cite{liu2009tensor,tomioka2010estimation}. We examine the solution algorithm according to the structure of $R(\chi)$.

\subsection{TNN-1 and TV Regularization}

The first case is when $R_\ast(\chi)$ is combined with $TV(\chi)$:
\begin{equation*}
\label{case1}
R_1(\chi) = R_\ast(\chi) + R_{TV}(\chi)
\end{equation*}

First, the optimization problem given in (\ref{optF1}) for $R_1(\chi)$ is reformulated as
\begin{equation}
\label{optF2}
\begin{aligned}
&\underset{\chi,\Zm_1,\Zm_2,\Zm_3}{\mbox{minimize}} & & \frac{1}{2} \sum_{k=1}^{N_3} L_k(\mathbf{x}_k) +  \sum_{l=1}^{3} \gamma_k\Vert \Zm_l \Vert_* + TV(\chi)\\
&\text{subject to} & & \Zm_l=\chi_{(l)}, \;\text{for}\; l=1,2,3.
\end{aligned}
\end{equation}
To solve (\ref{optF2}) we form the augmented Lagrangian as
\begin{equation}
\label{AL}
\begin{split}
L_\eta(\chi,\{\Zm_l\},\{\Ym_l\}) & = \frac{1}{2} \sum_{k=1}^{N_3} L_k(\mathbf{x}_k)   +  \sum_{l=1}^{3} \gamma_k\Vert \Zm_l \Vert_* \\
& + TV(\chi)+ \sum_{l=1}^{3} \left\langle\Ym_l,\chi_{(l)}-\Zm_l \right\rangle \\
& + \frac{\eta}{2}\sum_{k=l}^{3}\Vert \chi_{(l)}-\Zm_l \Vert^2_F,
\end{split}
\end{equation}
where $Y_l$'s are dual variables, $\eta>0$ is the penalty term and $\left\langle.\right\rangle$ is the inner product in the sense of Frobenius norm defined for $\mathbf{K}_1$ and $\mathbf{K}_2\in \mathbb{R}^{M\times N}$ as
\begin{equation*}
\left\langle\mathbf{K}_1,\mathbf{K}_2\right\rangle = \sum_{i=1}^{M}\sum_{j=1}^{N}[\mathbf{K}_1]_{ij}\cdot[\mathbf{K}_2]_{ij}
\end{equation*}

ADMM minimizes (\ref{AL}) for $\chi$ and $Z_k$'s in an alternating manner and then updates the dual variables:
\begin{equation}
\label{admm1}
\begin{array}{rcl}
\chi^{n+1}&:=& \underset{\chi}{\text{argmin}}\,L_{\eta}\left(\chi,\{\Zm_l\}^n,\{\Ym_l\}^n\right), \\
\Zm_l^{n+1}&:=& \underset{\Zm_l}{\text{argmin}}\,L_{\eta}\left(\chi^{n+1},\{\Zm_l\},\{\Ym_l\}^n\right), \;\text{for}\; l=1,2,3, \\
\Ym_l^{n+1}&:=& \Ym_l^n+\eta(\chi^{n+1}_{(l)}-Z^{n+1}_l), \;\text{for}\; l=1,2,3.
\end{array}
\end{equation}
Using the permutation matrix notation given in Section \ref{sec:preliminaries} we can write
%
%
\begin{equation*}
\begin{split}
\sum_{l=1}^{3} \left\langle\Ym_l,\chi_{(l)}-\Zm_l \right\rangle  & = \sum_{l=1}^{3} \left\langle\mathbf{P}_l^{\mathrm{T}}\mathbf{y}_l,\mathbf{x}-\mathbf{P}_l^{\mathrm{T}}\mathbf{z}_l \right\rangle\\
& = \sum_{k=1}^{N_3}\sum_{l=1}^{3} \left\langle\{\mathbf{P}_l^{\mathrm{T}}\mathbf{y}_l\}_k,\mathbf{x}_k-\{\mathbf{P}_l^{\mathrm{T}}\mathbf{z}_l\}_k \right\rangle
\end{split}
\end{equation*}
and
\begin{equation*}
\sum_{l=1}^{3}\Vert \chi_{(l)}-\Zm_l \Vert^2_F = \sum_{k=1}^{N_3}\sum_{l=1}^{3} \Vert \mathbf{x}_k-\{\mathbf{P}_l^{\mathrm{T}}\mathbf{z}_l\}_k \Vert^2,
\end{equation*}
where $\{.\}_k$ refers to the index set of corresponding energy (\emph{e.g.,} $1,\hdots,N_3$ for $k=1$). Hence, the $\chi$ update in (\ref{admm1}) can be decoupled and each $\mathbf{x}_k$ can be treated separately:
\begin{equation*}
\label{x_update}
\begin{split}
\mathbf{x}_k^{n+1}:= \underset{\mathbf{x}_k}{\text{argmin}}\,\,& L_k(\mathbf{x}_k) + \sum_{l=1}^{3} \left\langle\{\mathbf{P}_l^{\mathrm{T}}\mathbf{y}_l\}_k,\mathbf{x}_k-\{\mathbf{P}_l^{\mathrm{T}}\mathbf{z}_l\}_k \right\rangle\\
& + \sum_{l=1}^{3} \Vert \mathbf{x}_k-\{\mathbf{P}_l^{\mathrm{T}}\mathbf{z}_l\}_k \Vert^2 + \alpha_k TV(\mathbf{x}_k),
\end{split}
\end{equation*}
which is a total variation regularized quadratic problem that can be solved using various methods \cite{chambolle2004algorithm,vogel1998fast}. We used FISTA \cite{beck2009fast} in this work.

With a straightforward reformulation one finds that the $Z_l$ updates can be obtained via the proximity operator of the nuclear norm as
\begin{equation*}
\label{Z_update}
\begin{array}{lll}
\Zm_l^{n+1}&:=& \underset{\Zm_l}{\text{argmin}}\,\, \Vert \Zm_l \Vert_* + \frac{\eta}{2\gamma_l} \left \Vert \frac{\Ym_l^n}{\eta}+\chi^{n+1}_{(l)} \right\Vert_*\\
&:=&\mathrm{prox}_{\frac{\gamma_l}{\eta}\Vert.\Vert_*}\left( \frac{\Ym_l^n}{\eta}+\chi^{n+1}_{(l)} \right)
\end{array}
\end{equation*}
which has an analytical solution via the \emph{ singular value shrinkage} operator \cite{cai2008singular}. Specifically
\begin{equation}
\label{proxy_def}
\mathrm{prox}_{\frac{\gamma_l}{\eta}\Vert.\Vert_*}(\mathbf{Z})=\mathbf{U}S_{\frac{\gamma_l}{\eta}}(\boldsymbol{\Sigma})V^{\textbf{T}},
\end{equation}
where $\mathbf{Z}=\mathbf{U}\boldsymbol{\Sigma}V^{\textbf{T}}$ is the singular value decomposition of $\mathbf{Z}$ and $S\rho(\boldsymbol{\Sigma})=\text{diag}(\{(\sigma_i-\rho)_+\})$ is the shrinkage operator with $t_+=\text{max}(t,0)$ applied to the singular values.


%
%
%
%
\subsection{TNN-2 and TV Regularization}

Replacing TNN-1 with TNN-2 gives
\begin{equation*}
\label{case2}
R_2(\chi) = R_{\circledast}(\chi) + R_{TV}(\chi)
\end{equation*}
and
\begin{equation}
\label{optF3}
\begin{aligned}
&\underset{\chi,\Zm}{\mbox{minimize}} & & \frac{1}{2} \sum_{k=1}^{N_3} L_k(\mathbf{x}_k) +  \gamma\Vert \Zm \Vert_* + TV(\chi)\\
&\text{subject to} & & \Zm =\bcircu(\chi).
\end{aligned}
\end{equation}
needs to be solved. The augmented Lagrangian for (\ref{optF3}) is given as
\begin{equation*}
\label{AL2}
\begin{split}
L_\eta(\chi,\Zm,\Ym) & = \frac{1}{2} \sum_{k=1}^{N_3} L_k(\mathbf{x}_k)   +  \gamma\Vert \Zm \Vert_{*} \\
& + TV(\chi)+ \left\langle\Ym,\bcircu(\chi)-\Zm \right\rangle \\
& + \frac{\eta}{2}\Vert \bcircu(\chi)-\Zm \Vert^2_F,
\end{split}
\end{equation*}
In order to update $\bfx_k$'s separately as in the TNN-1 case, using the definition of $\bcircu(.)$ operation given in (\ref{bcircu_matrix}), we can write
\begin{equation*}
\left\langle\Ym,\bcircu(\chi)-\Zm \right\rangle = 
\sum_{k=1}^{N_3}\left\langle \{\mathbf{y}\}_k,\bfx_k-\{\mathbf{z}\}_k 
\right\rangle
\end{equation*}
and
\begin{equation*}
 \Vert \bcircu(\chi)-\Zm \Vert^2_F = 
\sum_{k=1}^{N_3}\Vert \bfx_k-\{\mathbf{z}\}_k \Vert^2_F
\end{equation*}
where $\{.\}_k$ refers to the index set of $k\tho$ energy (\emph{e.g.,} for $k=1$ we have $\big\{[1,\hdots,N_1 N_2],[(N_3+1) N_1 N2+1,\hdots, (N_3+2) N_1 N2+1],\hdots,[(N_3^2-1)N_1 N_2 +1,\hdots,N_3^2N_1 N_2] \big\}$). Given this notation, we note that the solution algorithm of (\ref{optF3}) is identical to the  TNN-1 case described in Section VI-A.

\begin{figure*}[h!t!]
\centering
$\begin{array}{cc}
\includegraphics[width=1.39in, trim = 18mm 5mm 15mm 0mm, clip=true]{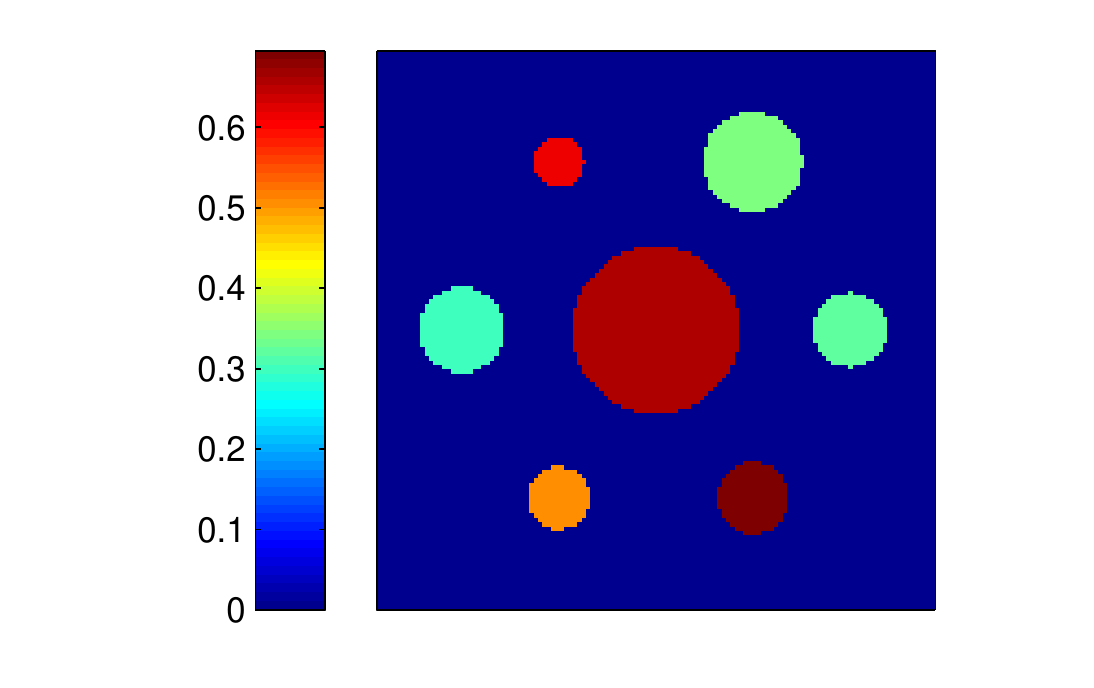}  &
\includegraphics[width=1.39in, trim = 18mm 5mm 15mm 0mm, clip=true]{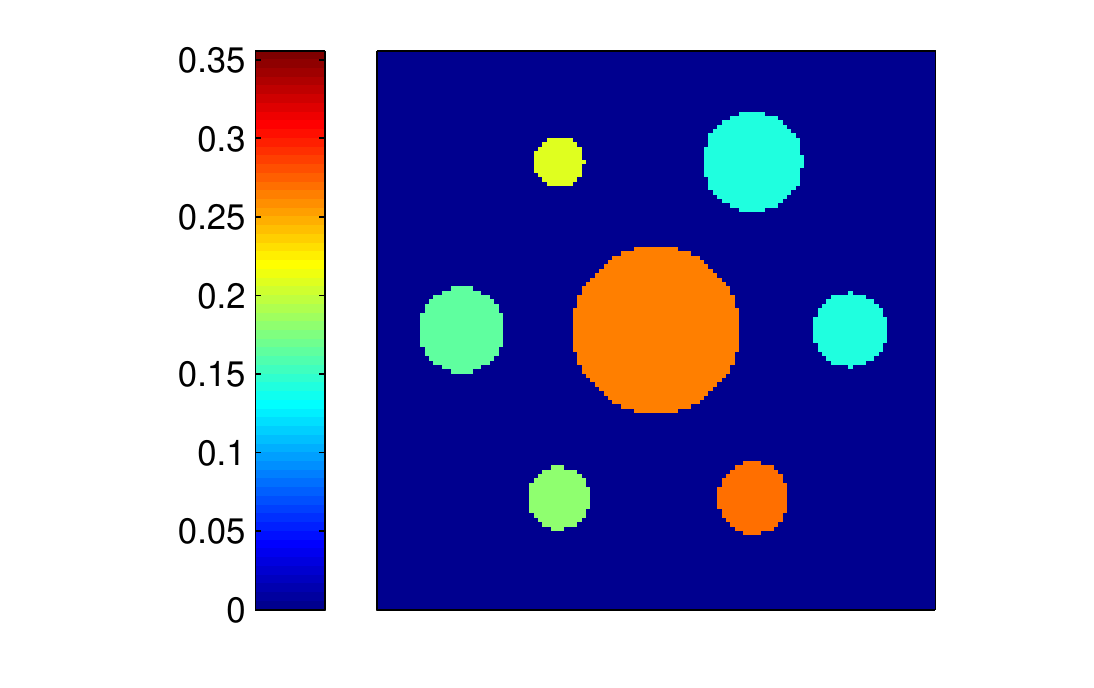}\vspace{-0.2cm} \\
\text{\small   Ground truth 25 keV}&\text{\small Ground  truth 85 keV}
\end{array}$
\caption{ Ground truth for Phantom-1. Left: 25 keV. Right: 85 keV}
\label{fig:truth1}
\end{figure*}

In Section \ref{sec:results} we show examples where $R(\chi) = R_\ast(\chi)$, $R(\chi) = R_{\circledast}(\chi)$, $R(\chi) =  R_{TV}(\chi)$ and $R(\chi) =  R_{TV_{3\text{D}}}(\chi)$. Solution to first two cases are straightforward variations where the latter case corresponds to reconstructing images for each energy independently using TV regularization. The last case results in a 3D linear inverse problem with TV regularization, for which we have used the UPN algorithm described in \cite{jensen2012implementation}.

\begin{figure*}[h!t!]
\centering
$\begin{array}{cccc}
\includegraphics[width=1.39in, trim = 18mm 5mm 15mm 0mm, clip=true]{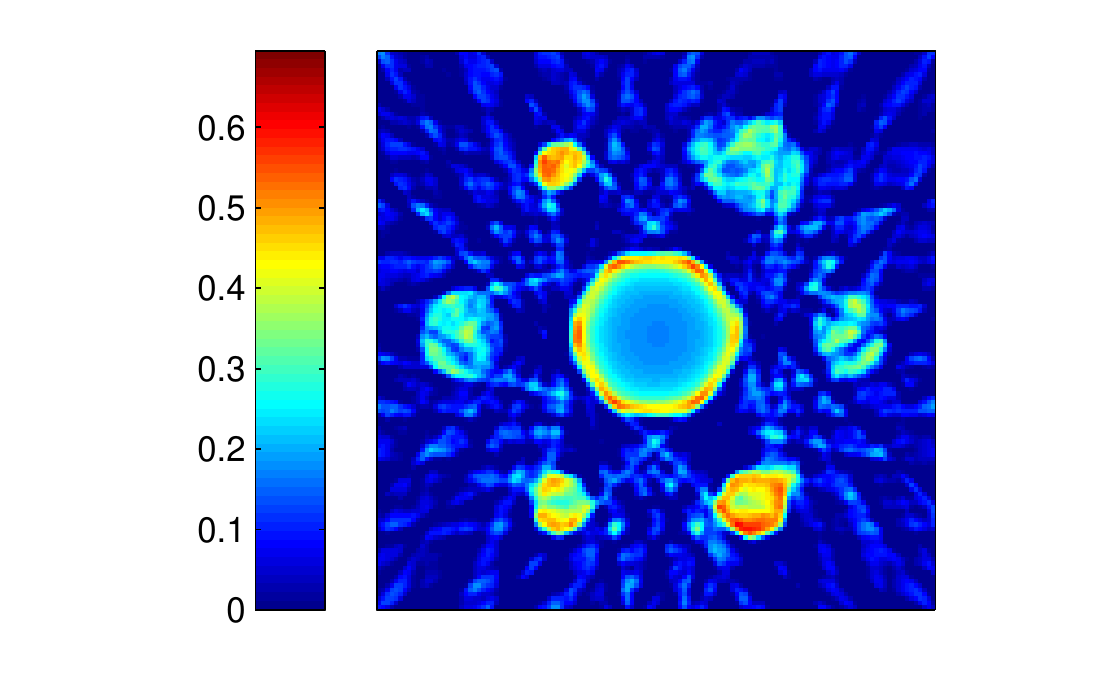}  &
\includegraphics[width=1in, trim = 31mm 6mm 25mm 0mm, clip=true]{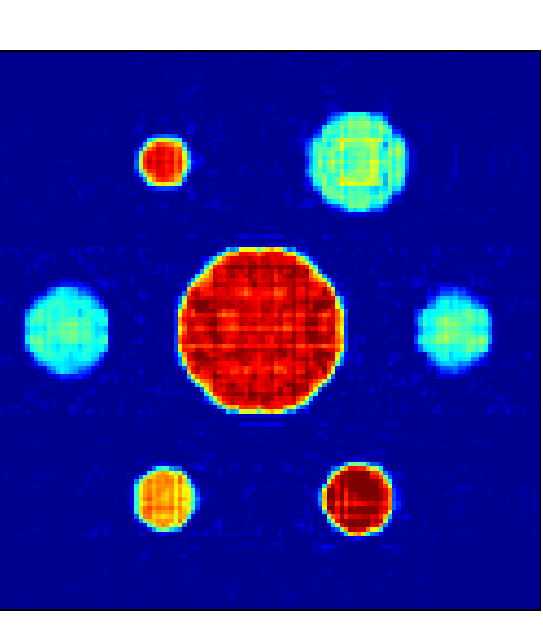}   &
\includegraphics[width=1in, trim = 31mm 6mm 25mm 0mm, clip=true]{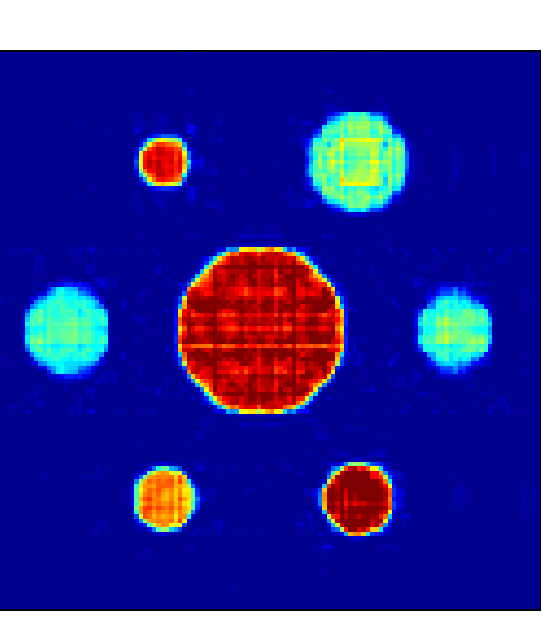} & \vspace{-0.2cm}  \\
\text{\small    FBP}&\text{\small TNN-1}&\text{\small TNN-2} \\

\includegraphics[width=1.39in, trim = 18mm 5mm 15mm 0mm, clip=true]{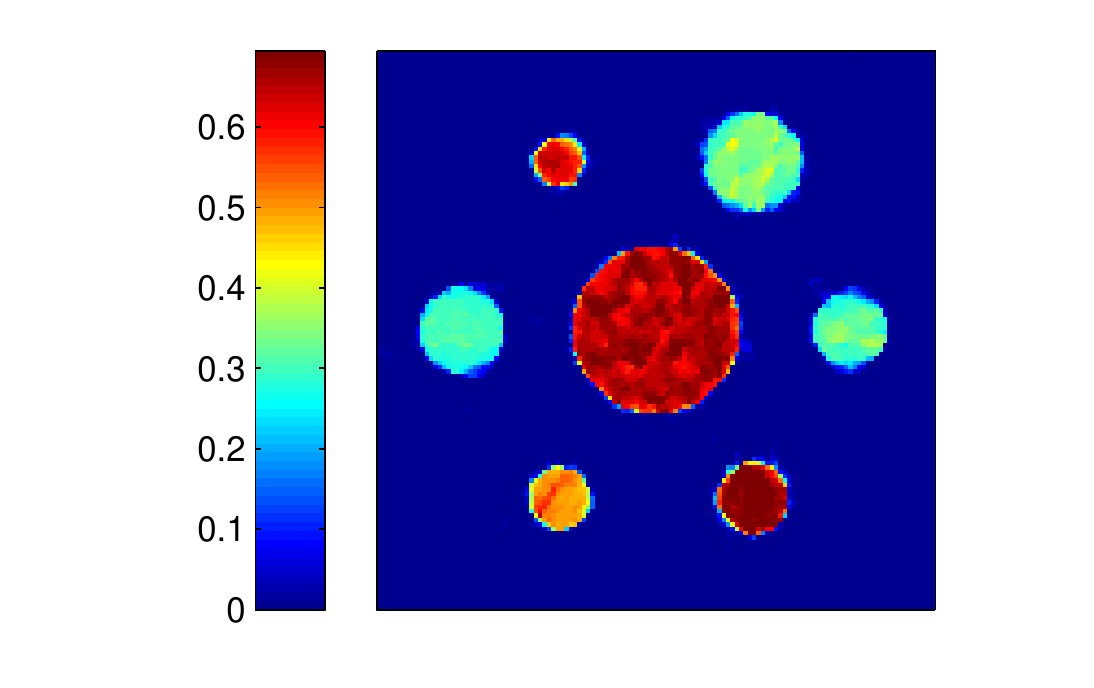}   &
\includegraphics[width=1.in, trim = 38mm 6mm 32mm 3mm, clip=true]{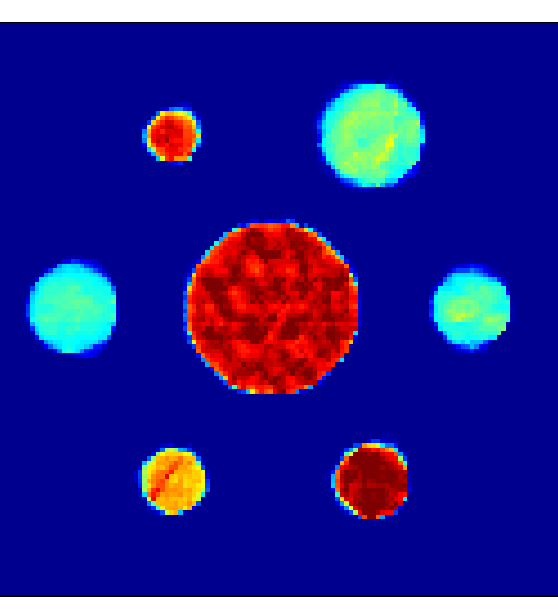}   &
\includegraphics[width=1in, trim = 31mm 6mm 25mm 0mm, clip=true]{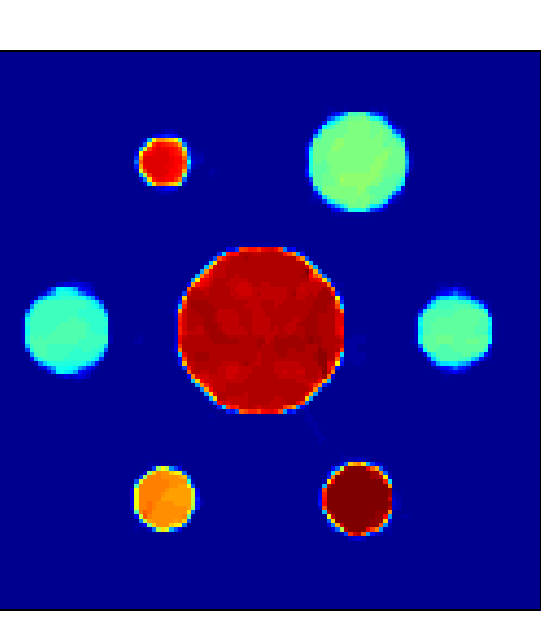}  &
\includegraphics[width=1in, trim = 31mm 6mm 25mm 0mm, clip=true]{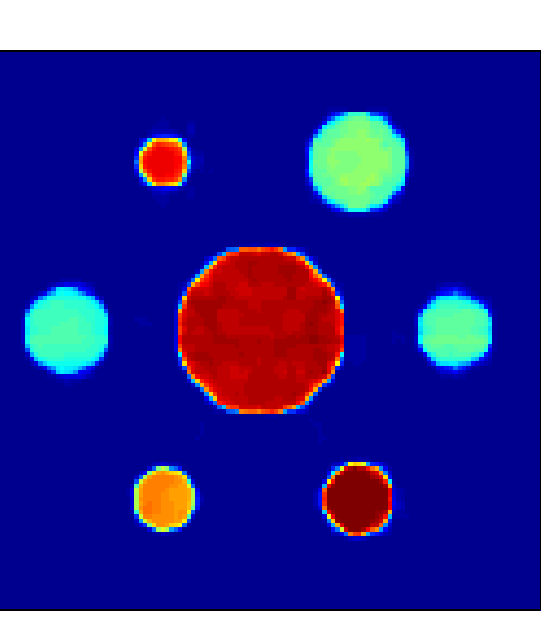}\vspace{-0.2cm} \\
\text{T\small V}&\text{\small 3D-TV}&\text{\small TNN-1+TV}&\text{\small TNN-2+TV}
\end{array}$
\caption{Phantom-1: Reconstructions results for 25 keV.}
\label{fig:results_original_25}
\end{figure*}

\begin{figure*}[h!t!]
\centering
$\begin{array}{cccc}
\includegraphics[width=1.39in, trim = 18mm 5mm 15mm 0mm, clip=true]{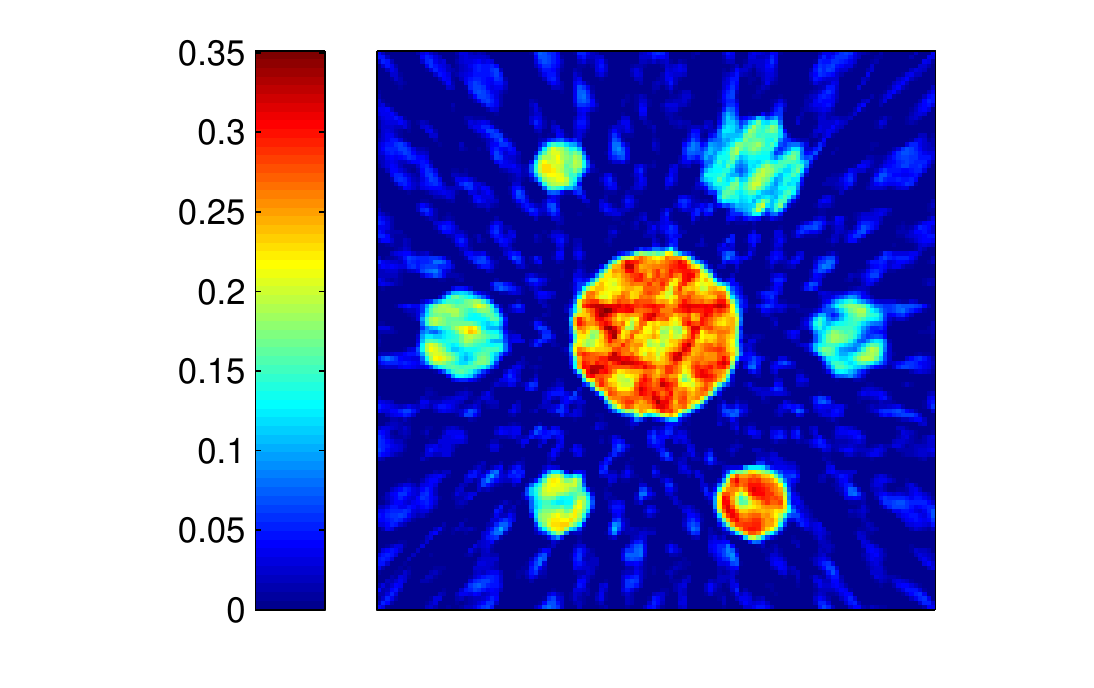}  &
\includegraphics[width=1in, trim = 31mm 6mm 25mm 0mm, clip=true]{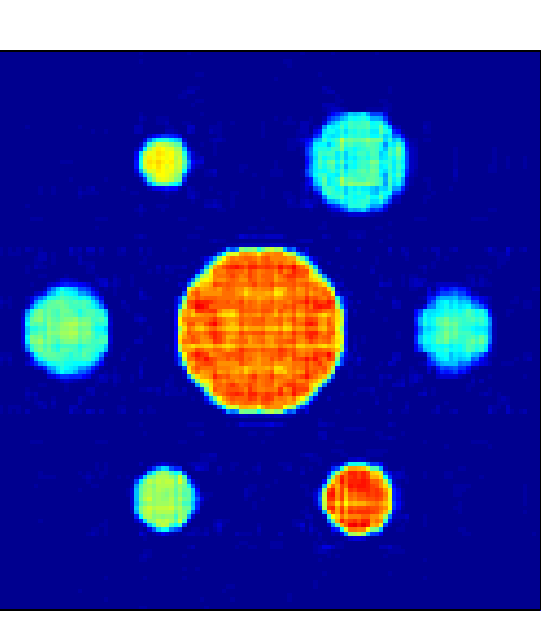}   &
\includegraphics[width=1in, trim = 31mm 6mm 25mm 0mm, clip=true]{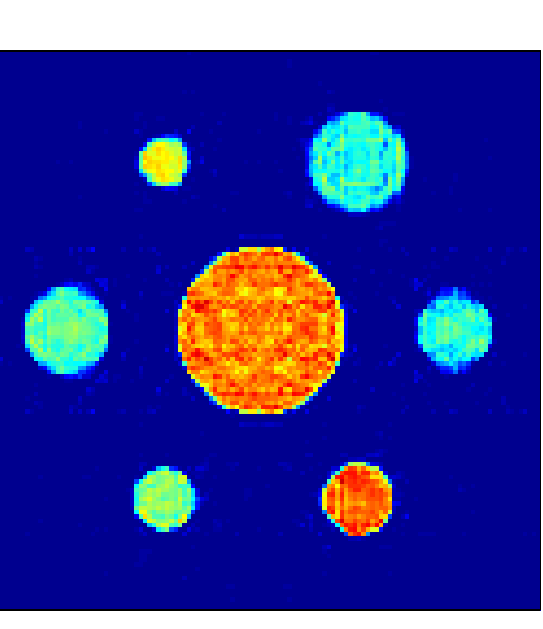}&\vspace{-0.2cm}   \\
\text{\small    FBP}&\text{\small TNN-1}&\text{\small TNN-2} \\

\includegraphics[width=1.39in, trim = 18mm 5mm 15mm 0mm, clip=true]{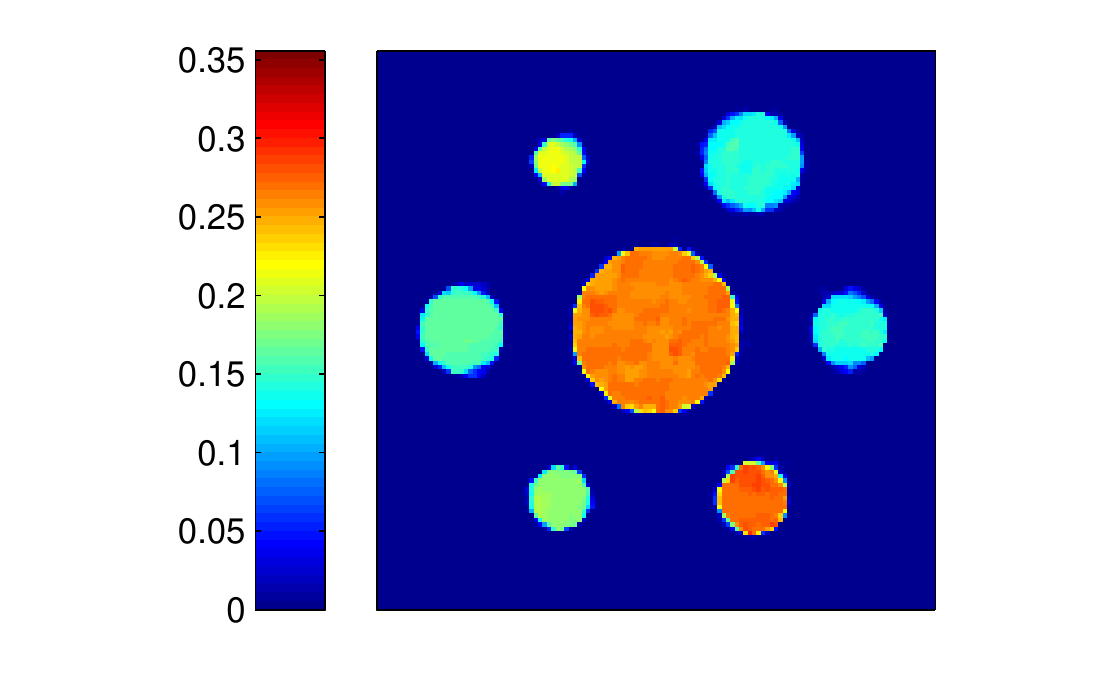}   &
\includegraphics[width=1.in, trim = 38mm 6mm 32mm 3mm, clip=true]{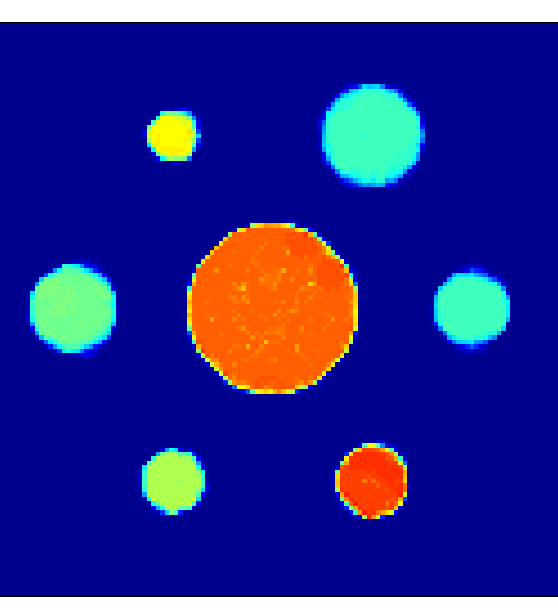}   &
\includegraphics[width=1in, trim = 31mm 5mm 25mm 0mm, clip=true]{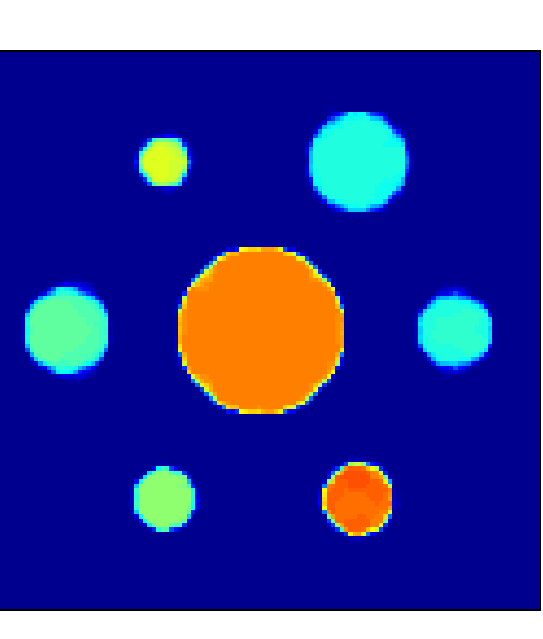}  &
\includegraphics[width=1in, trim = 31mm 5mm 25mm 0mm, clip=true]{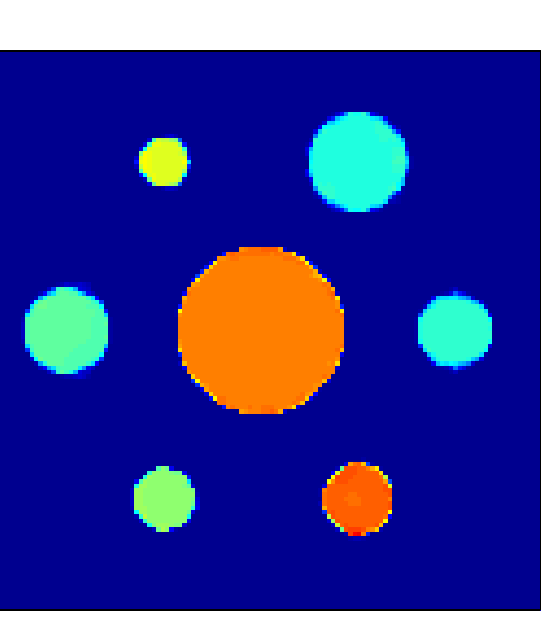}\vspace{-0.2cm} \\
\text{\small TV}&\text{\small 3D-TV}&\text{\small TNN-1+TV}&\text{\small TNN-2+TV}
\end{array}$
\caption{Phantom-1: Reconstructions results for 85 keV.}
\label{fig:results_original_85}
\end{figure*}

\section{Reconstruction Examples} \label{sec:results}

We compared the following methods in our simulations:
\begin{itemize}
\item[1.]Filtered back projection (FBP) \cite{slaney1988principles} algorithm applied to each energy bin separately. A RamLak filter multiplied with a Hamming window was used in the FBP inversion \cite{slaney1988principles}.
\item[2.]Only TV regularization at each energy bin separately (i.e., $\lambda_{\text{nuc}}=0$ in (\ref{optF1})).
\item[2.]3D-TV regularization.
\item[4.]Only TNN-1 regularization (i.e., $\alpha_k=0$ for $k=1,\hdots,N_3$ in (\ref{TV}).
\item[5.]Only TNN-2 regularization.
\item[6.]TNN-1 and TV regularization.
\item[7.]TNN-2 and TV regularization.
\item[9.]TNN-1 with $\gamma_1$ and $\gamma_2$ are set to 0.
\end{itemize}

Quantitative accuracy of the reconstructions were determined by relative $\ell_2$ error, $E_{\ell_2}$ which is given as
$$
E_{\ell_2} = \frac{\Vert \hat{\mathbf{x}}_k-\mathbf{x}^*_{k} \Vert^2_2}{\Vert \mathbf{x}^*_{k} \Vert^2_2  }.
$$
Here, $\hat{\mathbf{x}}_k$ and $\mathbf{x}^*_{k}$ are the reconstruction and the true images at the k$^{\text{th}}$ energy respectively, and $\Vert\cdot\Vert$ is the Euclidean norm.

\begin{table}
\caption{Error performance with respect to FBP: Elapsed times at particular iterations when FBP is outperformed for each method. The 3D-TV method uses an optimized C code that is called from Matlab \cite{jensen2012implementation}. All other methods use a non-optimized Matlab code.}

\centering
\begin{tabular}{lccc}
\hline\hline
Method & $E_{\ell_2} (25 \text{keV})$ & Iteration Number & Comp. time (sec) \\
\hline
FBP        & 0.2010  & -  &4     \\
TNN-1      & 0.1401  & 2  &1.25  \\
TNN-2      & 0.1837  & 3  &8.2     \\
TV         & 0.1919  & 17 &24.66\\
3D-TV	   & 0.1376  & 1  &0.15\\
TV+TNN-1   & 0.1093  & 1  &165\\
TV+TNN-2   & 0.1526  & 2  &342\\
\hline
\end{tabular}
\label{table:compare_fbp}
\end{table}

\begin{figure}
\centering
\includegraphics[width=2.4in, trim = 0mm 0mm 0mm 0mm, clip=true]{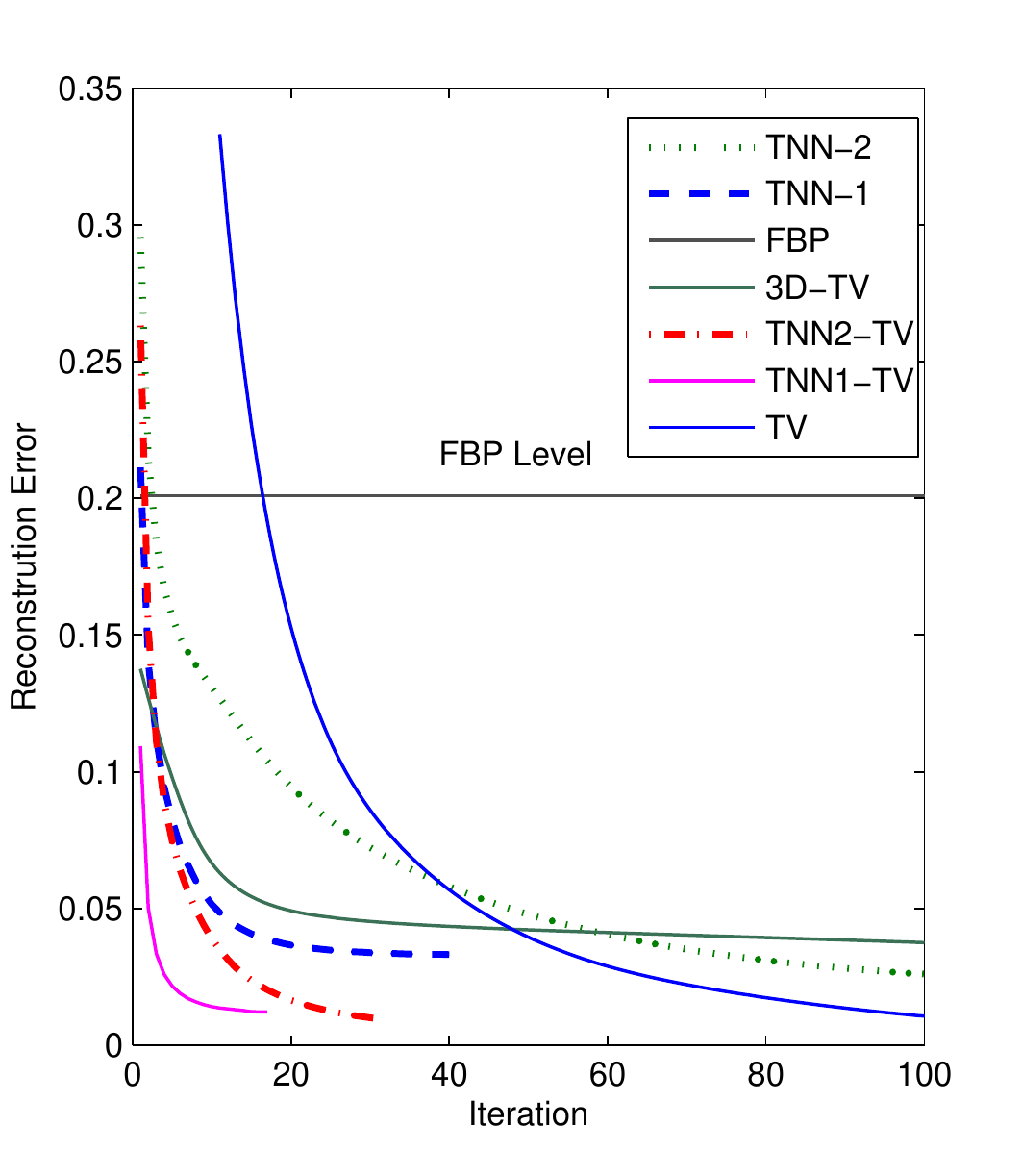}
\caption{Relative $E_{\ell_2}$ error versus iteration number for 85keV results from Phantom-1. The horizontal line represents the error level achieved by FBP. Note that all methods are implemented in Matlab except for 3D-TV which is written in C.}
\label{fig:error}
\end{figure}

\begin{figure*}[h!t!]
\centering
$\begin{array}{cc}
\includegraphics[width=1.39in, trim = 18mm 5mm 15mm 0mm, clip=true]{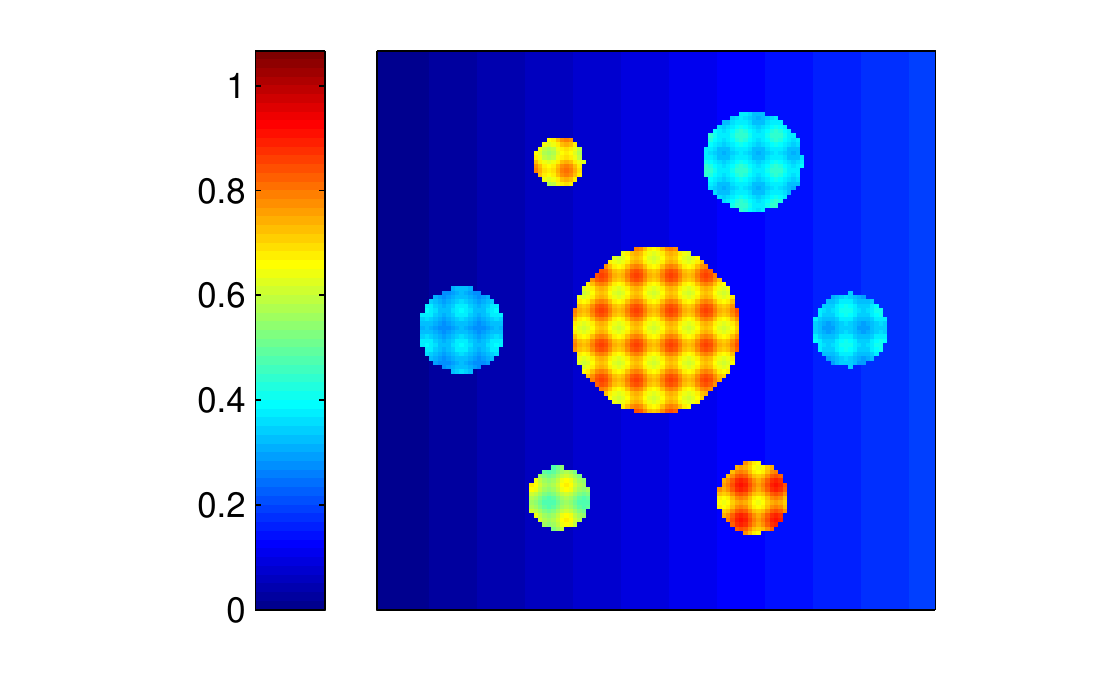}  &
\includegraphics[width=1.39in, trim = 18mm 5mm 15mm 0mm, clip=true]{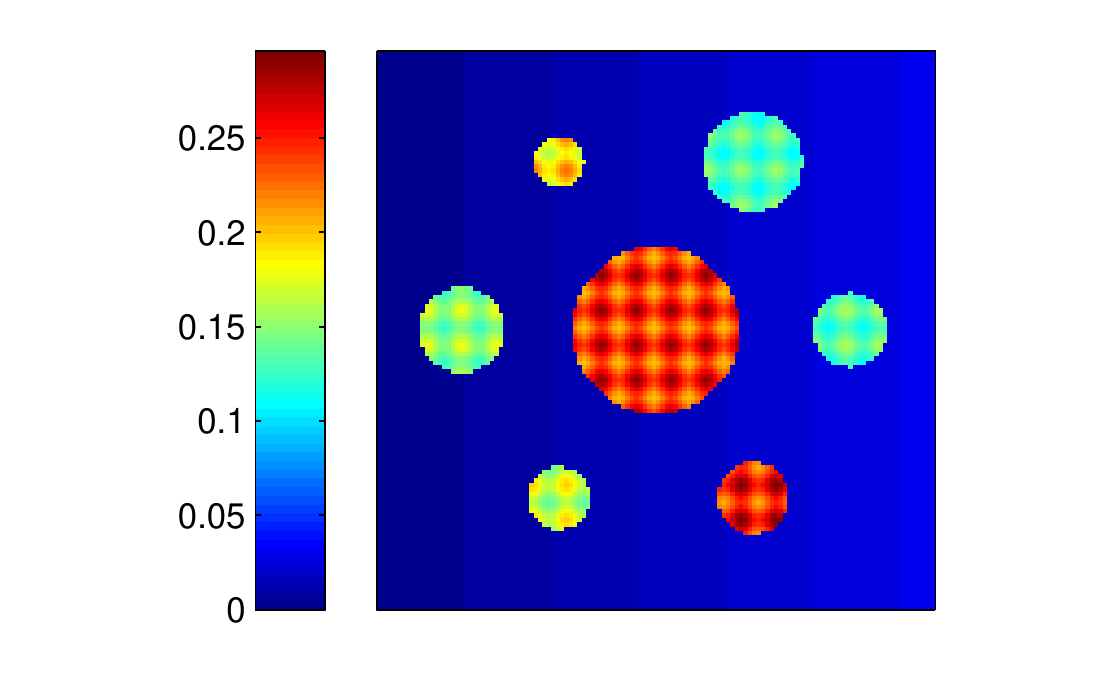}\vspace{-0.2cm} \\
\text{\small   Ground truth 25 keV}&\text{\small Ground  truth 85 keV}
\end{array}$
\caption{ Ground truth for Phantom-2. Left: 25 keV. Right: 85 keV}
\label{fig:truth2}
\end{figure*}

We simulated multi-spectral data for 12 energies between 25 and 85 keV for 16 uniformly distributed angles between 0 and 180 degrees. We chose 25-85 keV range as it covers the lower portion of the X-ray source spectra of 20-140 keV used in CT \cite{beutel2000handbook}, where materials are better differentiated (see Fig. \ref{fig:phantom}). We assumed the X-ray spectra is uniform with $10^6$ photons at each energy. We have used 2 different phantoms in our experiments. In addition  to the piecewise constant phantom (128x128 pixels) shown in Fig. \ref{fig:phantom} and \ref{fig:truth1}, which we call Phantom-1, we generated another phantom with isotropic texture on the objects and with a small linear variation to the background for more realistic experiments. This second phantom is called Phantom-2 and is shown in see Fig. \ref{fig:truth2}. To explore the performance of the approach using a more  realistic phantom we employed a DICOM image obtained from a CT scan of a duffel bag and artificially assigned attenuation values that are in the same range as Phantom-1 (see Fig. \ref{fig:truth3}). For all cases considered here, simulations are performed in {\sc MATLAB} \cite{guide1998mathworks} except for the 3D-TV implementation we have used the code available at http://www2.imm.dtu.dk/$\sim$pcha/TVReg/, which is written in C. We have used a 8 core Intel CPU with 16 gigabytes of memory. Note that the code we have used for these experiments is written in Matlab and is in no sense optimized for efficiencies which could be obtained using a lower level language and exploiting parallel architectures. Indeed, the bulk of the computation time comes from the projection and back-projection operations (\emph{i.e., $\textbf{A},\textbf{A}^{\mathrm{T}}$}) and from the computation singular values, both of which can be performed efficiently using fast and parallel algorithms \cite{de2004distance,andrecut2009parallel} in order
to achieve a real-time reconstruction algorithm, which is crucial for the baggage inspection application. In order to implement the singular value soft thresholding operation given in (\ref{proxy_def}), we used the PCA (principle component analysis) \cite{golub1996matrix} algorithm given in \cite{woolfe2008fast}, which returns a rank $k$ approximation of a $n\times m$ matrix in $\mathcal{O}(mn \log k + l^2(m + n))$ operations, where $l$ is an integer bigger than and close to $k$. Hence, we avoided the explicit calculation of SVD at each iteration, which can be calculated in $\mathcal{O}(knm)$ operations using a standard $QR$ decomposition based algorithm \cite{golub1996matrix}.The linear attenuation values for the materials in Phantom-1 are taken from XCOM: Photon cross sections database \cite{berger1998xcom}.

In all examples we set $\eta,\gamma_1,\gamma_2,\gamma_3=0.4$ and $\gamma=0.1$. We let $\alpha_i$'s reduce from 0.05 to 0.03 in a quadratic manner from $k=1,\hdots,N_3$ as low energy images need stronger regularization due to the higher level of Poisson noise. The 3D-TV regularization parameter $\alpha$ was set to 0.1. We tuned the regularization parameters manually and used the same set of parameters for both phantoms, as they gave the best error performance. We emphasize that systematic selection of regularization parameters is an important problem, which continues to be an active area of research, especially for non-quadratic regularization techniques such as TV and nuclear norm regularization \cite{ahn2008analysis,ramani2012regularization,candes2012unbiased}.

\begin{figure*}[h!t!]
\centering
$\begin{array}{cccc}
\includegraphics[width=1.39in, trim = 26mm 6mm 22mm 3mm, clip=true]{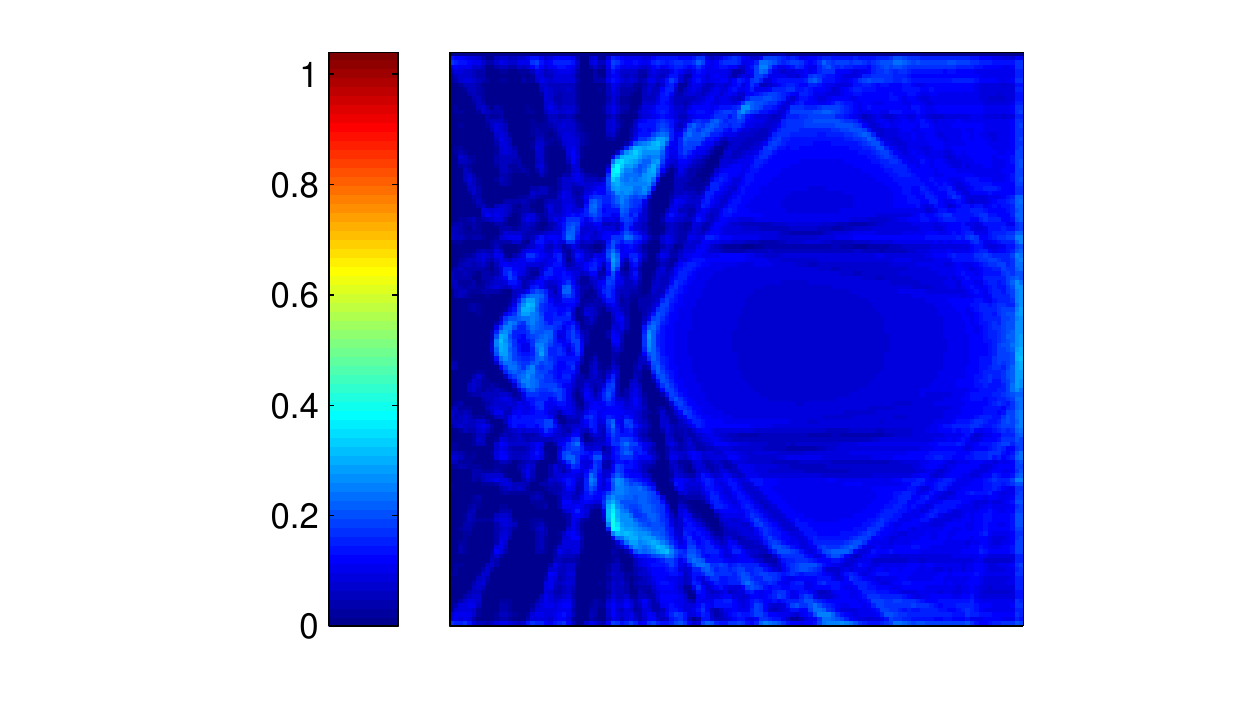}  &
\includegraphics[width=1in, trim = 31mm 5mm 25mm 0mm, clip=true]{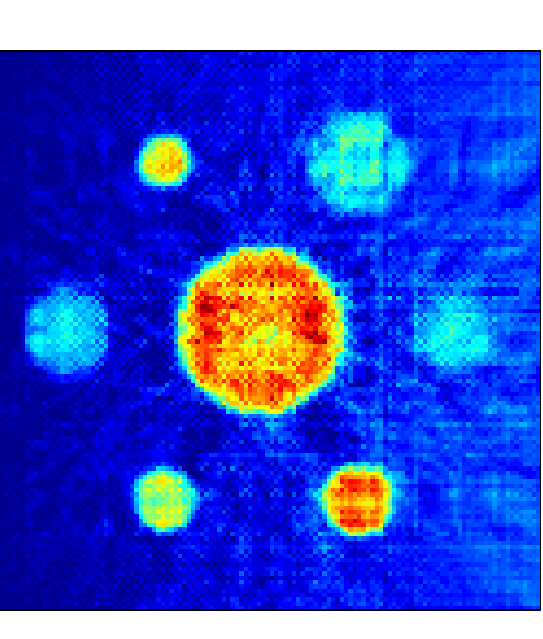}   &
\includegraphics[width=1in, trim = 31mm 5mm 25mm 0mm, clip=true]{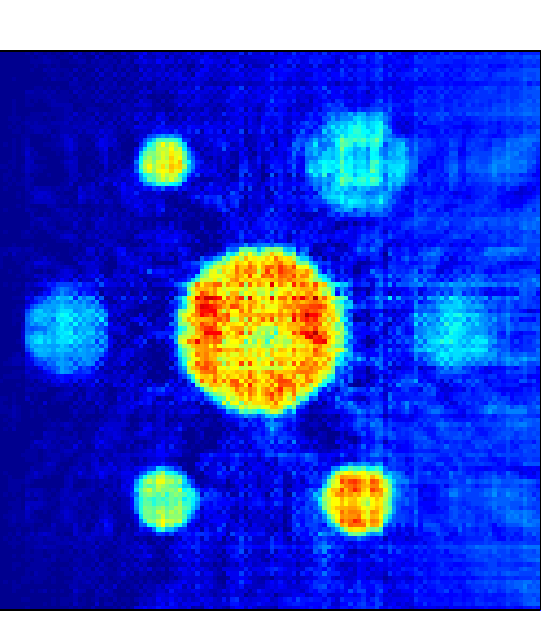} & \vspace{-0.2cm}  \\
\text{\small    FBP}&\text{\small TNN-1}&\text{\small TNN-2} \\

\includegraphics[width=1.39in, trim = 18mm 5mm 15mm 0mm, clip=true]{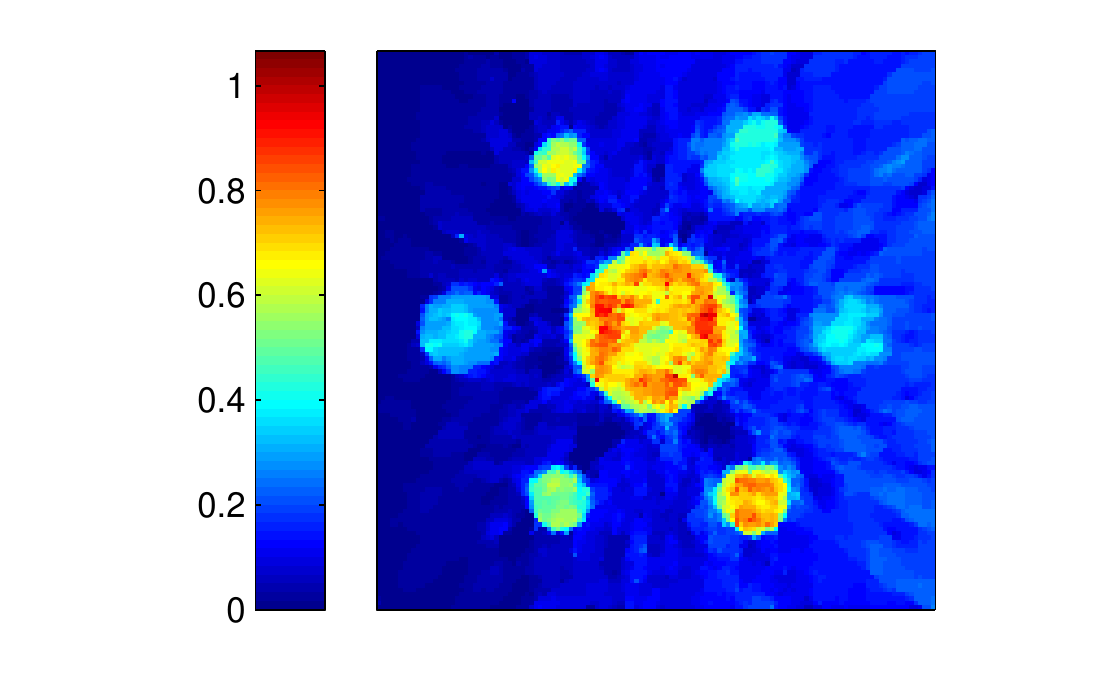}   &
\includegraphics[width=1.in, trim = 38mm 6mm 32mm 3mm, clip=true]{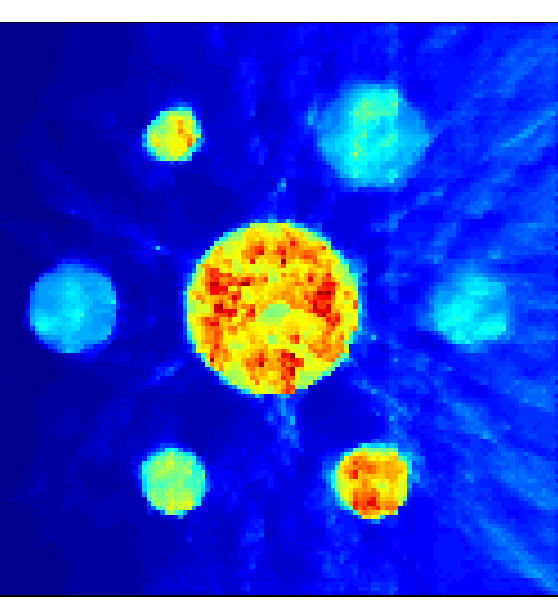}  &
\includegraphics[width=1in, trim = 31mm 5mm 25mm 0mm, clip=true]{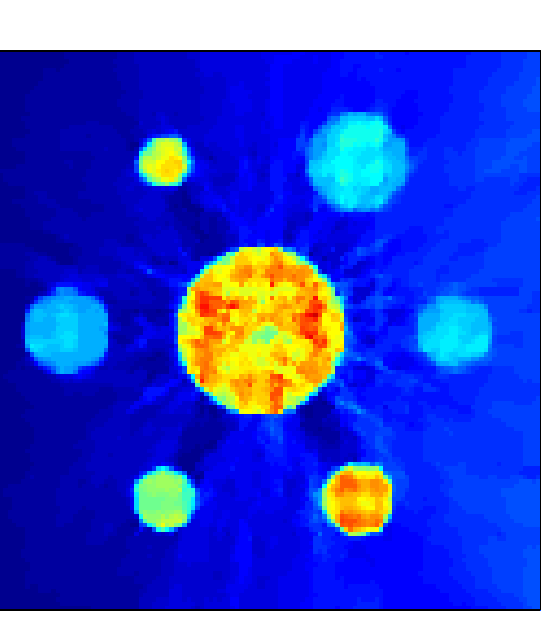}  &
\includegraphics[width=1in, trim = 31mm 5mm 25mm 0mm, clip=true]{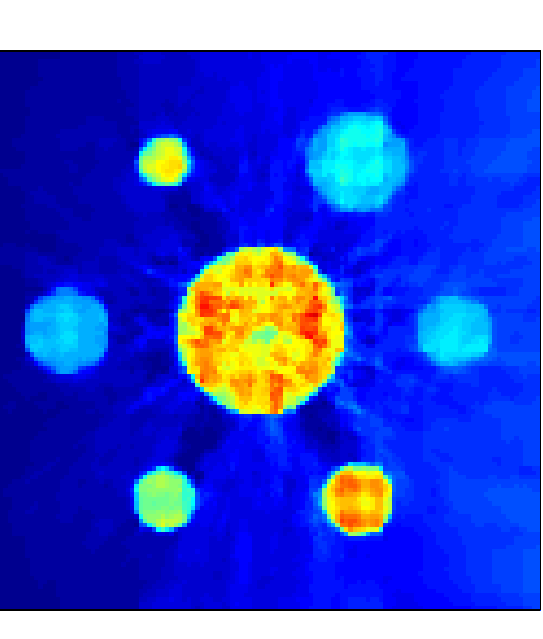}\vspace{-0.2cm} \\
\text{\small TV}&\text{\small 3D-TV}&\text{\small TNN-1+TV}&\text{\small TNN-2+TV}
\end{array}$
\caption{Phantom-2: Reconstructions results for 25 keV.}
\label{fig:results_textured_25}
\end{figure*}

\begin{figure*}[h!t!]
\centering
$\begin{array}{cccc}
\includegraphics[width=1.39in, trim = 18mm 5mm 15mm 0mm, clip=true]{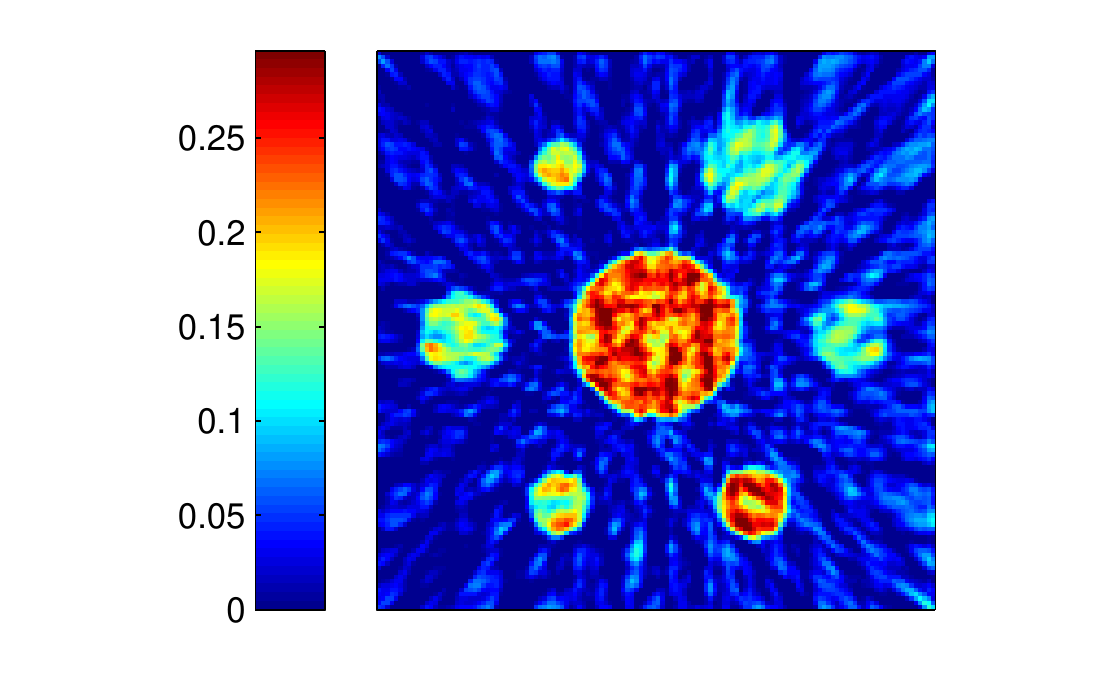}  &
\includegraphics[width=1in, trim = 31mm 5mm 25mm 0mm, clip=true]{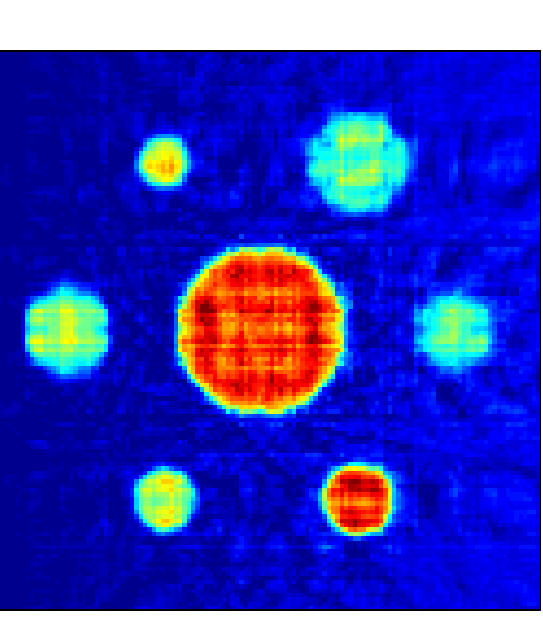}   &
\includegraphics[width=1in, trim = 31mm 5mm 25mm 0mm, clip=true]{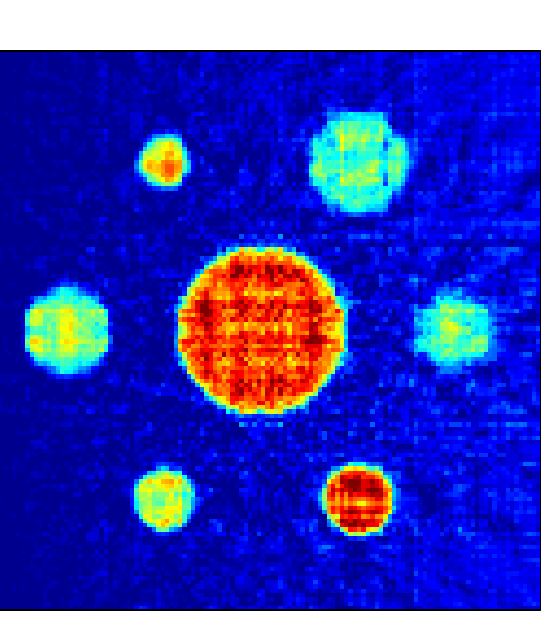} & \vspace{-0.2cm}  \\
\text{\small    FBP}&\text{\small TNN-1}&\text{\small TNN-2} \\

\includegraphics[width=1.39in, trim = 18mm 5mm 15mm 0mm, clip=true]{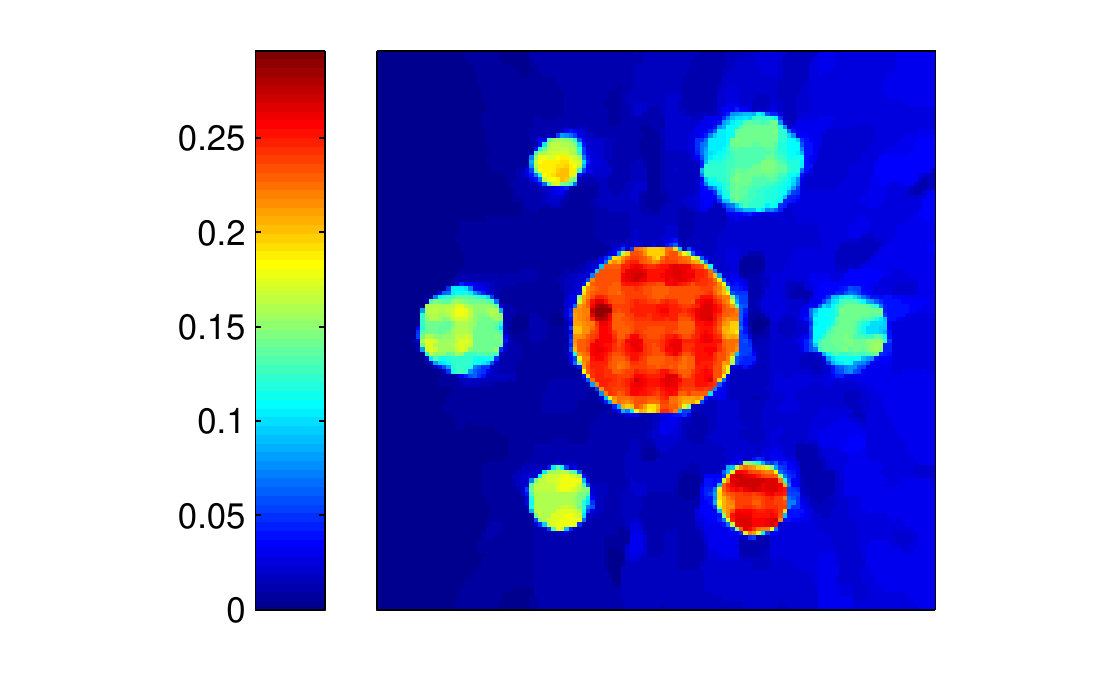}   &
\includegraphics[width=1.in, trim = 38mm 6mm 32mm 3mm, clip=true]{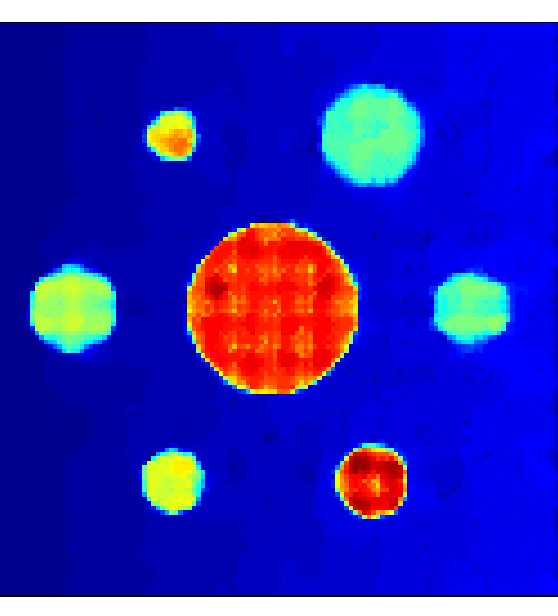}   &
\includegraphics[width=1in, trim = 31mm 5mm 25mm 0mm, clip=true]{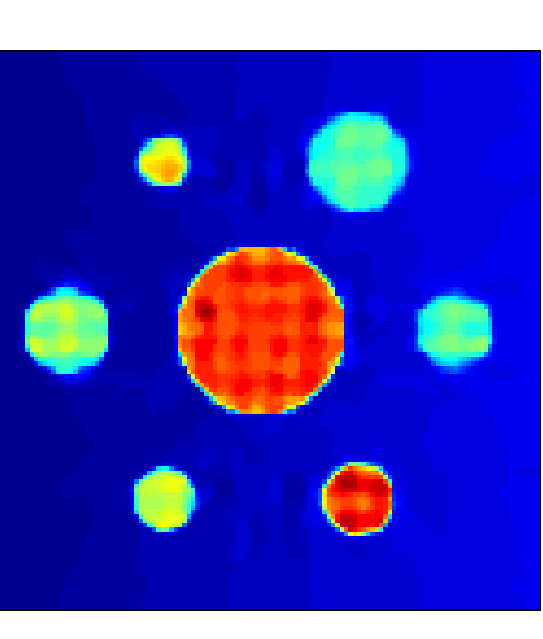}  &
\includegraphics[width=1in, trim = 31mm 5mm 25mm 0mm, clip=true]{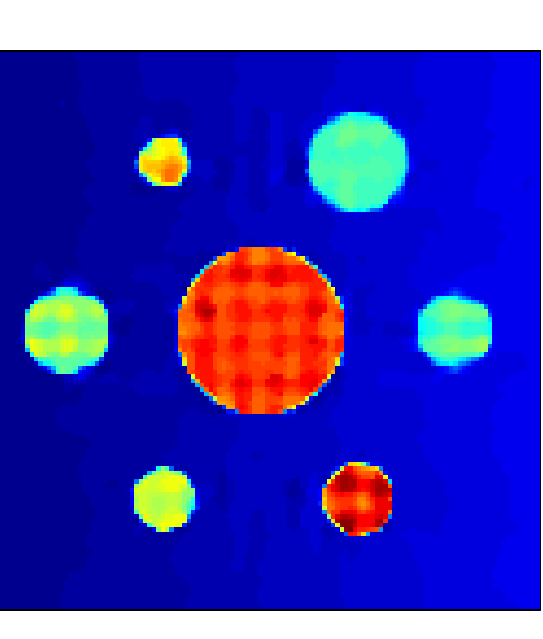} \vspace{-0.2cm} \\
\text{\small TV}&\text{\small 3D-TV}&\text{\small TNN-1+TV}&\text{\small TNN-2+TV}
\end{array}$
\caption{Phantom-2: Reconstructions results for 85 keV.}
\label{fig:results_textured_85}
\end{figure*}

\begin{figure*}[h!t!]
\centering
$\begin{array}{cc}
\includegraphics[width=1.37in, trim = 28mm 6mm 22mm 3mm, clip=true]{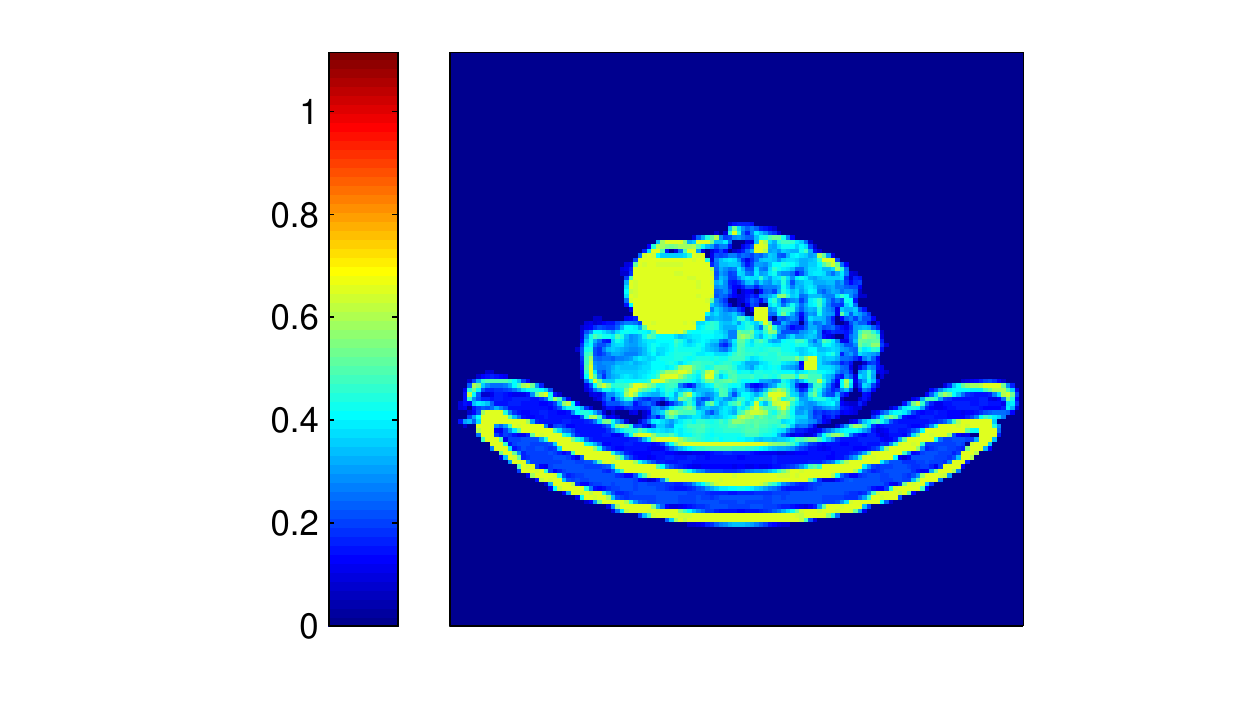}  &
\includegraphics[width=1.37in, trim = 28mm 6mm 22mm 3mm, clip=true]{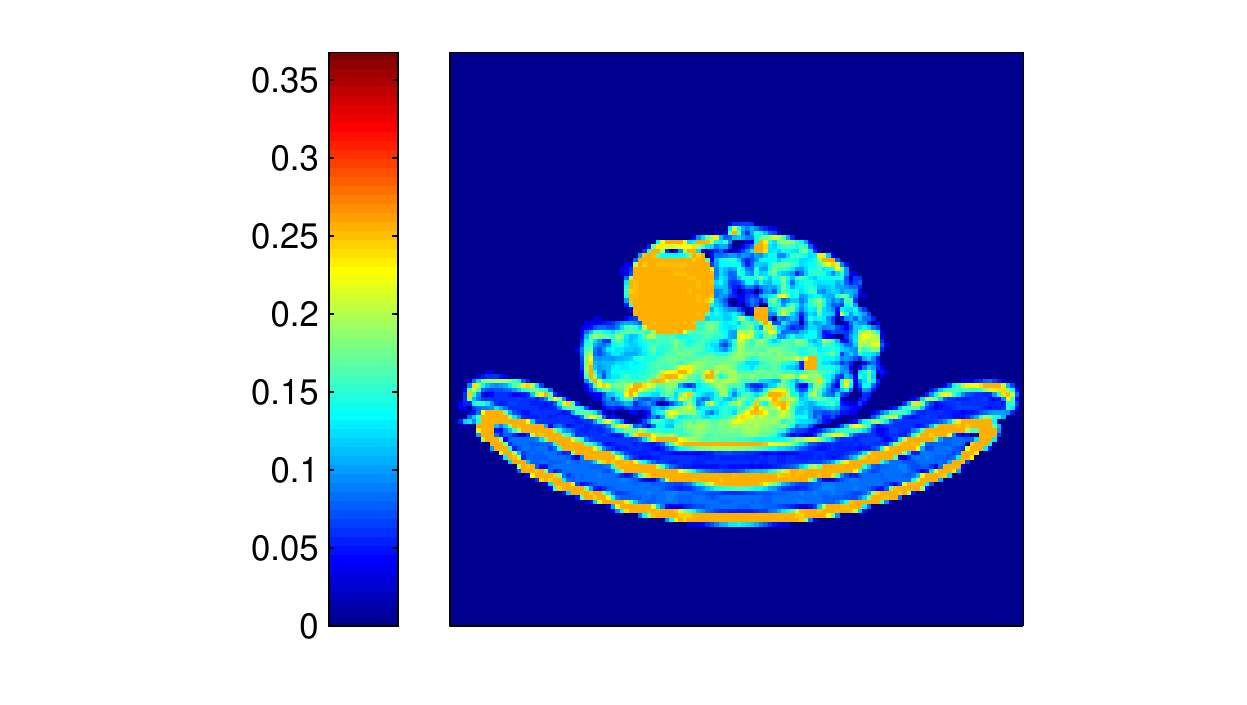}\vspace{-0.2cm} \\
\text{\small   Ground truth 25 keV}&\text{\small Ground  truth 85 keV}
\end{array}$
\caption{ Ground truth for Phantom-3. Left: 25 keV. Right: 85 keV}
\label{fig:truth3}
\end{figure*}

\begin{figure*}[h!t!]
\centering
$\begin{array}{cccc}
\includegraphics[width=1.37in, trim = 28mm 6mm 22mm 3mm, clip=true]{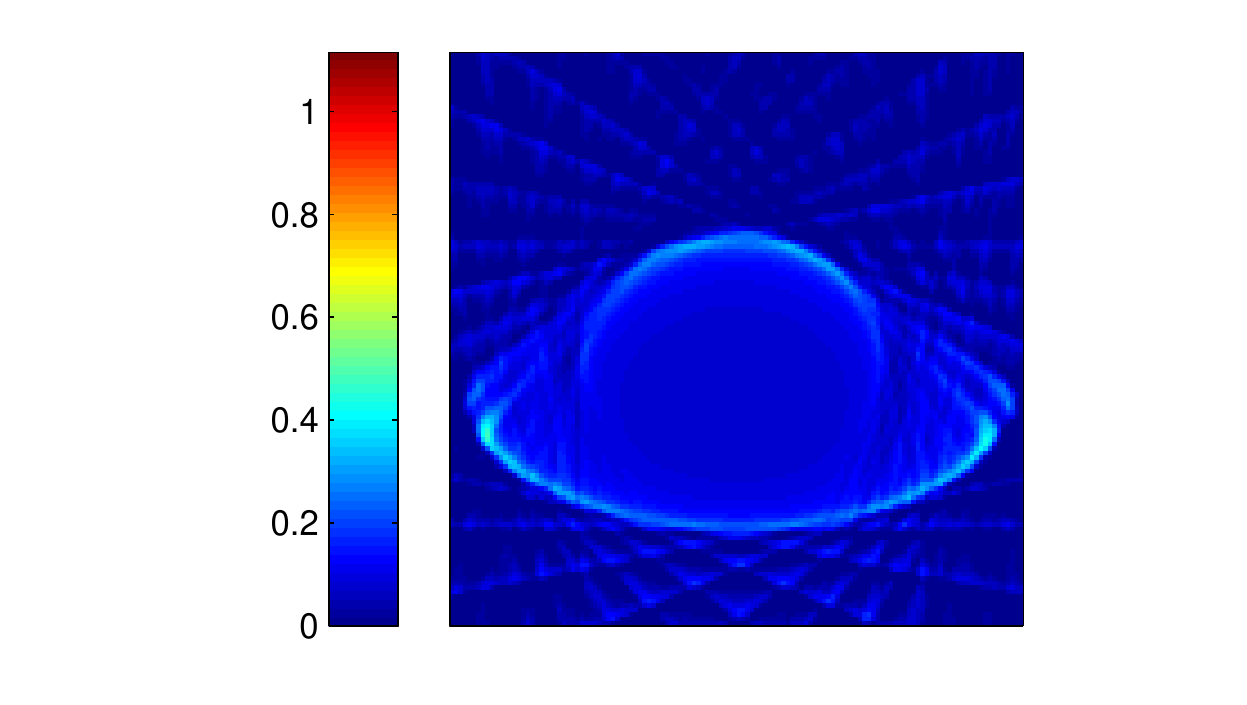}  &
\includegraphics[width=1.03in, trim = 38mm 6mm 32mm 3mm, clip=true]{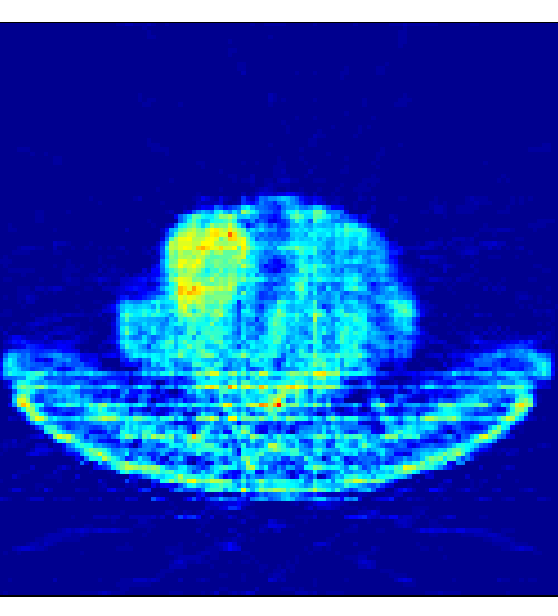}   &
\includegraphics[width=1.03in, trim = 38mm 6mm 32mm 3mm, clip=true]{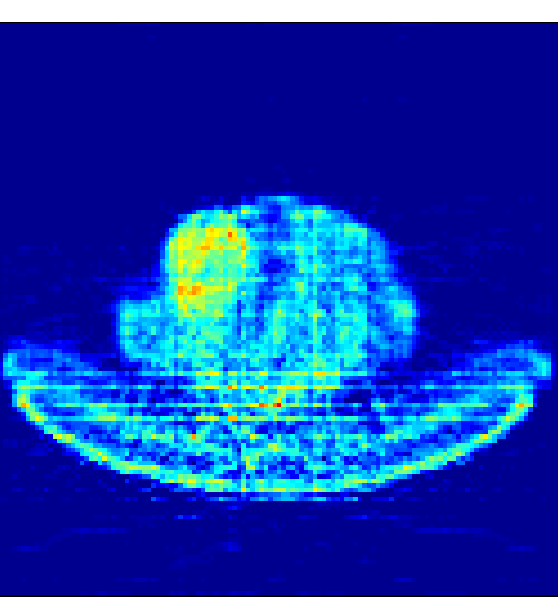} & \vspace{-0.2cm}  \\
\text{\small    FBP}&\text{\small TNN-1}&\text{\small TNN-2} \\

\includegraphics[width=1.37in, trim = 28mm 6mm 22mm 3mm, clip=true]{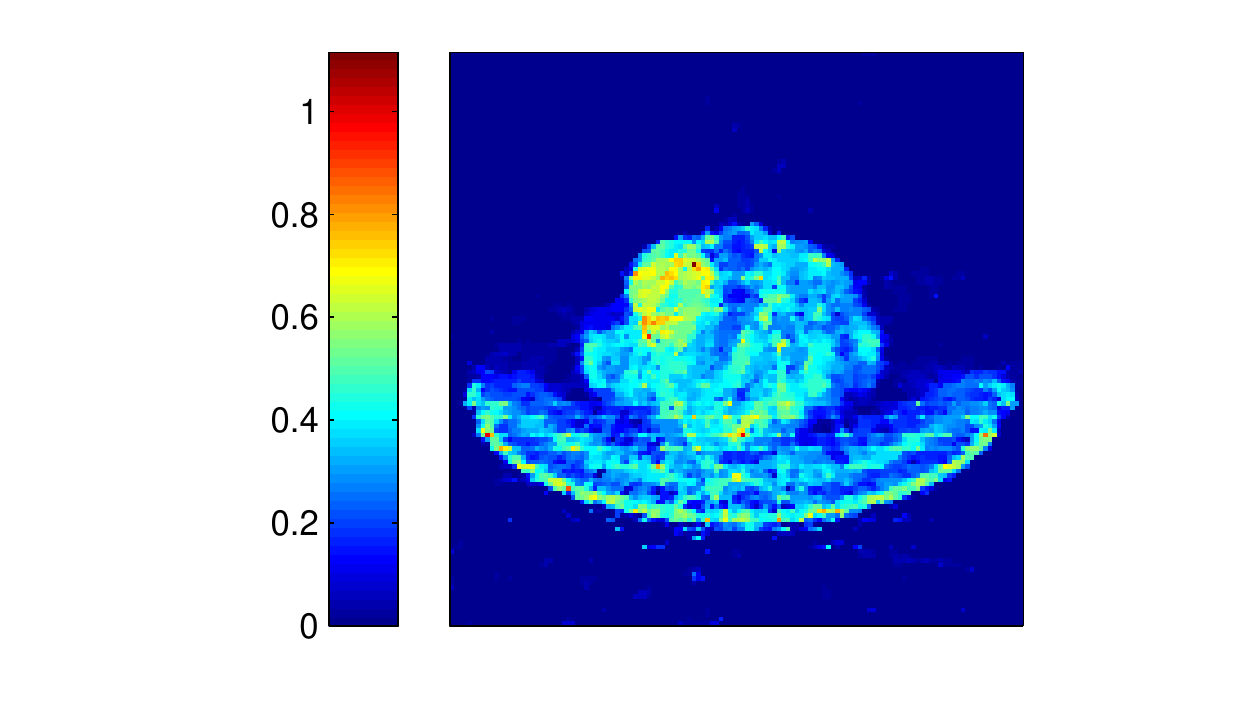}   &
\includegraphics[width=1.03in, trim = 38mm 6mm 32mm 3mm, clip=true]{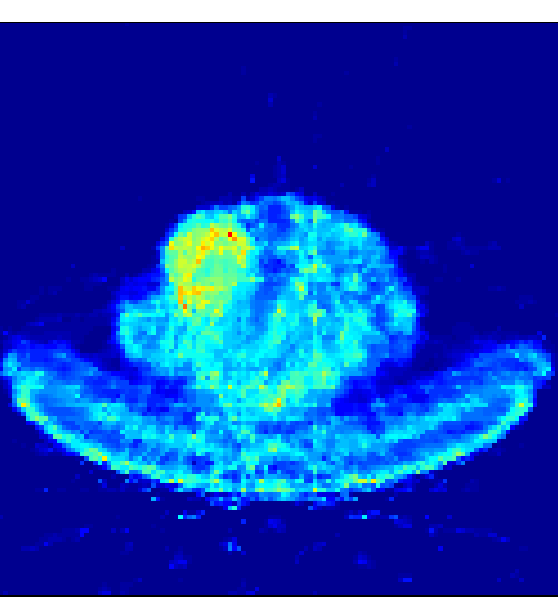}   &
\includegraphics[width=1.03in, trim = 38mm 6mm 32mm 3mm, clip=true]{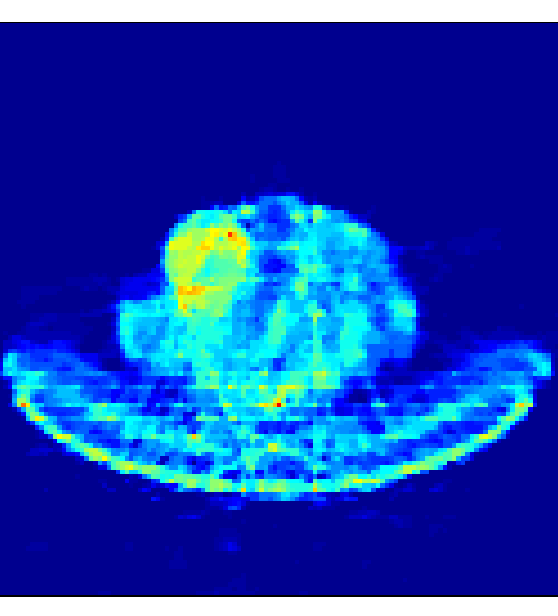}  &
\includegraphics[width=1.03in, trim = 38mm 6mm 32mm 3mm, clip=true]{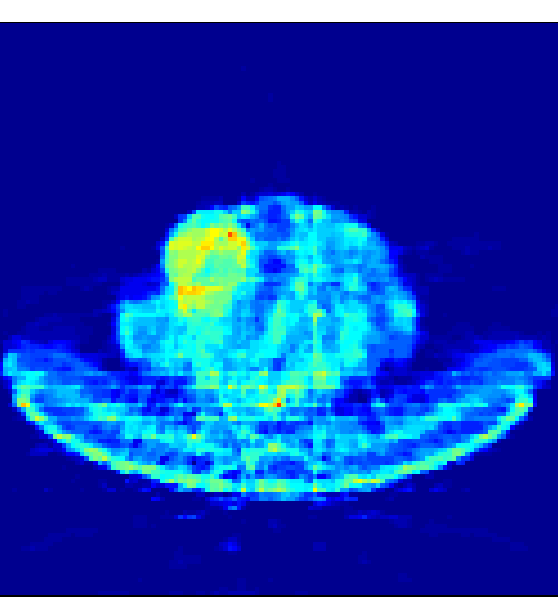} \vspace{-0.2cm} \\
\text{\small TV}&\text{\small 3D-TV}&\text{\small TNN-1+TV}&\text{\small TNN-2+TV}
\end{array}$
\caption{Phantom-3: Reconstructions results for 25 keV.}
\label{fig:results_duffle_25}
\end{figure*}

\begin{figure*}[h!t!]
\centering
$\begin{array}{cccc}
\includegraphics[width=1.37in, trim = 28mm 6mm 22mm 3mm, clip=true]{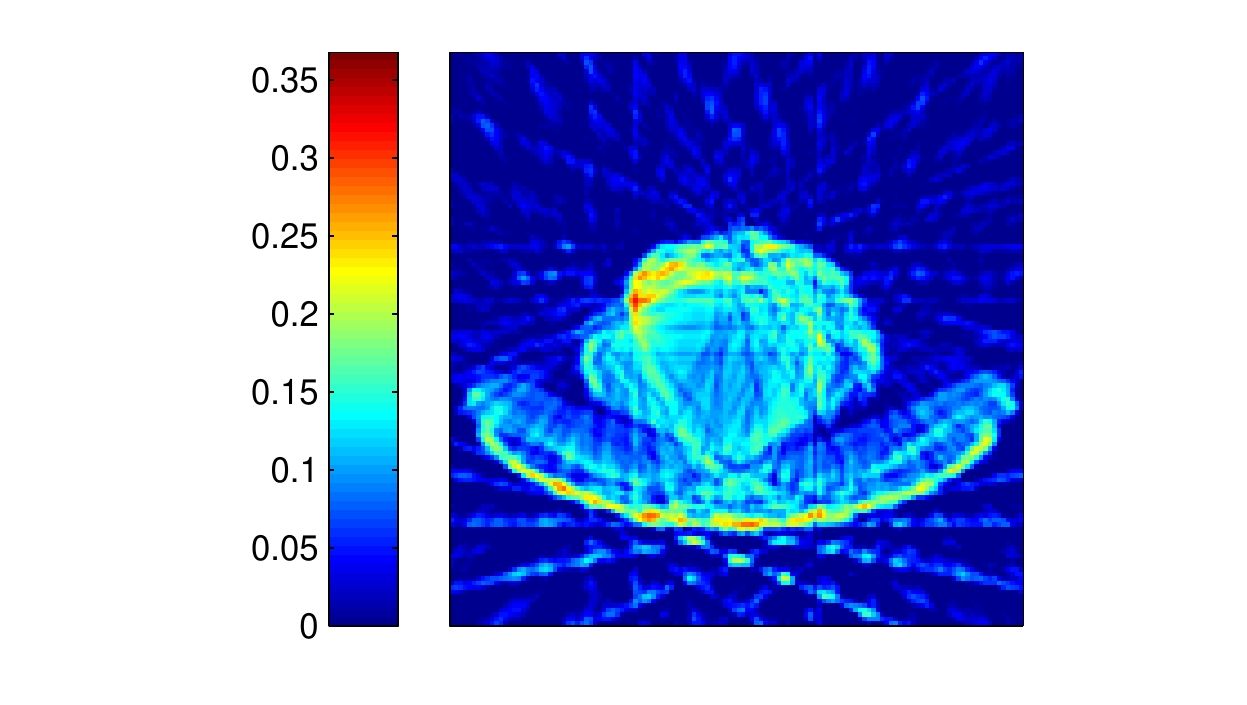}  &
\includegraphics[width=1.03in, trim = 38mm 6mm 32mm 3mm, clip=true]{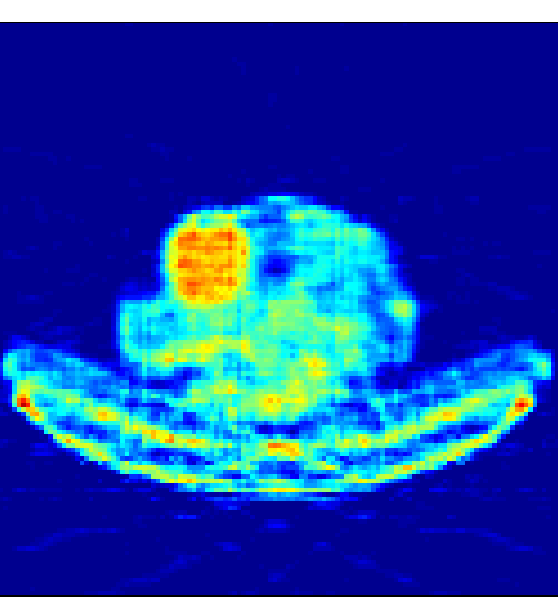}   &
\includegraphics[width=1.03in, trim = 38mm 6mm 32mm 3mm, clip=true]{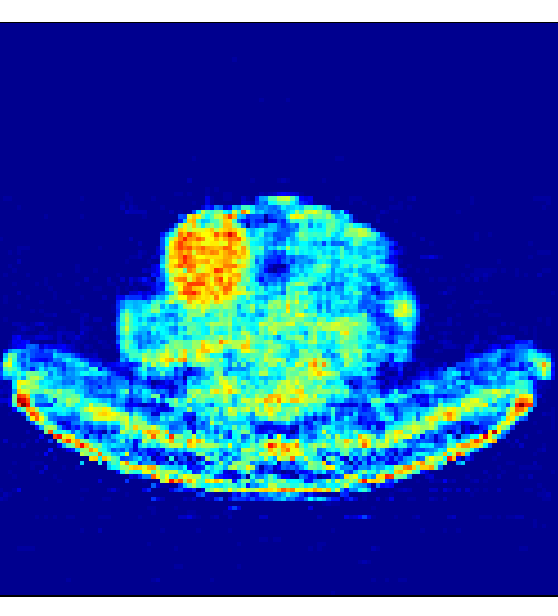} & \vspace{-0.2cm}  \\
\text{\small    FBP}&\text{\small TNN-1}&\text{\small TNN-2} \\

\includegraphics[width=1.37in, trim = 28mm 6mm 22mm 3mm, clip=true]{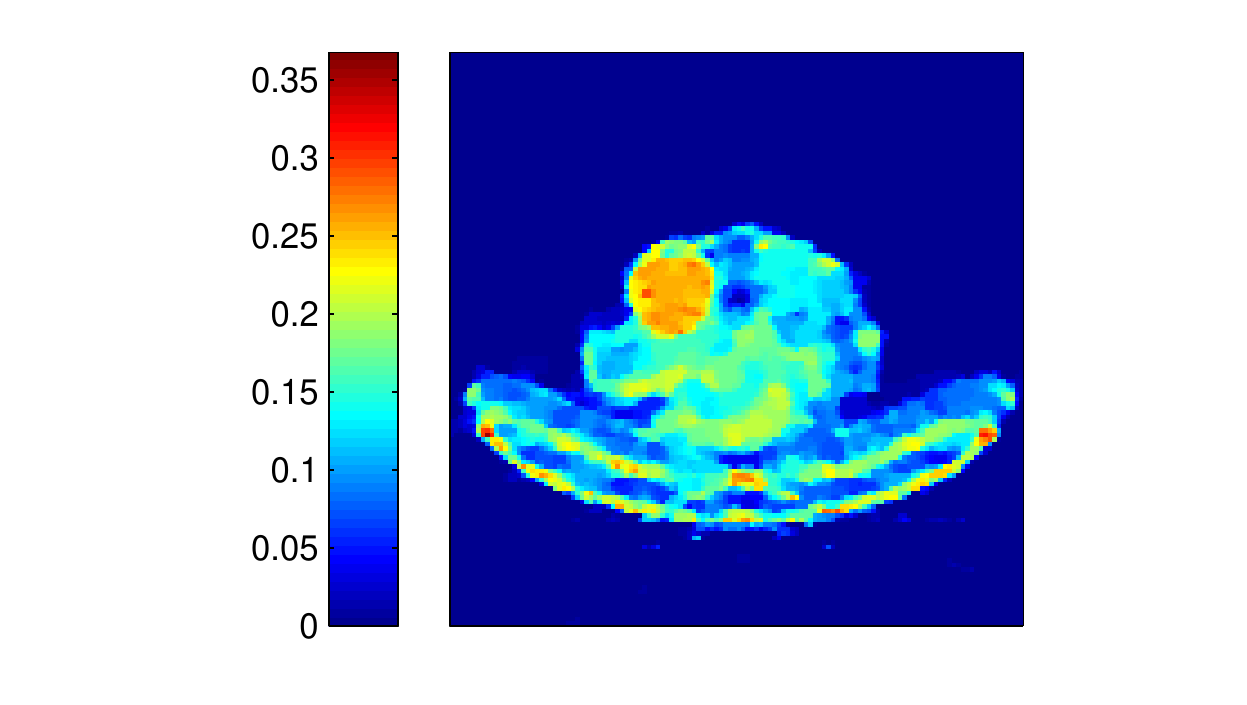}   &
\includegraphics[width=1.03in, trim = 38mm 6mm 32mm 3mm, clip=true]{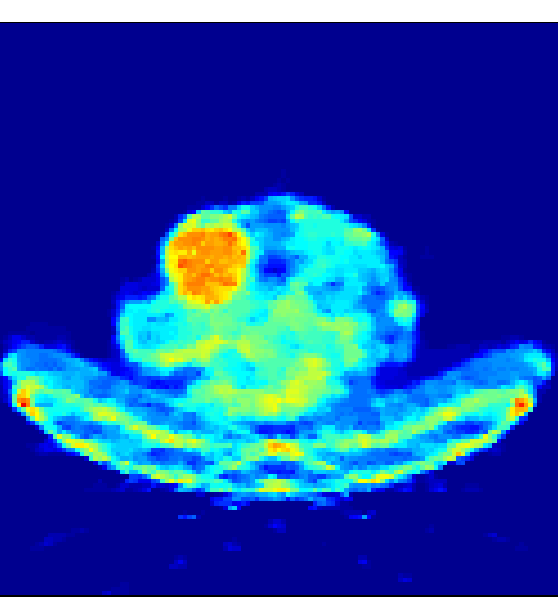}   &
\includegraphics[width=1.03in, trim = 38mm 6mm 32mm 3mm, clip=true]{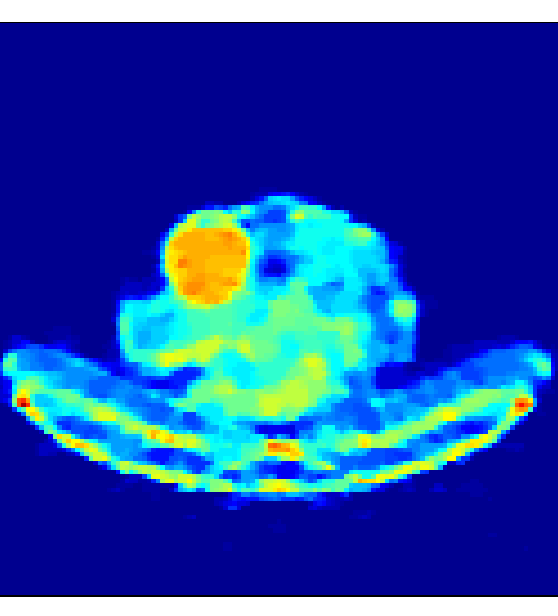}  &
\includegraphics[width=1.03in, trim = 38mm 6mm 32mm 3mm, clip=true]{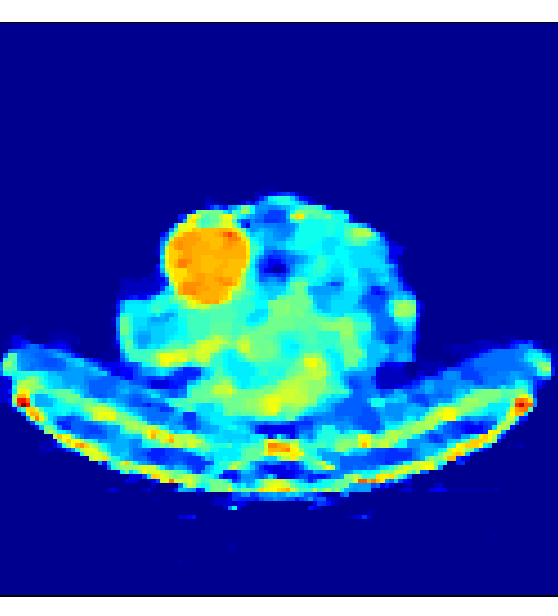} \vspace{-0.2cm} \\
\text{\small TV}&\text{\small 3D-TV}&\text{\small TNN-1+TV}&\text{\small TNN-2+TV}
\end{array}$
\caption{Phantom-3: Reconstructions results for 85 keV.}
\label{fig:results_duffle_85}
\end{figure*}

We show the error performance vs. iteration number for each method for 85keV and Phantom-1 in Fig. \ref{fig:error}. We display the reconstructions for the 25 keV and 85 keV bins as representatives of high and low regions of the spectra. Fig. \ref{fig:results_original_25}, Fig. \ref{fig:results_textured_25} and Fig. \ref{fig:results_duffle_25} show reconstruction results for 25 keV images; Fig. \ref{fig:results_original_85}, Fig. \ref{fig:results_textured_85} and Fig. \ref{fig:results_duffle_85} show reconstruction results for 85 keV images for both phantoms. Table \ref{table:compare_fbp} gives the elapsed times at particular iterations when FBP is outperformed for each method. Table \ref{table:results1}, Table \ref{table:results2} and Table \ref{table:results3} give quantitative error performance as well as the computation times.

Firstly, we observe that pure FBP fails to provide reasonable reconstructions at low energies due to limited number of views and noise.The proposed methods outperformed FBP in at most 3 iterations for 85keV and Phantom-1. Second, when TNN-1 or TNN-2 is used as the only regularizer, they provide considerable noise reduction while preserving much of the detail. Additionally, as seen in Table \ref{table:results1} they allow rapid processing relative to the other methods considered here. When they are combined with TV, TNN-1 and TNN-2 regularizers enhance its detail preserving capabilities and increases the reconstruction quality of low energy at the price of increased computational burden, which can be observed especially in the examples with Phantom-2. Although 3D-TV outperforms TV as it incorporates the smoothness in the energy dimension, TNN-1 and TNN-2 combined with TV gives superior results.  

Finally, we observe that TNN-1 and TNN-2 perform similarly in terms of image quality when they are the only regularizers that are used. However, for the 85 keV images we see that when combined with TV, TNN-2 outperforms TNN-1.

\begin{figure*}[h!t!]
\centering
$\begin{array}{cccc}
\includegraphics[width=1.39in, trim = 18mm 5mm 15mm 0mm, clip=true]{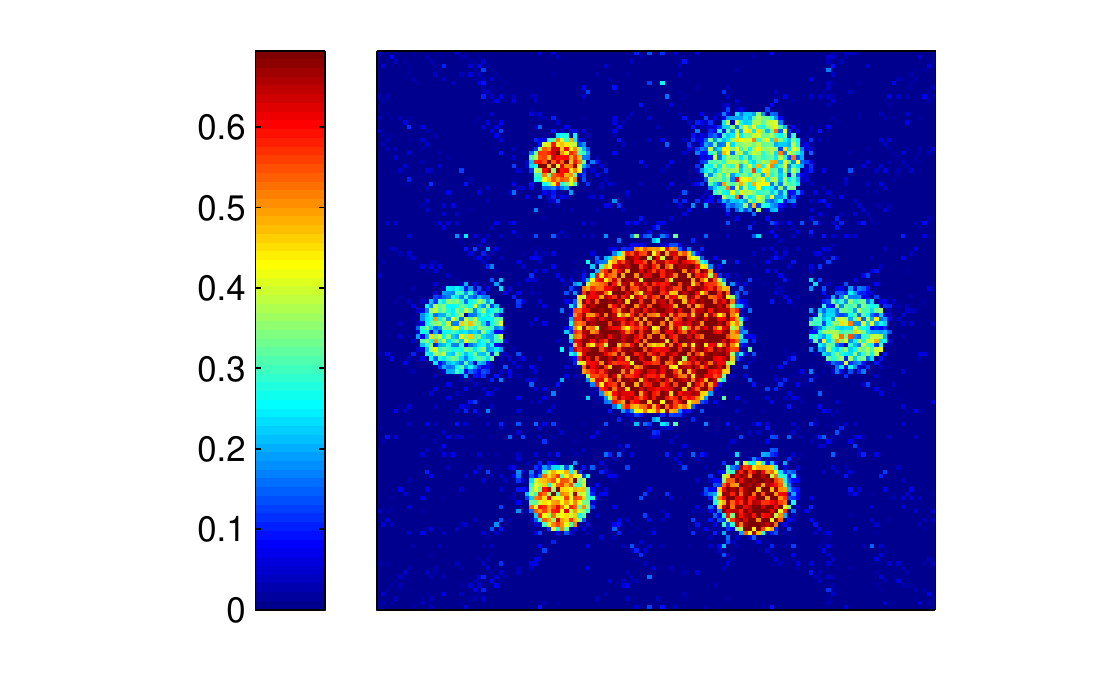}  &
\includegraphics[width=1in, trim = 31mm 5mm 25mm 0mm, clip=true]{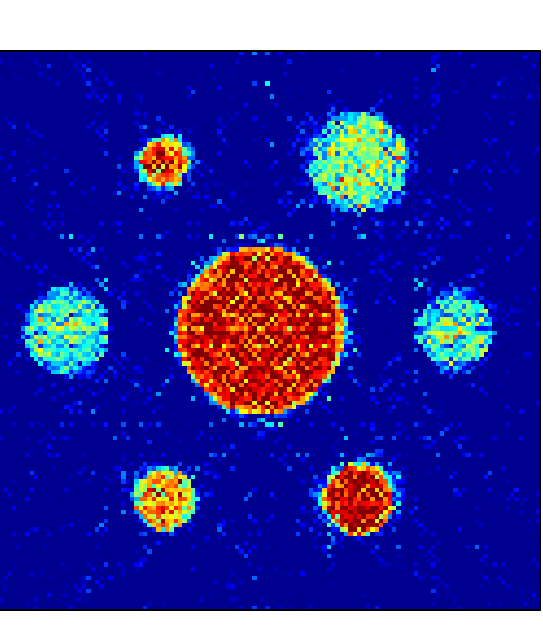}
&
\includegraphics[width=1in, trim = 31mm 5mm 25mm 0mm, clip=true]{nuc_25.pdf}  &
\includegraphics[width=1in, trim = 31mm 5mm 25mm 0mm, clip=true]{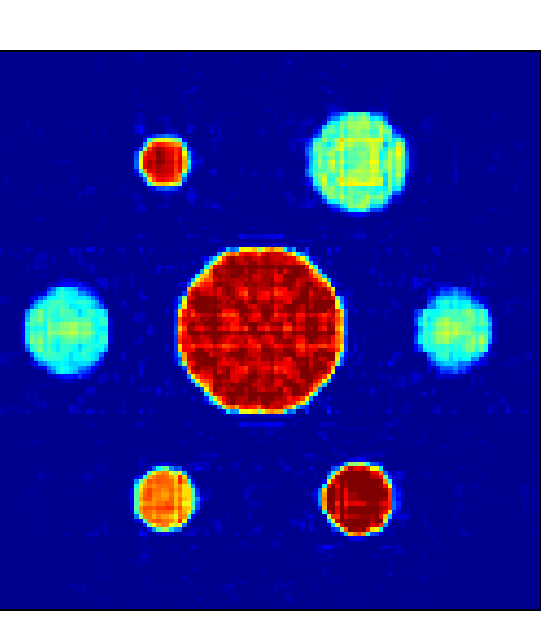} \vspace{-0.2cm} \\

\text{\small TNN-1,} {\small\gamma_{1,2}=0} &\text{\small TNN-1,} {\small\gamma_{1,2}=0} & \text{\small TNN-1}&\text{\small TNN-1} \\
 &\text{25 Bins} &  & \text{25 Bins}

\end{array}$
\caption{Phantom-1: Reconstructions results for 20 keV with TNN-1 where $\gamma_1$ and $\gamma_2$ are set to 0.}
\label{fig:results3_1}
\end{figure*}
\begin{figure*}[h!t!]
\centering
$\begin{array}{cccc}
\includegraphics[width=1.39in, trim = 18mm 5mm 15mm 0mm, clip=true]{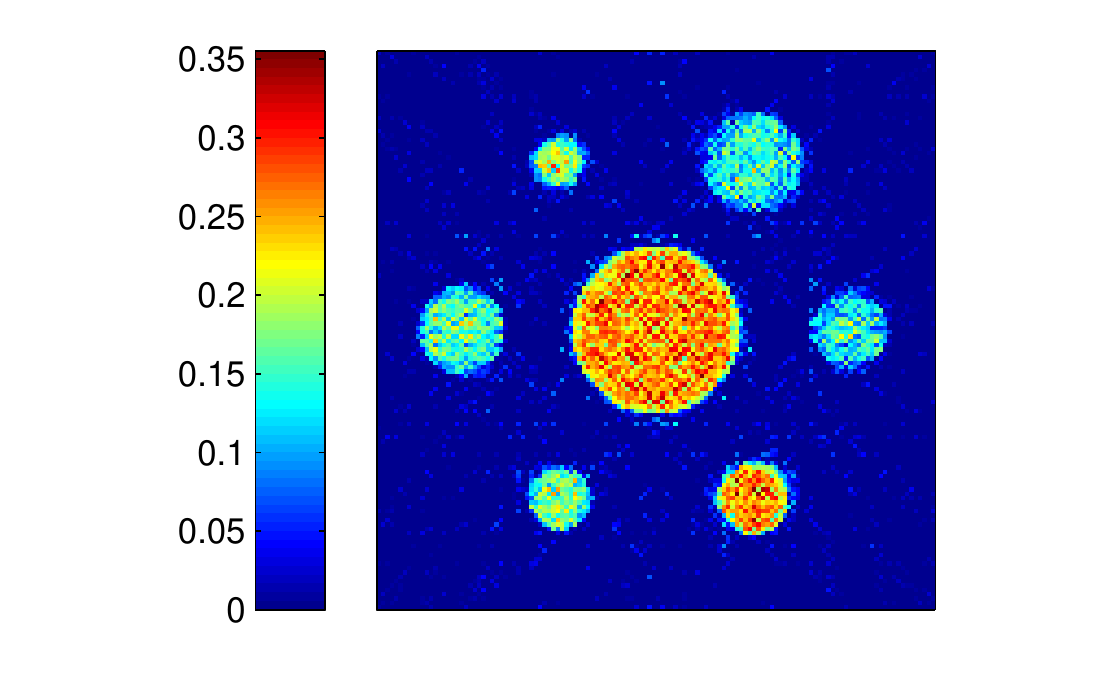}  &
\includegraphics[width=1in, trim = 31mm 5mm 25mm 0mm, clip=true]{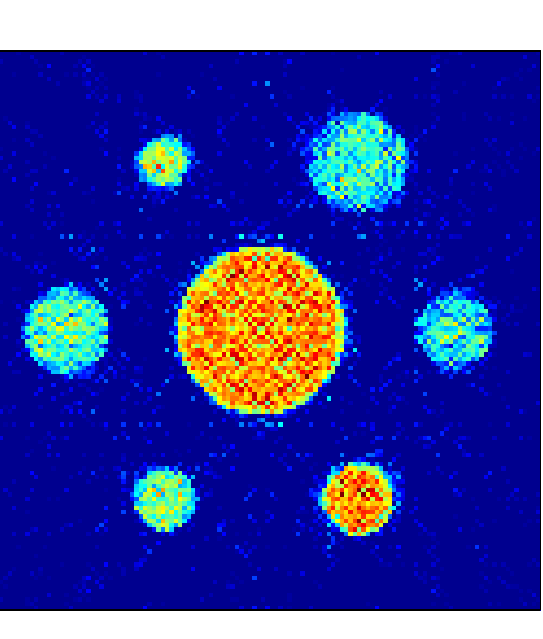}
&
\includegraphics[width=1in, trim = 31mm 5mm 25mm 0mm, clip=true]{nuc_80.pdf}  &
\includegraphics[width=1in, trim = 31mm 5mm 25mm 0mm, clip=true]{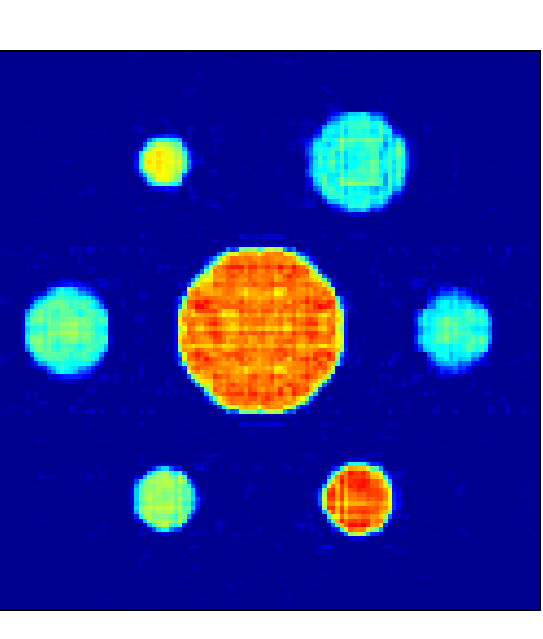} \vspace{-0.2cm} \\

\text{\small TNN-1,} {\small\gamma_{1,2}=0} &\text{\small TNN-1,} {\small\gamma_{1,2}=0} & \text{\small TNN-1}&\text{\small TNN-1} \\
 &\text{25 Bins} &  & \text{25 Bins}

\end{array}$
\caption{Phantom-1: Reconstructions results for 85 keV with TNN-1 where $\gamma_1$ and $\gamma_2$ are set to 0.}
\label{fig:results3_2}
\end{figure*}

In the last example, we demonstrate the efficiency of the \emph{tensor} model in constructing a regularizer based on low-rank assumptions. We consider using the TNN-1 regularizer with $\gamma_1$ and $\gamma_2$ are set to 0. This is equivalent to applying the low rank prior to the multi-spectral matrix whose columns are vectorized images at different energies. In addition to the data set with 12 energy bins, we simulated data for 25 bins uniformly distributed between the same range of 25 keV and 85 keV with Phantom-1. All other parameters were kept the same. Fig. \ref{fig:results3_1}, Fig. \ref{fig:results3_2} and Table \ref{table:results33} shows the results for both data sets. We observe that that the tensor-based representation is needed to design useful nuclear norm regularization with the TNN-1 approach. Although increasing the number of bins from 12 to 25 introduces more redundancy in the energy dimension, the incorporation of the unfoldings in the spatial dimensions is still essential.

\begin{table}
\caption{Error performance of different methods for Phantom-1}

\centering
\begin{tabular}{lccc}
\hline\hline
Method & $E_{\ell_2} (25 \text{keV})$ & $E_{\ell_2} (85 \text{keV})$ & Comp. time (sec) \\
\hline
FBP        & 0.4507  & 0.2010  &4\\
TNN-1      & 0.0492  & 0.0335  &25\\
TNN-2      & 0.0299  & 0.0215  &112\\
TV         & 0.0149  & 0.0101  &249\\
3D-TV	   & 0.0078  & 0.0118  &196\\
TV+TNN-1   & 0.0056  & 0.0122  & 3615\\
TV+TNN-2   & 0.0066  & 0.0045  & 10879\\
\hline
\end{tabular}
\label{table:results1}
\end{table}

\begin{table}
\caption{Error performance of different methods for Phantom-2}

\centering
\begin{tabular}{lccc}
\hline\hline
Method & $E_{\ell_2} (25 \text{keV})$ & $E_{\ell_2} (85 \text{keV})$ & Comp. time (sec) \\
\hline
FBP        & 0.2102  & 0.1845  & 4\\
TNN-1      & 0.0583  & 0.0492  & 28\\
TNN-2      & 0.0598  & 0.0514  & 140\\
TV         & 0.0465  & 0.0202  & 284\\
3D-TV	   & 0.0140  & 0.0185  & 162\\
TV+TNN-1   & 0.0093  & 0.0175  & 3911\\
TV+TNN-2   & 0.0096  & 0.0102  & 11470\\
\hline
\end{tabular}
\label{table:results2}
\end{table}
\begin{table}
\caption{Error performance of different methods for Phantom-3}

\centering
\begin{tabular}{lccc}
\hline\hline
Method & $E_{\ell_2} (25 \text{keV})$ & $E_{\ell_2} (85 \text{keV})$ & Comp. time (sec) \\
\hline
FBP        & 0.6216  & 0.3482  & 4\\
TNN-1      & 0.1632  & 0.1057  & 18\\
TNN-2      & 0.1634  & 0.1124  & 51\\
TV         & 0.1482  & 0.0879  & 203\\
3D-TV	   & 0.1553  & 0.1028  & 112\\
TV+TNN-1   & 0.1365  & 0.0881  & 2891\\
TV+TNN-2   & 0.1395  & 0.0886  & 5570\\
\hline
\end{tabular}
\label{table:results3}
\end{table}
\begin{table}
\caption{ Error performance of the different unfolding trials with Phantom-1}

\centering
\begin{tabular}{lccc}
\hline\hline
Method & $E_{\ell_2} (20 \text{keV})$ & $E_{\ell_2} (85 \text{keV})$ & Comp. time (sec) \\
\hline
TNN                          & 0.0341  & 0.0335  & 25\\
TNN, $\gamma_{1,2}=0$          & 0.0708  & 0.0694  &25\\
TNN 25 bins                  & 0.0327  & 0.0314  & 64 \\
TNN, 25 bins $\gamma_{1,2}=0$  & 0.0713  & 0.0693  &72\\
\hline
\end{tabular}
\label{table:results33}
\end{table}

\section{Conclusions}\label{sec:conclusion}

In this paper, we provided an algorithmic framework for iterative multi-energy CT and showed that generalized tensor nuclear norm ideas can be used as regularizers. Additionally we proposed an alternative tensor nuclear norm based on t-SVD and a regularizer based on this new tensor nuclear norm. The ideas presented here can be extended to any type of inverse problem where a multi-linear description of the unknown is possible. Additionally, the tensor nuclear norm regularization can be generalized to higher dimensions. For instance, one can consider a the 5D problem with an additional spatial dimension and time dependency.

In future, incorporation of low rank-sparse decomposition approaches \cite{golbabaee2011hyperspectral,gao2011multi} in the tensor-based framework will be investigated. The applicability of Tucker and CANDECOMP/PARAFAC decomposition techniques especially to reduce the dimension of the multi-energy data cube will also be considered. These directions will allow the design of more sophisticated tensor nuclear norm regularizers. Another important extension is to design an algorithm to estimate the redundancy along different dimensions, which will allow us to quantify the requirement of low rank priors.

As the goal of the multi-energy tomography problem is to reconstruct structurally similar images of the X-ray attenuation at each energy, it is sensible to design Tikhonov type, fast regularization techniques that explicitly enforce structural similarity. Design of such alternative regularizers is important especially for medical imaging applications considering the cartooning effect of TV regularization.

Development of an automatic determination of the regularization parameters is also an important future direction to increase the practicality of the algorithms presented. Considerable reduction of computation time is required in order to be able to use our methods in practice. Therefore, an important area of future work should be devoted to increasing the efficiency of implementation by,\emph{e.g.,} parallel computing and code optimization. Finally, the algorithms should be tested for real-life scenarios with experimental data and higher resolution reconstructions.


\begin{thebibliography}{10}
\providecommand{\url}[1]{#1}

\bibitem{whiting2006properties}
B.~Whiting, P.~Massoumzadeh, O.~Earl, J.~OSullivan, D.~Snyder, and
  J.~Williamson, ``Properties of preprocessed sinogram data in x-ray computed
  tomography,'' \emph{Medical physics}, vol.~33, p. 3290, 2006.

\bibitem{bouman}
C.~Bouman and K.~Sauer, ``A unified approach to statistical tomography using
  coordinate descent optimization,'' \emph{Image Processing, IEEE Transactions
  on}, vol.~5, no.~3, pp. 480--492, mar 1996.

\bibitem{pan2009commercial}
X.~Pan, E.~Sidky, and M.~Vannier, ``Why do commercial ct scanners still employ
  traditional, filtered back-projection for image reconstruction?''
  \emph{Inverse problems}, vol.~25, p. 123009, 2009.

\bibitem{elbakri2002statistical}
I.~Elbakri and J.~Fessler, ``Statistical image reconstruction for polyenergetic
  x-ray computed tomography,'' \emph{Medical Imaging, IEEE Transactions on},
  vol.~21, no.~2, pp. 89--99, 2002.

\bibitem{semerci}
O.~Semerci and E.~Miller, ``A parametric level-set approach to simultaneous
  object identification and background reconstruction for dual-energy computed
  tomography,'' \emph{Image Processing, IEEE Transactions on}, vol.~21, no.~5,
  pp. 2719 --2734, may 2012.

\bibitem{wang2008outlook}
G.~Wang, H.~Yu, and B.~De~Man, ``An outlook on x-ray ct research and
  development,'' \emph{Medical physics}, vol.~35, p. 1051, 2008.

\bibitem{de2001iterative}
B.~De~Man, J.~Nuyts, P.~Dupont, G.~Marchal, and P.~Suetens, ``An iterative
  maximum-likelihood polychromatic algorithm for ct,'' \emph{Medical Imaging,
  IEEE Transactions on}, vol.~20, no.~10, pp. 999--1008, 2001.

\bibitem{shikhaliev2008energy}
P.~Shikhaliev, ``Energy-resolved computed tomography: first experimental
  results,'' \emph{Physics in medicine and biology}, vol.~53, p. 5595, 2008.

\bibitem{gao2011multi}
H.~Gao, H.~Yu, S.~Osher, and G.~Wang, ``Multi-energy ct based on a prior rank,
  intensity and sparsity model (prism),'' \emph{Inverse problems}, vol.~27, p.
  115012, 2011.

\bibitem{shikhaliev2011photon}
P.~Shikhaliev and S.~Fritz, ``Photon counting spectral ct versus conventional
  ct: comparative evaluation for breast imaging application,'' \emph{Physics in
  medicine and biology}, vol.~56, p. 1905, 2011.

\bibitem{schlomka2008experimental}
J.~Schlomka, E.~Roessl, R.~Dorscheid, S.~Dill, G.~Martens, T.~Istel,
  C.~B{\"a}umer, C.~Herrmann, R.~Steadman, G.~Zeitler \emph{et~al.},
  ``Experimental feasibility of multi-energy photon-counting k-edge imaging in
  pre-clinical computed tomography,'' \emph{Physics in medicine and biology},
  vol.~53, p. 4031, 2008.

\bibitem{ivakhnenko2010novel}
V.~Ivakhnenko, ``A novel quasi-linearization method for ct image reconstruction
  in scanners with a multi-energy detector system,'' \emph{Nuclear Science,
  IEEE Transactions on}, vol.~57, no.~2, pp. 870--879, 2010.

\bibitem{iwanczyk2009photon}
J.~Iwanczyk, E.~Nygard, O.~Meirav, J.~Arenson, W.~Barber, N.~Hartsough,
  N.~Malakhov, and J.~Wessel, ``Photon counting energy dispersive detector
  arrays for x-ray imaging,'' \emph{Nuclear Science, IEEE Transactions on},
  vol.~56, no.~3, pp. 535--542, 2009.

\bibitem{barber2009characterization}
W.~Barber, E.~Nygard, J.~Iwanczyk, M.~Zhang, E.~Frey, B.~Tsui, J.~Wessel,
  N.~Malakhov, G.~Wawrzyniak, N.~Hartsough \emph{et~al.}, ``Characterization of
  a novel photon counting detector for clinical ct: count rate, energy
  resolution, and noise performance,'' in \emph{Proceedings of the SPIE Medical
  Imaging Conference}, vol. 7258, 2009, p. 725824.

\bibitem{ying}
Z.~Ying, R.~Naidu, and C.~Crawford, ``Dual energy computed tomography for
  explosive detection,'' \emph{J. of X-ray Sci. and Tech.}, vol.~14, no.~4, pp.
  235--256, 2006.

\bibitem{singh2003explosives}
S.~Singh and M.~Singh, ``{Explosives detection systems (EDS) for aviation
  security},'' \emph{Signal Processing}, vol.~83, no.~1, pp. 31--55, 2003.

\bibitem{candes2009robust}
E.~Candes, X.~Li, Y.~Ma, and J.~Wright, ``Robust principal component
  analysis?'' \emph{Arxiv preprint ArXiv:0912.3599}, 2009.

\bibitem{cai2008singular}
J.~Cai, E.~Candes, and Z.~Shen, ``A singular value thresholding algorithm for
  matrix completion,'' \emph{Arxiv preprint Arxiv:0810.3286}, 2008.

\bibitem{candes2009exact}
E.~Cand{\`e}s and B.~Recht, ``Exact matrix completion via convex
  optimization,'' \emph{Foundations of Computational Mathematics}, vol.~9,
  no.~6, pp. 717--772, 2009.

\bibitem{tomioka2010estimation}
R.~Tomioka, K.~Hayashi, and H.~Kashima, ``Estimation of low-rank tensors via
  convex optimization,'' \emph{Arxiv preprint arXiv:1010.0789}, 2010.

\bibitem{liu2009tensor}
J.~Liu, P.~Musialski, P.~Wonka, and J.~Ye, ``Tensor completion for estimating
  missing values in visual data,'' in \emph{Computer Vision, 2009 IEEE 12th
  International Conference on}.\hskip 1em plus 0.5em minus 0.4em\relax IEEE,
  2009, pp. 2114--2121.

\bibitem{cai2012cine}
J.~Cai, X.~Jia, H.~Gao, S.~Jiang, Z.~Shen, and H.~Zhao, ``Cine cone beam ct
  reconstruction using low-rank matrix factorization: algorithm and a
  proof-of-princple study,'' \emph{Arxiv preprint arXiv:1204.3595}, 2012.

\bibitem{carroll1970analysis}
J.~Carroll and J.~Chang, ``Analysis of individual differences in
  multidimensional scaling via an n-way generalization of “eckart-young”
  decomposition,'' \emph{Psychometrika}, vol.~35, no.~3, pp. 283--319, 1970.

\bibitem{tucker1966some}
L.~Tucker, ``Some mathematical notes on three-mode factor analysis,''
  \emph{Psychometrika}, vol.~31, no.~3, pp. 279--311, 1966.

\bibitem{acar2009unsupervised}
E.~Acar and B.~Yener, ``Unsupervised multiway data analysis: A literature
  survey,'' \emph{Knowledge and Data Engineering, IEEE Transactions on},
  vol.~21, no.~1, pp. 6--20, 2009.

\bibitem{andersen2003practical}
C.~Andersen and R.~Bro, ``Practical aspects of parafac modeling of fluorescence
  excitation-emission data,'' \emph{Journal of Chemometrics}, vol.~17, no.~4,
  pp. 200--215, 2003.

\bibitem{kolda2008scalable}
T.~Kolda and J.~Sun, ``Scalable tensor decompositions for multi-aspect data
  mining,'' in \emph{Data Mining, 2008. ICDM'08. Eighth IEEE International
  Conference on}.\hskip 1em plus 0.5em minus 0.4em\relax IEEE, 2008, pp.
  363--372.

\bibitem{de2000multilinear}
L.~De~Lathauwer, B.~De~Moor, and J.~Vandewalle, ``A multilinear singular value
  decomposition,'' \emph{SIAM Journal on Matrix Analysis and Applications},
  vol.~21, no.~4, pp. 1253--1278, 2000.

\bibitem{vasilescu2002multilinear}
M.~Vasilescu and D.~Terzopoulos, ``Multilinear image analysis for facial
  recognition,'' in \emph{Pattern Recognition, 2002. Proceedings. 16th
  International Conference on}, vol.~2.\hskip 1em plus 0.5em minus 0.4em\relax
  IEEE, 2002, pp. 511--514.

\bibitem{kolda2009tensor}
T.~Kolda and B.~Bader, ``Tensor decompositions and applications,'' \emph{SIAM
  review}, vol.~51, no.~3, p. 455, 2009.

\bibitem{bro2003new}
R.~Bro and H.~Kiers, ``A new efficient method for determining the number of
  components in parafac models,'' \emph{Journal of Chemometrics}, vol.~17,
  no.~5, pp. 274--286, 2003.

\bibitem{recht2010guaranteed}
B.~Recht, M.~Fazel, and P.~Parrilo, ``Guaranteed minimum-rank solutions of
  linear matrix equations via nuclear norm minimization,'' \emph{SIAM review},
  vol.~52, no.~3, pp. 471--501, 2010.

\bibitem{gandy2011tensor}
S.~Gandy, B.~Recht, and I.~Yamada, ``Tensor completion and low-n-rank tensor
  recovery via convex optimization,'' \emph{Inverse Problems}, vol.~27, p.
  025010, 2011.

\bibitem{signoretto2011tensor}
M.~Signoretto, R.~Van~de Plas, B.~De~Moor, and J.~Suykens, ``Tensor versus
  matrix completion: a comparison with application to spectral data,''
  \emph{IEEE Signal Processing Letters}, vol.~18, no.~7, p. 403, 2011.

\bibitem{signoretto2010nuclear}
M.~Signoretto, L.~De~Lathauwer, and J.~Suykens, ``Nuclear norms for tensors and
  their use for convex multilinear estimation,'' \emph{Technical Report 10-186,
  ESAT-SISTA, K.U.Leuven}, 2010.

\bibitem{signoretto2011learning}
M.~Signoretto, Q.~Dinh, L.~De~Lathauwer, and J.~Suykens, ``Learning with
  tensors: a framework based on convex optimization and spectral
  regularization,'' Internal Report 11-129, ESATSISTA, KU Leuven (Leuven,
  Belgium), Tech. Rep., 2011.

\bibitem{semerci:12:air}
O.~Semerci, N.~Hao, M.~E. Kilmer, and E.~L. Miller, ``An iterative
  reconstruction method for spectral {CT} with tensor-based formulation and
  nuclear norm regularization,'' in \emph{Proc. 2nd Intl. Mtg. on image
  formation in X-ray CT}, 2012, pp. {314--7}.

\bibitem{Kilmer2011641}
M.~E. Kilmer and C.~D. Martin, ``Factorization strategies for third-order
  tensors,'' \emph{Linear Algebra and its Applications}, vol. 435, no.~3, pp.
  641 -- 658, 2011, <ce:title>Special Issue: Dedication to Pete Stewart on the
  occasion of his 70th birthday</ce:title>. [Online]. Available:
  \url{http://www.sciencedirect.com/science/article/pii/S0024379510004830}

\bibitem{kilmer2011third}
M.~KILMER, K.~BRAMAN, N.~HAO, and R.~Hoover, ``Third order tensors as operators
  on matrices: a theoretical and computational framework with applications in
  imaging,'' \emph{Department of Computer Science, Tufts University}, 2011.

\bibitem{rudin1992nonlinear}
L.~Rudin, S.~Osher, and E.~Fatemi, ``Nonlinear total variation based noise
  removal algorithms,'' \emph{Physica D: Nonlinear Phenomena}, vol.~60, no.~1,
  pp. 259--268, 1992.

\bibitem{sidky2008image}
E.~Sidky and X.~Pan, ``Image reconstruction in circular cone-beam computed
  tomography by constrained, total-variation minimization,'' \emph{Physics in
  medicine and biology}, vol.~53, p. 4777, 2008.

\bibitem{tang2009performance}
J.~Tang, B.~Nett, and G.~Chen, ``Performance comparison between total variation
  (tv)-based compressed sensing and statistical iterative reconstruction
  algorithms,'' \emph{Physics in Medicine and Biology}, vol.~54, p. 5781, 2009.

\bibitem{golub1996matrix}
G.~Golub and C.~Van~Loan, \emph{Matrix computations}.\hskip 1em plus 0.5em
  minus 0.4em\relax Johns Hopkins University Press, 1996, vol.~3.

\bibitem{semerci2012tomographic}
O.~Semerci and E.~Miller, ``Tomographic imaging-a parametric level-set approach
  to simultaneous object identification and background reconstruction for
  dual-energy computed tomography,'' \emph{IEEE Transactions on Image
  Processing}, vol.~21, no.~5, p. 2719, 2012.

\bibitem{beutel2000handbook}
J.~Beutel, \emph{Handbook of medical imaging: Physics and psychophysics}.\hskip
  1em plus 0.5em minus 0.4em\relax Spie Press, 2000, vol.~1.

\bibitem{schmidt2012empirical}
T.~Schmidt, ``An empirical method for correcting the detector spectral response
  in energy-resolved ct,'' in \emph{Proceedings of SPIE}, vol. 8313, 2012, p.
  831312.

\bibitem{ye2005generalized}
J.~Ye, ``Generalized low rank approximations of matrices,'' \emph{Machine
  Learning}, vol.~61, no.~1, pp. 167--191, 2005.

\bibitem{jolliffe2005principal}
I.~Jolliffe, \emph{Principal component analysis}.\hskip 1em plus 0.5em minus
  0.4em\relax Wiley Online Library, 2005.

\bibitem{lefkimmiatis2012hessian}
S.~Lefkimmiatis, J.~Ward, and M.~Unser, ``A hessian schatten-norm
  regularization approach for solving linear inverse problems,'' \emph{arXiv
  preprint arXiv:1209.3318}, 2012.

\bibitem{fazel2001rank}
M.~Fazel, H.~Hindi, and S.~Boyd, ``A rank minimization heuristic with
  application to minimum order system approximation,'' in \emph{American
  Control Conference, 2001. Proceedings of the 2001}, vol.~6.\hskip 1em plus
  0.5em minus 0.4em\relax IEEE, 2001, pp. 4734--4739.

\bibitem{turk1991face}
M.~Turk and A.~Pentland, ``Face recognition using eigenfaces,'' in
  \emph{Computer Vision and Pattern Recognition, 1991. Proceedings CVPR'91.,
  IEEE Computer Society Conference on}.\hskip 1em plus 0.5em minus 0.4em\relax
  IEEE, 1991, pp. 586--591.

\bibitem{hsieh2012low}
C.~Hsieh, K.~Chiang, and I.~Dhillon, ``Low rank modeling of signed networks,''
  in \emph{Proceedings of the 18th ACM SIGKDD international conference on
  Knowledge discovery and data mining}.\hskip 1em plus 0.5em minus 0.4em\relax
  ACM, 2012, pp. 507--515.

\bibitem{donoho2006compressed}
D.~Donoho, ``Compressed sensing,'' \emph{Information Theory, IEEE Transactions
  on}, vol.~52, no.~4, pp. 1289--1306, 2006.

\bibitem{sauer1993local}
K.~Sauer and C.~Bouman, ``A local update strategy for iterative reconstruction
  from projections,'' \emph{Signal Processing, IEEE Transactions on}, vol.~41,
  no.~2, pp. 534--548, 1993.

\bibitem{elad2005simultaneous}
M.~Elad, J.~Starck, P.~Querre, and D.~Donoho, ``Simultaneous cartoon and
  texture image inpainting using morphological component analysis (mca),''
  \emph{Applied and Computational Harmonic Analysis}, vol.~19, no.~3, pp.
  340--358, 2005.

\bibitem{boyd2011distributed}
S.~Boyd, N.~Parikh, E.~Chu, B.~Peleato, and J.~Eckstein, \emph{Distributed
  optimization and statistical learning via the alternating direction method of
  multipliers}.\hskip 1em plus 0.5em minus 0.4em\relax Now Publishers, 2011.

\bibitem{bertsekas1999nonlinear}
D.~Bertsekas, ``Nonlinear programming,'' 1999.

\bibitem{xu2011statistical}
Q.~Xu, X.~Mou, G.~Wang, J.~Sieren, E.~A. Hoffman, and H.~Yu, ``Statistical
  interior tomography,'' \emph{Medical Imaging, IEEE Transactions on}, vol.~30,
  no.~5, pp. 1116--1128, 2011.

\bibitem{ramani2012splitting}
S.~Ramani and J.~A. Fessler, ``A splitting-based iterative algorithm for
  accelerated statistical x-ray ct reconstruction,'' \emph{Medical Imaging,
  IEEE Transactions on}, vol.~31, no.~3, pp. 677--688, 2012.

\bibitem{vandeghinste2011split}
B.~Vandeghinste, B.~Goossens, J.~De~Beenhouwer, A.~Pizurica, W.~Philips,
  S.~Vandenberghe, and S.~Staelens, ``Split-bregman-based sparse-view ct
  reconstruction,'' in \emph{11th International meeting on Fully
  Three-Dimensional Image Reconstruction in Radiology and Nuclear Medicine
  (Fully 3D 11)}, 2011, pp. 431--434.

\bibitem{afonso2010fast}
M.~V. Afonso, J.~M. Bioucas-Dias, and M.~A. Figueiredo, ``Fast image recovery
  using variable splitting and constrained optimization,'' \emph{Image
  Processing, IEEE Transactions on}, vol.~19, no.~9, pp. 2345--2356, 2010.

\bibitem{figueiredo2010restoration}
M.~A. Figueiredo and J.~M. Bioucas-Dias, ``Restoration of poissonian images
  using alternating direction optimization,'' \emph{Image Processing, IEEE
  Transactions on}, vol.~19, no.~12, pp. 3133--3145, 2010.

\bibitem{chambolle2004algorithm}
A.~Chambolle, ``An algorithm for total variation minimization and
  applications,'' \emph{Journal of Mathematical imaging and vision}, vol.~20,
  no.~1, pp. 89--97, 2004.

\bibitem{vogel1998fast}
C.~Vogel and M.~Oman, ``Fast, robust total variation-based reconstruction of
  noisy, blurred images,'' \emph{Image Processing, IEEE Transactions on},
  vol.~7, no.~6, pp. 813--824, 1998.

\bibitem{beck2009fast}
A.~Beck and M.~Teboulle, ``A fast iterative shrinkage-thresholding algorithm
  for linear inverse problems,'' \emph{SIAM Journal on Imaging Sciences},
  vol.~2, no.~1, pp. 183--202, 2009.

\bibitem{jensen2012implementation}
T.~L. Jensen, J.~H. J{\o}rgensen, P.~C. Hansen, and S.~H. Jensen,
  ``Implementation of an optimal first-order method for strongly convex total
  variation regularization,'' \emph{BIT Numerical Mathematics}, vol.~52, no.~2,
  pp. 329--356, 2012.

\bibitem{slaney1988principles}
M.~Slaney and A.~Kak, ``Principles of computerized tomographic imaging,''
  \emph{SIAM, Philadelphia}, 1988.

\bibitem{guide1998mathworks}
M.~Guide, ``The mathworks inc,'' \emph{Natick, MA}, vol.~4, 1998.

\bibitem{de2004distance}
B.~De~Man and S.~Basu, ``Distance-driven projection and backprojection in three
  dimensions,'' \emph{Physics in medicine and biology}, vol.~49, no.~11, p.
  2463, 2004.

\bibitem{andrecut2009parallel}
M.~Andrecut, ``Parallel gpu implementation of iterative pca algorithms,''
  \emph{Journal of Computational Biology}, vol.~16, no.~11, pp. 1593--1599,
  2009.

\bibitem{woolfe2008fast}
F.~Woolfe, E.~Liberty, V.~Rokhlin, and M.~Tygert, ``A fast randomized algorithm
  for the approximation of matrices,'' \emph{Applied and Computational Harmonic
  Analysis}, vol.~25, no.~3, pp. 335--366, 2008.

\bibitem{berger1998xcom}
M.~Berger, J.~Hubbell, S.~Seltzer, J.~Chang, J.~Coursey, R.~Sukumar, and
  D.~Zucker, ``Xcom: Photon cross sections database,'' \emph{NIST Standard
  Reference Database}, vol.~8, pp. 87--3597, 1998.

\bibitem{ahn2008analysis}
S.~Ahn and R.~Leahy, ``Analysis of resolution and noise properties of
  nonquadratically regularized image reconstruction methods for pet,''
  \emph{Medical Imaging, IEEE Transactions on}, vol.~27, no.~3, pp. 413--424,
  2008.

\bibitem{ramani2012regularization}
S.~Ramani, Z.~Liu, J.~Rosen, J.~Nielsen, and J.~Fessler, ``Regularization
  parameter selection for nonlinear iterative image restoration and mri
  reconstruction using gcv and sure-based methods,'' \emph{Image Processing,
  IEEE Transactions on}, vol.~21, no.~8, pp. 3659--3672, 2012.

\bibitem{candes2012unbiased}
E.~Candes, C.~Sing-Long, and J.~Trzasko, ``Unbiased risk estimates for singular
  value thresholding and spectral estimators,'' \emph{arXiv preprint
  arXiv:1210.4139}, 2012.

\bibitem{golbabaee2011hyperspectral}
M.~Golbabaee and P.~Vandergheynst, ``Hyperspectral image compressed sensing via
  low-rank and joint-sparse matrix recovery,'' in \emph{Indernational
  Conference on Acoustics, Speech and Signal Processing ICASSP}, 2011.

\end{thebibliography}
\end{document}